\documentclass[pdflatex,sn-mathphys-num]{sn-jnl}

\usepackage{graphicx}%
\usepackage{multirow}%
\usepackage{amsmath,amssymb,amsfonts,bbm}%
\usepackage{amsthm}%
\usepackage{mathrsfs}%
\usepackage[title]{appendix}%
\usepackage{xcolor}%
\usepackage{textcomp}%
\usepackage{manyfoot}%
\usepackage{booktabs}%
\usepackage{booktabs}%
\usepackage{subcaption}%
\usepackage{algorithm}%
\usepackage{algorithmicx}%
\usepackage{algpseudocode}%
\usepackage{listings}%
\usepackage{booktabs}
\usepackage{tabularx}
\usepackage{siunitx}
\usepackage{xspace}

\usepackage{amsmath,amsfonts,bm}
\usepackage{xifthen}

\def\eqref#1{equation~\ref{#1}}

\def\1{\bm{1}}

\DeclareMathAlphabet{\mathsfit}{\encodingdefault}{\sfdefault}{m}{sl}
\SetMathAlphabet{\mathsfit}{bold}{\encodingdefault}{\sfdefault}{bx}{n}

\def\gG{{\mathcal{G}}}

\def\gR{{\mathcal{R}}}
\def\gS{{\mathcal{S}}}

\def\gX{{\mathcal{X}}}
\def\gY{{\mathcal{Y}}}

\newcommand{\funcd}[1][]{%
    \ifthenelse{\isempty{#1}}%
    {\mathrm{d}}%
    {\mathrm{d} \left( #1 \right) }%
}
\newcommand{\funcf}[1][]{%
    \ifthenelse{\isempty{#1}}%
    {\mathrm{f}}%
    {\mathrm{f} \left( #1 \right) }%
}

\newcommand{\fPhi}[1][]{%
    \ifthenelse{\isempty{#1}}%
    {\Phi}%
    {\Phi \left( #1 \right) }%
} 

\def\sR{{\mathbb{R}}}

\newcommand{\poisson}[1][]{%
    \ifthenelse{\isempty{#1}}%
    {\mathrm{Poisson}}%
    {\mathrm{Poisson(#1)}}%
} 
\newcommand{\uniform}[1][]{%
    \ifthenelse{\isempty{#1}}%
    {\mathrm{Uni}}%
    {\mathrm{Uni(#1)}}%
} 

\newcommand{\bern}[1][]{%
    \ifthenelse{\isempty{#1}}%
    {\mathrm{Bern}}%
    {\mathrm{Bern}\left(#1\right)}%
} 

\newcommand{\E}[2][]{%
    \ifx\\#1\\
        \mathbb{E}\left[#2\right]%
    \else
        \mathbb{E}_{#1}\left[#2\right]%
    \fi
}

\newcommand{\R}{\mathbb{R}}

\definecolor{emerald}{RGB}{80, 200, 120} 

\newcommand{\graphmodel}{\textit{THiGER}\xspace}

\newcommand{\llmmodel}{\textit{THiGERLLM}\xspace}

\newcommand{\agatha}{\textit{Agatha}\xspace}

\newcommand{\predmodel}{THiGERLLM}

\newcommand{\explainer}{\textit{PHELInE}\xspace}

\newcommand{\retrfn}{\mathrm{F}}
\newcommand{\rerankfn}{\mathrm{R}}

\newcommand{\true}{\text{True}}

\newcommand{\scorefn}[2][]{\mathrm{f}_{#1} \left( #2 \right)}
\newcommand{\scoref}[1][]{\mathrm{f}_{#1}}

\newcommand{\surro}[2][]{\mathrm{s}_{#1} \left( #2 \right)}

\newcommand{\Gtrain}[1][]{\mathcal{G}_{\mathrm{train}}^{#1}}
\newcommand{\Gunseen}[1][]{\mathcal{G}_{\mathrm{unseen}}^{#1}}

\newcommand{\suff}[1][]{\text{Suff}\left({#1} \right)}
\newcommand{\nece}[1][]{\text{Nec}\left({#1} \right)}

\newcommand{\prompt}{\mathbf{x}}
\NewDocumentCommand{\tok}{o}{%
  x\IfValueT{#1}{_{#1}}%
}

\newcommand{\nodeset}{\mathcal{V}}

\NewDocumentCommand{\relset}{o}{%
  \mathcal{R}\IfValueT{#1}{^{(#1)}}%
}

\NewDocumentCommand{\factset}{o}{%
  \mathcal{F}\IfValueT{#1}{^{(#1)}}%
}

\newcommand{\ctx}{\mathbf{x}}
\newcommand{\timevec}{\mathbf{t}}
\newcommand{\instance}[2]{(#1,#2,\ctx_{#1#2},\timevec_{#1#2})}
\newcommand{\supinstance}[2]{(#1,#2,\ctx_{#1#2},\mathbf{y}_{#1#2})}

\newcommand{\labscore}{\boldsymbol{\ell}}

\newcommand{\lbase}[1]{\labscore^{\text{base}}_{#1}}
\newcommand{\lbasee}[1]{\ell^{\mathrm{base}}_{#1}}

\newcommand{\ltext}[1]{\labscore^{\text{text}}_{#1}}
\newcommand{\ltexte}[1]{\ell^{\mathrm{text}}_{#1}}

\newcommand{\lgraph}[1]{\labscore^{\text{graph}}_{#1}}
\newcommand{\lgraphe}[1]{\ell^{\mathrm{graph}}_{#1}}

\newcommand{\lense}[1]{\ell^{\mathrm{ens}}_{#1}}

\newcommand{\pairemb}{\mathbf{p}}

\newcommand{\pairembt}[2]{\pairemb^{(#1)}_{#2}}

\theoremstyle{thmstyleone}%
\theoremstyle{thmstyletwo}%

\theoremstyle{thmstylethree}%

\begin{document}

\title[Hakken: Predicting future discoveries]{Hakken: Predicting future discoveries to fill the gaps in today's knowledge}


\author*[1]{\fnm{Tarek R.} \sur{Besold}}\email{tarek.besold@gmail.com}

\author[1]{\fnm{Uchenna} \sur{Akujuobi}}

\author[1]{\fnm{Pablo} \sur{Sanchez}}

\author[2]{\fnm{Alessandra} \sur{Toniato}}

\author[3]{\fnm{Kana} \sur{Maruyama}}

\author[3]{\fnm{Jihun} \sur{Choi}}

\author[1]{\fnm{Samy} \sur{Badreddine}}

\author[4]{\fnm{Frederick} \sur{Gifford}}

\author[6]{\fnm{Daniel} \sur{Evans-Yamamoto}}

\author[6]{\fnm{Sucheendra K.} \sur{Palaniappan}}

\author[1]{\fnm{Miquel} \sur{Ferrer}}

\author[3]{\fnm{Kae} \sur{Nagano}}

\author[1]{\fnm{Iris} \sur{Rossell}}

\author[1]{\fnm{Tom} \sur{Joy}}

\author[1]{\fnm{Hatem} \sur{ElShazly}}

\author[1]{\fnm{Chrysa} \sur{Iliopoulou}}

\author[1]{\fnm{Christoph} \sur{Wehner}}

\author[1]{\fnm{Thiviyan} \sur{Thanapalasingam}}

\author[1]{\fnm{Susana} \sur{Nunes}}

\author[1]{\fnm{Pedro G.} \sur{Cotovio}}

\author[5]{\fnm{Peter} \sur{Wurman}}

\author[5]{\fnm{Peter} \sur{Stone}}

\author[3,6,7]{\fnm{Hiroaki} \sur{Kitano}}

\author[3]{\fnm{Michael} \sur{Spranger}}

\affil*[1]{\orgname{SonyAI}, \orgaddress{\city{Barcelona}, \country{Spain}}}

\affil[2]{\orgname{SonyAI}, \orgaddress{\city{Zurich}, \country{Switzerland}}}

\affil[3]{\orgname{SonyAI}, \orgaddress{\city{Tokyo}, \country{Japan}}}

\affil[4]{\orgname{SonyAI}, \orgaddress{\city{London}, \country{UK}}}

\affil[5]{\orgname{SonyAI}, \orgaddress{\city{New York, NY}, \country{USA}}}

\affil[6]{\orgname{The Systems Biology Institute}, \orgaddress{\city{Tokyo}, \country{Japan}}}

\affil[7]{\orgname{Okinawa Institute of Science and Technology (OIST)}, \orgaddress{\city{Okinawa}, \country{Japan}}}


\abstract{
We present Hakken, a domain-agnostic prediction and explanation system performing \emph{knowledge prediction}, i.e., growing scientific knowledge by establishing novel relationships, ones that are not limited to the deductive hull of previous knowledge. 
Hakken uses a transformer-based prediction model built on temporal sequences of knowledge graphs extracted from vast bodies of research publications, fused with an LLM's semantic knowledge, to predict the presence and define the type of as-yet undocumented relationships between scientific concepts. It then calls a model-agnostic explanation framework to provide accompanying information for each prediction that allows scientists to evaluate the suggested new relationships. While general purpose, we demonstrate Hakken’s practical capabilities by applying it to the biomedical domain. There, Hakken's prediction model establishes a new benchmark for time-aware multi-label relation prediction, and we show that the model's outputs stay coherent and informative over extended time spans in historic data. In addition, we scored 1.5 million above-confidence-threshold hypotheses related to aging, qualitatively validated batches of these predictions with biologists and progressed three of them for empirical validation in wet-lab. Two predictions with potentially significant impact in the context of drug discovery and repurposing were confirmed, introducing previously undocumented interactions between TP53 and BAMBI, and between RAF1 and TNF, to biomedical science.
}

\keywords{Knowledge Prediction, Hypothesis Generation, Literature-Based Discovery, Explainability, Biomedical Science}



\maketitle

\section{Introduction}\label{sec1}

One of the foundational principles of science is ``standing on the shoulders of giants''---building on past discoveries to make new ones. For some time, this principle has been at risk: Scientific knowledge is being produced at unprecedented and ever-increasing speeds~\cite{nsfPublicationsOutput}, and individual researchers are practically unable to stay abreast of all potentially relevant developments in their fields. Additionally, alongside the growth and fragmentation of the knowledge base, the individual fragments themselves are becoming more complex, increasingly pushing the mental limits of scientists who may struggle to comprehend overly intricate sets of facts, and who may be more subject to biases and preconceptions in their assessment of novel insights. 
In contrast, from an AI perspective the exponentially growing number of publications offers a clear opportunity: The vast body of expert-reviewed research constitutes an evolving collection of highest-quality data concerning the governing laws of our physical world and the relationships between entities within it. Technologies like Literature Based Discovery (LBD)~\cite{bhasuran2023literaturebaseddiscoverylbd}, and more recently generative AI~\cite{alkan2025surveyhypothesisgenerationscientific}, have been deployed to help researchers handle the double-challenge posed by the growing volume and complexity of scientific knowledge, with varying but notable success. 

Going beyond these methods, we introduce \emph{Hakken} (named after the Japanese noun for ``discovery'' or ``finding''), an AI prediction and explanation system that uses the knowledge stored in research literature to predict novel knowledge in the form of as yet unpublished relationships between scientific concepts. In doing knowledge prediction, Hakken takes the next step past current methods that extract, simplify and collate existing knowledge for researchers, or suggest unidentified connections from within the deductive closure of pre-existing knowledge. Instead, due to its predictive nature, Hakken is closer in spirit to scientific AI tools like AlphaFold~\cite{jumper2021highly} or GNoME~\cite{merchant2023scaling}. Its output is a vast new dataset of novel predicted scientific relationships, many of them not inferrable based on today's state of the literature alone, all of them specific in the subject, the object, and the precise type of the relationship. Each prediction comes with a confidence score, representing the counterfactual probability that it could already exist in the literature given everything else that is known---i.e., the weight of evidence in favor of the hypothesis. 
%
%
Hakken additionally delivers an explainability function for researchers that identifies small sets of known facts
which help reveal the mechanisms underlying a prediction and the existing knowledge that supports it. This information helps to build confidence in the machine-generated insights and facilitates an informed decision to progress to validation studies.

While domain-general in the underlying models, we concretely developed and tested Hakken for the biomedical domain. We benchmarked its prediction model against alternative approaches on the same prediction task, and evaluated its performance against historic biomedical literature data. Additionally, we validated Hakken’s practical value in a collaborative setting with biomedical researchers, going through a full discovery cycle from hypothesis definition to wet-lab validation. We discovered and confirmed two previously undocumented biomedical facts in the form of a gene-gene interaction between TP53 and BAMBI, and an enzyme-gene interaction between RAF1 and TNF. 

In the following, we first introduce Hakken's prediction model \llmmodel (Temporal Hierarchical Graph-based Encoder Representation with LLM), the first approach that predicts future relation labels in a multi-label setting as the discovery target, models temporal evolution via a temporal Graph Neural Network~\cite{longa2023GNNs}, fuses an LLM’s semantic knowledge with the graph structure, and is successfully applied to biomedical knowledge prediction. We then describe PHELInE (Predicted Hypothesis Elucidation with Literature-Inferred Explanations), a novel model-agnostic explanation framework we deploy within Hakken to provide the already mentioned knowledge-graph information for \llmmodel's predictions. We continue with the results of the different benchmarks and a summary of the co-discovery process with biomedical researchers, before discussing the major opportunities that lie ahead on the path that Hakken opens.

\section{Predicting novel relationships with \llmmodel}\label{sec2}

\begin{figure}
    \centering
    \includegraphics[width=1\linewidth]{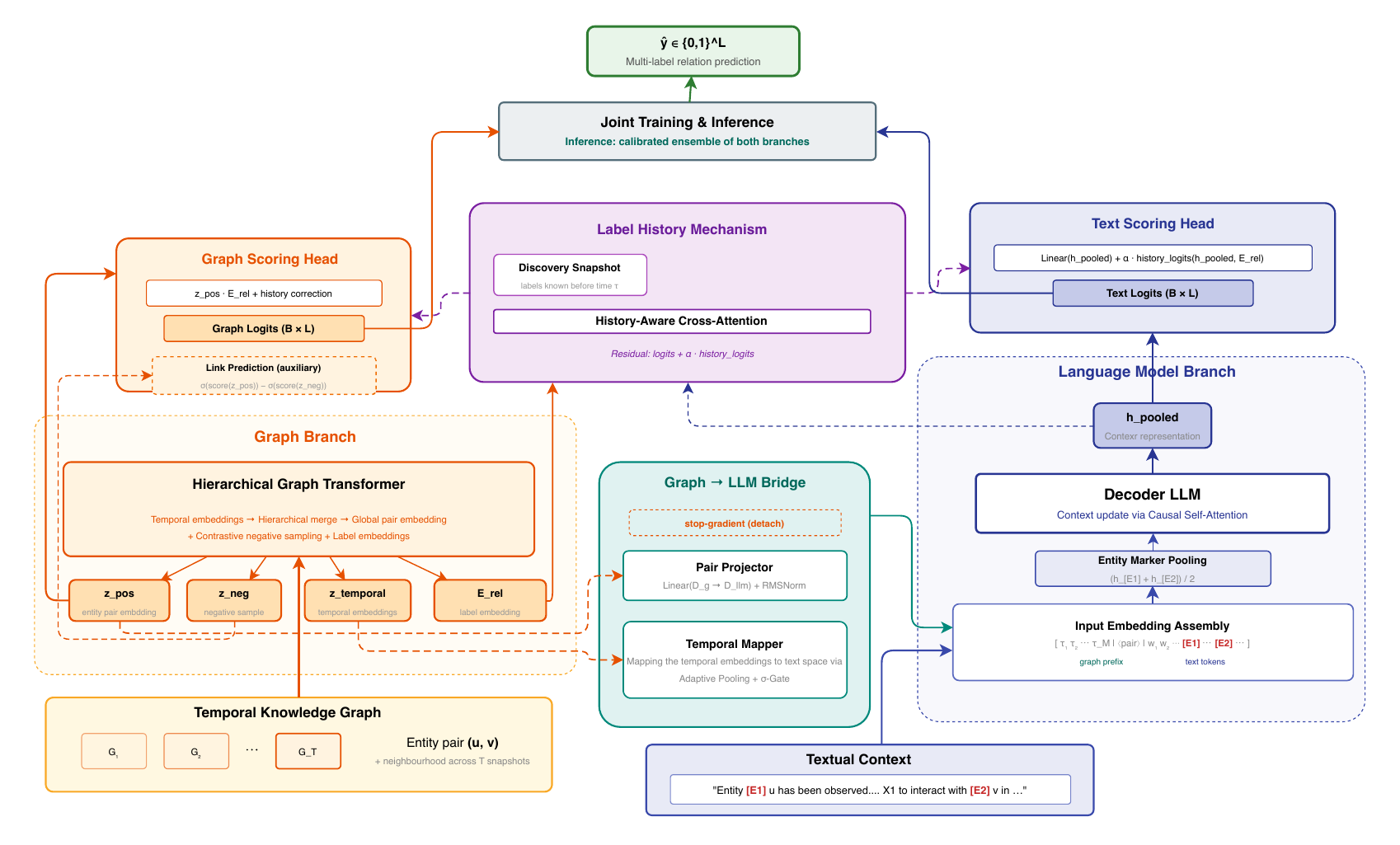}
    \caption{\textbf{\llmmodel for multi-label relation prediction.}
Given an entity pair $(u,v)$, the model combines a hierarchical temporal graph encoder with a language model. The graph branch summarizes structural and temporal evidence into compact graph embeddings, which are projected as prefix tokens and injected into the LLM together with publication-derived text. Graph-based and text-based label scores are then combined to produce the final predicted relations, and the resulting probabilities are calibrated to yield confidence scores.}
    \label{fig:core_model_overview}
\end{figure}

Our unified temporal graph and language model \llmmodel addresses literature-based knowledge prediction as a multi-label relation prediction task for an entity pair $(s,o)$ in a scientific-literature knowledge graph. As a simple example from the biomedical domain, $s$ may be a drug, such as Mefloquine, and $o$ a disease, such as malaria. One possible relation type $r$ could indicate that the drug prevents the disease, although multiple relation types may hold for the same pair. The model then asks which biomedical relations may hold between them.

In order to formally define the problem, we have to introduce some notation. Let $\nodeset$ denote the set of entities (graph nodes) and $\relset={1,\dots,C}$ the set of relation types.
The space of all possible triples is defined as $\Omega := \nodeset \times \gR \times \nodeset$.
A knowledge graph is represented as $\gG=(\nodeset,\relset,\factset)$, where each fact $(s,r,o)\in\factset\subseteq\Omega$ consists of a subject entity $s\in\nodeset$, an object entity $o\in\nodeset$, and a directed relation $r\in\relset$.
We consider a temporal knowledge graph represented as a sequence of $T$ graph snapshots
$\gG^{(t)}=(\nodeset,\relset[t],\factset[t]) \text{ with } t\in\{1,\dots,T\}$,
where $\factset[t]\subseteq\Omega$ denotes the set of triples observed within (or up to) time window $t$ and $\relset[t] \subseteq \relset$.
Given an entity pair $(s,o)\in\nodeset\times\nodeset$, the task is to predict the set of relations that hold between them, treating the pair as undirected so that $(s,o)$ and $(o,s)$ yield the same predictions.
We formulate this as a multi-label classification problem by associating each pair $(s,o)$ with a binary relation vector $\mathbf{y}_{so} \in \{0, 1\}^{|\relset|}$ where the $r$-th entry $y_{so, r}=1$ indicates that relation $r$ holds between $s$ and $o$; $y_{so, r}=0$ otherwise.

Building on our previous model THiGER \cite{akujuobi_link_2024}, which used temporal graph evidence alone to predict whether a relationship exists between two entities but not its type, \llmmodel takes as input the ontology-based node representations of $s$ and $o$, the temporally evolving graph around the pair, and text evidence from research publications. The text evidence consists of sentences---sourced from articles available via PMC Open Access---mentioning each entity separately, without mentioning the other entity. The graph records how related findings accumulate over time, while the text provides scientific context from the literature. 

For each candidate relation $r$, the model outputs a prediction for the hypothesis $(s,r,o)$ together with a confidence score, which is subsequently calibrated \cite{Platt1999ProbabilisticOF,Zadrozny2001ObtainingCP} to represent the estimated probability that the triple $(s,r,o)$ is true, i.e., that predictions assigned probability $p$ are correct with empirical frequency approximately $p$.
This task is challenging because the evidence is structural, temporal, and semantic at the same time. To address this, \llmmodel uses two branches that are learned jointly, as shown in Fig.~\ref{fig:core_model_overview}: a graph branch that captures how evidence around $(s,o)$ changes across time, and a language branch that interprets the pair using publication text. Each branch produces relation scores, and the final prediction combines them so that structural and semantic evidence can complement each other. To further refine these predictions, the model also uses label history, meaning relations already known for the pair up to the current time, as additional context. 

In the graph branch of \llmmodel, entity representations are first constructed from learned node features together with available ontology and domain information, and are updated by neighborhood aggregation with a GraphSAGE-style encoder~\cite{hamilton2017graphsage}. For the target pair, this yields a sequence of pair representations over time, which is processed by a temporal encoder \cite{Dao2022FlashAttentionFA,Vaswani2017AttentionIA} to capture persistence, accumulation, and change in relational evidence. The encoder produces a global pair embedding for graph-based prediction and a short temporal sequence that preserves coarse temporal structure. These outputs are then projected into a small set of learned vectors that are prepended to the language-model input, allowing the language branch to use a compact summary of the graph and its temporal evolution.

The language branch processes the publication text associated with each entity pair together with the prepended graph vectors.
For the present study, we use Mistral-7B-Instruct-v0.3 
(released May 22, 2024) as the underlying large language model.
For each entity, we retrieve 20 randomly sampled sentences that mention it individually from publications dated prior to the training-time cutoff.
This constraint reduces the likelihood that predictions rely on future task-specific information and encourages the model to ground its outputs in the provided evidence.
However, although it is very unlikely, because the model was pretrained on broad corpora with an unspecified cutoff date, residual temporal leakage cannot be fully excluded.
When entity markers are available, the model uses the hidden states at the marked positions to build a text representation focused on the target pair; otherwise, it uses the final valid token representation.
This text representation is used to produce relation scores from the text, while the graph embedding produces a separate set of relation scores from the graph alone. 
The two sets of scores are then combined in a label-wise ensemble so that the model can benefit from both sources, which may be informative in different situations. 
We also incorporate label history: relations already known for the pair up to the current time are encoded as contextual cues that adjust the candidate relation scores in both branches, helping the model account for prior knowledge and dependencies among relation labels. 
In this way, the model first learns from known examples during training and then applies what it has learned to new entity pairs during inference to predict relations.

Training is designed for incomplete supervision, where observed labels are treated as positives but unobserved labels are not assumed to be negatives. We therefore use a class-balanced multi-label objective that downweights unlabeled entries, and the graph branch is additionally trained with an auxiliary ranking loss. During inference, the graph and text scores are combined in a label-wise ensemble and then calibrated on a held-out validation set before per-label thresholding. We report the resulting calibrated probabilities as confidence scores. 
Overall, \llmmodel combines temporal graph evidence with publication text to produce more reliable knowledge predictions than either source alone.

\section{Explaining novel relationships with \explainer}

\begin{figure}
    \centering
    \includegraphics[width=1\linewidth]{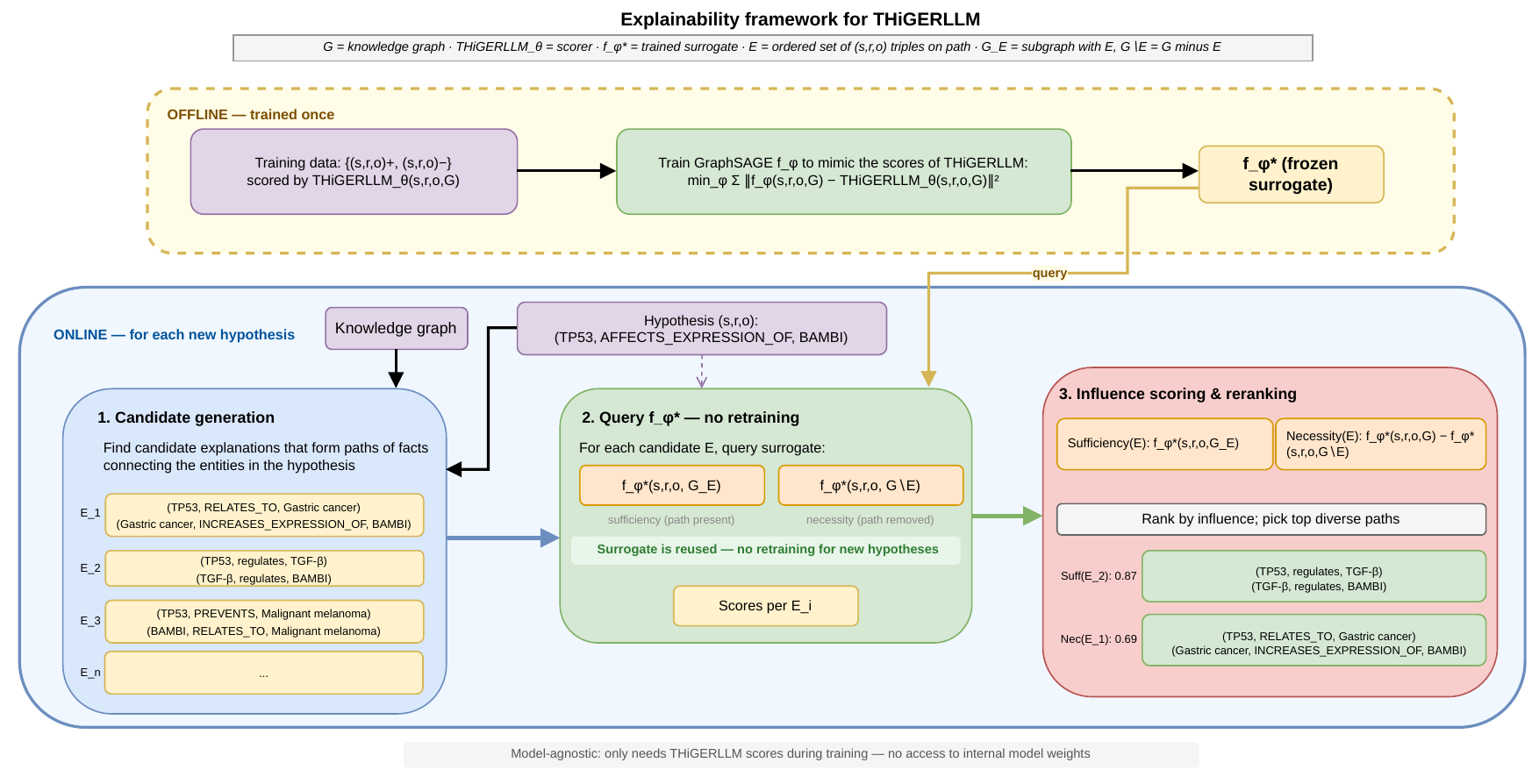}

    \caption{\textbf{\explainer providing explanations for \llmmodel's predictions.} A surrogate model is trained once (offline) to approximate prediction scores of \llmmodel. At inference time (online), for each hypothesis---a candidate triple (subject, relation, object)---we enumerate relational paths in the knowledge graph that connect the entities. For each path, we query the surrogate on subgraphs with and without that path to compute sufficiency (how much the path supports the prediction) and necessity (how much removing it weakens the prediction). Paths are ranked by influence and reranked by diversity to produce final explanations.}
    \label{fig:explainability_framework}
\end{figure}

In scientific contexts, interpretability is essential as researchers must be able to assess how a prediction relates to existing knowledge.
To address this challenge, we developed \explainer, a model-agnostic explanation framework (Figure~\ref{fig:explainability_framework}) that explains relation predictions $(s,r,o)$ on knowledge graphs through relational patterns, i.e., contextual subgraphs of facts.
Specifically, \explainer identifies small sets of related facts---typically forming multi-hop paths between subject and object---that significantly affect the model’s predicted score when present or removed.
For the wet-lab validation experiments (Section~\ref{subsec:wet-lab-validation}), candidate explanations were restricted to shortest subject--object paths, which were typically 2--4 hops long.
This design is motivated by the observation that knowledge is inherently structured as interconnected relationships rather than isolated facts \cite{Nickel2015ARO}, where relational patterns often capture known mechanisms (e.g., pathways, interactions, or functional associations) \cite{Barabsi2004NetworkBU,Peng2023KnowledgeGO} and allow researchers to assess whether predictions align with established domain knowledge.

\explainer interacts with the prediction model only through queries to its scoring function and therefore does not require access to model parameters or the training procedure.
The framework comprises three main components: initial path-based candidate explanation generation, subsequent surrogate-based pseudo-retraining to estimate prediction changes under graph perturbations, and final influence-based explanation scoring and reranking.

The initial candidate-generation stage identifies candidate explanations within a constrained search space of facts. 
In practice, explanations are constructed as paths of length $k$ connecting the entities involved in the predicted relationship.
Candidate paths are obtained by enumerating relational paths in the observed knowledge graph that connect the subject entity and object entity of the hypothesis. All triples occurring along these paths are collected as candidate explanatory facts. Such paths capture multi-hop relational patterns present in the knowledge graph and often reflect meaningful contextual relationships in scientific data.

As a next step, evaluating the influence of a candidate explanation requires estimating how the prediction of the model would change if the explanation facts were added to or removed from the training graph. 
Directly retraining the prediction model for each candidate explanation would be computationally infeasible.
To overcome this limitation, we introduce a pseudo-retraining procedure based on a surrogate model trained to approximate the scoring behaviour of the original predictor.
The surrogate learns to reproduce the scores produced by the original model while conditioning on the structure of the training graph. Once trained, it can efficiently estimate how predictions would change under hypothetical modifications to the graph, enabling rapid evaluation of candidate explanations without repeated retraining of the original model.
In our implementation, the surrogate model is a GraphSAGE graph neural network~\cite{hamilton2017graphsage} trained to predict the scores of the original predictive model. 
Once trained, it allows approximation of score changes for many candidate explanations simply by conditioning on different sets of facts.

In a final scoring and reranking step, candidate explanations are evaluated according to how strongly they influence the predicted score of the hypothesis triple.
A sufficiency score measures whether the triples in the explanation alone can reproduce a high prediction score. 
To compute this quantity, we evaluate the surrogate model on a graph containing only the candidate explanation triples and measure the resulting score for the hypothesis \cite{Lonardi2025UnifyingPE}.
A necessity score measures how strongly the prediction deteriorates when the explanation triples are removed from the training graph. 
This is estimated by evaluating the surrogate model on the graph obtained by removing the candidate explanation from the training graph and measuring the resulting drop in the predicted score \cite{Lonardi2025UnifyingPE}.
These two objectives capture complementary aspects of model behaviour: sufficiency identifies triples that independently support the prediction, whereas necessity identifies triples whose absence substantially weakens it.

After computing these influence scores for all candidate explanations, the framework selects the highest-scoring candidates---according to the chosen score strategy---and applies a reranking step to produce the final explanations. 
The reranking prioritizes explanations corresponding to distinct relational paths, ensuring that the final explanations capture diverse reasoning patterns rather than multiple variants of the same path structure. 
By promoting explanations that involve different intermediate entities and relations, this step encourages the discovery of multiple plausible mechanisms that could support the predicted relationship.

\section{Evaluation in Biomedical Science}

In order to assess Hakken's performance and practical usability and relevance, we chose to apply Hakken to research in biomedical science. This matches a general trend where AI models are increasingly demonstrating their potential in the domain~\cite{singhal2023large,saab2024capabilities,tian2024opportunities}. 

\subsection{Biomedical Data}

The primary dataset used for the evaluation of Hakken in biomedical science was made up of information derived from PMC Open Access (full text articles) and MEDLINE (titles and abstracts), together with data from a commercially licensed dataset. 
Initially, the dataset contains \num{249329650} raw triples, where each triple consists of two entities and the directed relationship connecting them (see formal definition in Section \ref{sec2}). Each triple is also associated with a timestamp indicating the year in which the article from which the triple has been derived was published and other auxiliary information (e.g. the original text sentence where available from the Open Access content used, the publication source, etc.). Alongside these triples, ontology files were provided that map each entity to one or more hierarchical domain descriptors, which we used to standardize entity domains and guide the subsequent data-cleaning steps (see Appendix~\ref{appendix_data} for details). The final cleaned dataset contains \num{254806} entities and \num{7127960} triples across 23 relation types and 20 macro-domains.

\subsection{Model Benchmarking}

Using the biomedical data at our disposal, we evaluate \llmmodel against a diverse set of baseline models for multi-label relation prediction under a temporally separated protocol designed to prevent training-time information leakage. 
We assess model performance on temporally held-out data, testing its ability to identify relations that emerge in later years.
The selected baselines represent complementary modeling paradigms: A random baseline assigning relationships randomly among candidate relation types, a structural knowledge-graph model based on ComplEx~\cite{trouillon_complex_2016}, k-nearest neighbors (kNN) over subject-object representations, a multi-label multi-layer perceptron (MLP) classifier, a rule-based multi-label classifier learning per-node label profiles and label-specific neighbor sets then scoring each label, a temporal graph-modeling baseline based on tNodeEmbed~\cite{singer_node_2019}, and our previous model \graphmodel~\cite{akujuobi_link_2024} that integrates temporal and structural signals for relational hypothesis ranking.

Table~\ref{tab:baseline-benchmark} summarizes both classification-based and ranking-based metrics using a 2020 temporal cut-off. 
All relation instances represented as triples $(s,r,o)$ first observed on or before 2020 are used for model training. Relation instances first observed after 2020 are reserved for evaluation. Models perform multi-label edge prediction by assigning one or more candidate relation types to a given subject--object entity pair. A predicted relation type is considered a true positive if the corresponding relation instance is first observed after 2020. 
Predictions not observed after 2020 are treated as negative instances for metric computation within the evaluation period. 
%
Appendix~\ref{appendix_benchmarks} reports further experiments, including a top-$m$ analysis examining the ranking behavior under a fixed prediction budget.

Overall, \llmmodel 
consistently ranks among the best-performing models across the evaluated metrics.
When focusing on weighted F1 or $nDCG_{\text{mean}}$, our previous model \graphmodel pulls slightly ahead of \llmmodel.
Weighted F1 emphasizes frequent relation types, while $nDCG_{\text{mean}}$ evaluates the quality of the predicted ranking.
Together, these results suggest that \graphmodel performs particularly well on dominant relations and produces slightly sharper rankings of the most relevant candidates.
However, while still closely tracking \graphmodel on weighted F1 or $nDCG_{\text{mean}}$, the trend reverses for macro F1, where \llmmodel achieves higher performance. 
Because macro metrics weight all relations equally, this indicates that \llmmodel performs better on average across relation types.
By additionally incorporating textual information, our new model substantially improves macro recall while maintaining similar weighted precision, despite some reduction in macro precision. 
This indicates recovery of more correct triples across relation types---particularly benefiting less frequent relations---which explains the improvement in macro F1.

When interpreting these results, it is important to consider the evaluation setting---and the application context---in a scientific domain.
While recall depends on correctly predicted true positives, precision implicitly assumes that predicted negatives correspond to true negatives. 
In practice, many negatives correspond to hitherto unobserved---rather than truly false---facts, meaning that predictions counted as false positives may actually represent valid but not yet labeled $(s,r,o)$ triples.
Consequently, improvements in recall as shown by \llmmodel are particularly meaningful, as they reflect the ability to recover true relations without relying on the completeness of negative labels.
We hypothesize that this improvement stems from the additional semantic information provided by the text component.
Graph structure and temporal signals capture relational patterns and dynamics, in addition to which textual descriptions provide complementary background knowledge about entities and relations. 
Moreover, this information is particularly valuable when graph connectivity is sparse---common for rare relations---where textual context can help infer plausible connections that are weakly supported by structural evidence alone. 
In science, rare relations often fall into at least one of three categories, corresponding to infrequently studied, novel, or quite specific types of interaction. 
For predicted scientific knowledge, novelty and specificity are strongly desirable properties, as is the ability for Hakken to also cover potentially understudied types of concept connections. 
This makes \llmmodel's consistent performance across relation types a key asset regarding Hakken's practical application value.

\begin{table}[h]
    \caption{Baseline comparison for time-aware multi-label relation prediction for a 2020 cut-off. All metrics are reported in percent to improve readability.}
    \centering
    \begin{tabular}{lccccccc}
    \toprule
    \textbf{Model} & $\mathrm{Prec}_{\mathrm{macro}}$ & $\mathrm{Rec}_{\mathrm{macro}}$ & $F_{1,\mathrm{macro}}$ & $\mathrm{Prec}_{\mathrm{w}}$ & $\mathrm{Rec}_{\mathrm{w}}$ & $F_{1,\mathrm{w}}$ & $\mathrm{nDCG}_{\mathrm{mean}}$ \\
    \midrule
    Random & 6.08 & 6.86 & 3.78 & 18.21 & 6.07 & 8.36 & 35.20 \\
    ComplEx & 5.34 & 8.29 & 5.80 & 17.15 & 19.68 & 18.15 & 47.21 \\
    KNN & 33.00 & 26.85 & 28.53 & 60.93 & 58.52 & 59.10 & 74.59 \\
    MLP & 39.46 & 26.81 & 28.75 & 66.96 & 60.19 & 61.52 & 86.81 \\
    RuleBased & 12.81 & 56.39 & 19.08 & 29.43 & $\mathbf{91.08}$ & 42.22 & 78.98 \\
    TNodeEmbed & 40.57 & 27.41 & 29.17 & 68.94 & 62.42 & 63.47 & 87.63 \\
    THiGER & $\mathbf{60.20}$ & 46.13 & 50.98 & $\mathbf{86.15}$ & 71.66 & $\mathbf{77.97}$ & $\mathbf{92.19}$ \\
    THiGERLLM & 53.77 & $\mathbf{60.72}$ & $\mathbf{51.16}$ & 84.01 & 66.54 & 73.73 & 90.30 \\
    \bottomrule
    \end{tabular}
    \label{tab:baseline-benchmark}
\end{table}

\subsection{Back-testing the Model on Historic Data}

Beyond strong performance on standard benchmarks, a temporal prediction model such as \llmmodel must demonstrate practical scientific relevance. In particular, it should not only generate plausible hypotheses from the knowledge available at a given point in time, but also maintain their validity as the time horizon extends—without degrading to random performance for more distant future predictions. In other words, meaningful signal should persist even as the evaluation window shifts further forward in time.
To evaluate these capabilities, we simulate historical forecasting conditions by restricting the model to past knowledge and then testing whether its proposed hypotheses remain meaningful over the following decade.
To quantify this behavior, we evaluate the model using disjoint future time bins. For a given cutoff time $t_k$, we partition the future into non-overlapping intervals such as $(t_k, t_k + 2],(t_k + 2, t_k + 4],\ldots,(t_k + 8, t_k + 10]$. Within each interval, we compute recall as the fraction of relations that first appear in that time window and were correctly predicted using only information available up to $t_k$. This provides a time-resolved view of predictive performance: early bins capture short-term, quickly testable hypotheses, while later bins reflect longer-term predictions that may depend on future advances. This separation allows us to distinguish between immediate and delayed validation of model-generated hypotheses.
An important consequence of this setup is that recall is well-defined temporally, since we can identify when a true relation first appears in the literature and check whether it was predicted in advance. In contrast, precision becomes fundamentally ill-defined in this setting. While we can label true positives (relations later observed) and false negatives (relations missed by the model), we cannot reliably label false positives, as predictions that are not yet observed may simply correspond to discoveries that have not been made—or not yet published—within the evaluation window. For this reason, temporal recall and ranking-based metrics provide a more meaningful assessment of predictive performance in this context.

\begin{figure*}[!h]
\centering
\begin{subfigure}[t]{0.32\textwidth}
    \centering
    \includegraphics[width=\linewidth]{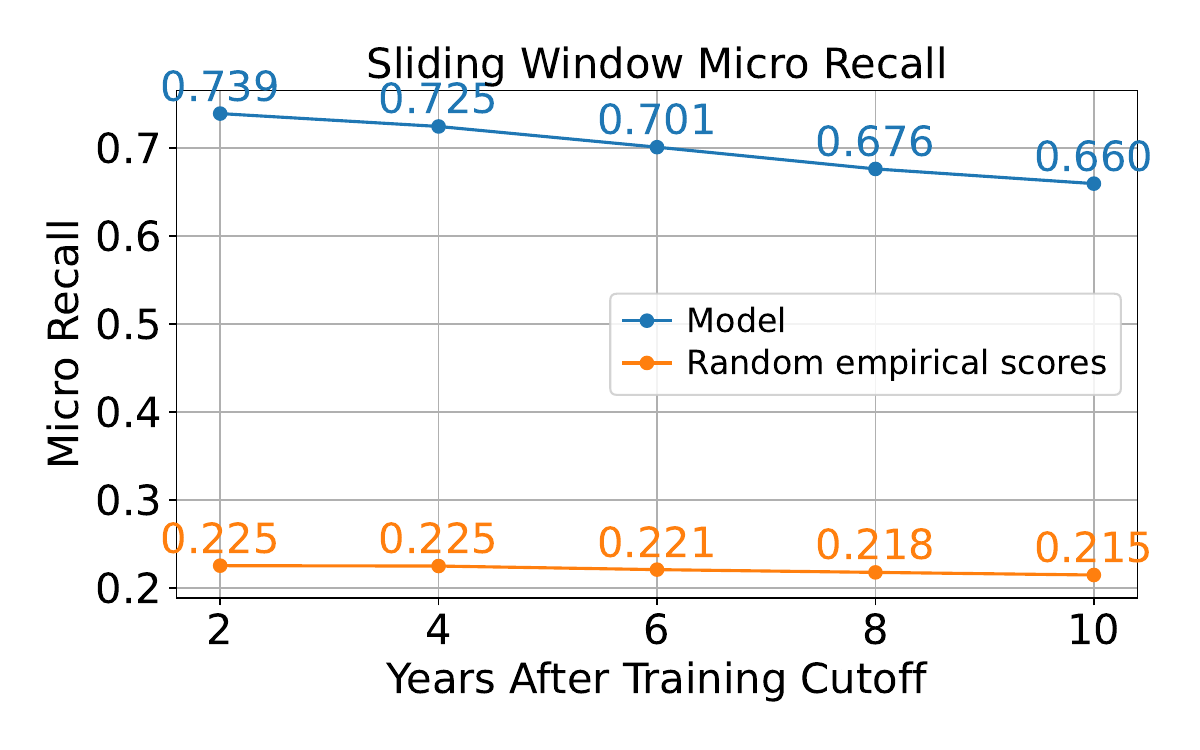}
    \caption{Interval Micro Recall for training cutoff 1990.}
    \label{fig:recall_graph_y1990}
\end{subfigure}
\hfill
\begin{subfigure}[t]{0.32\textwidth}
    \centering
    \includegraphics[width=\linewidth]{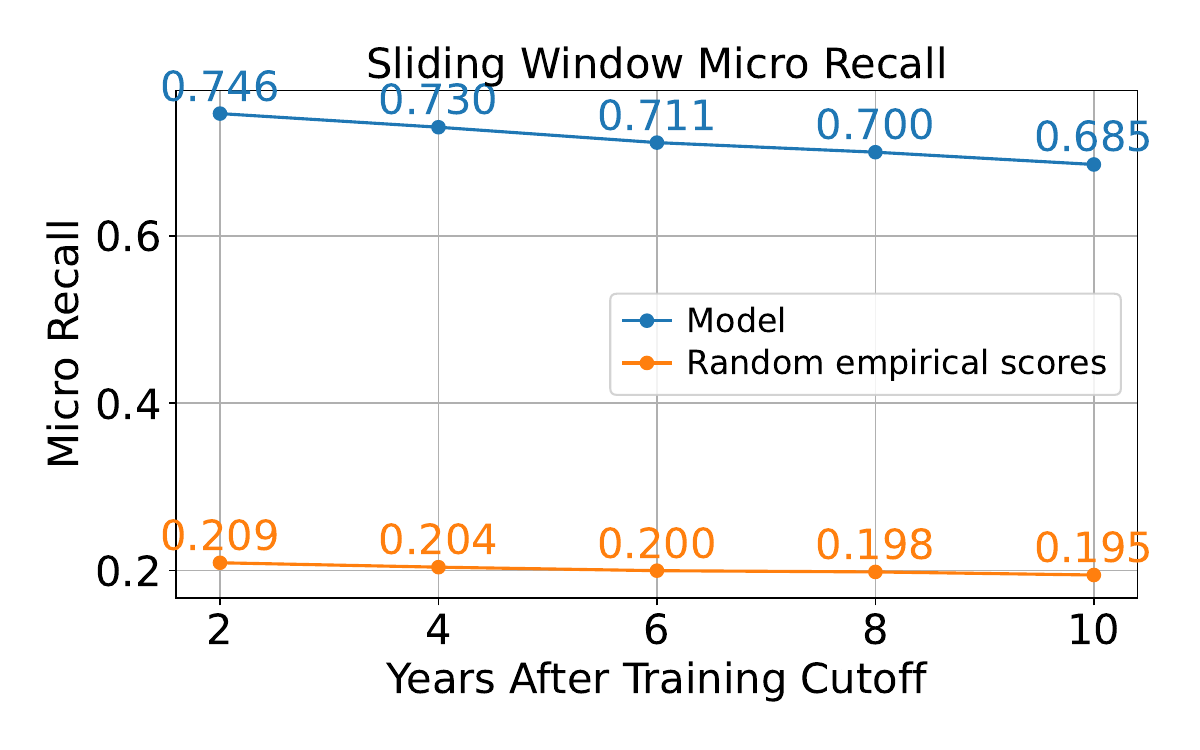}
    \caption{Interval Micro Recall for training cutoff 2000.}
    \label{fig:recall_graph_y2000}
\end{subfigure}
\hfill
\begin{subfigure}[t]{0.32\textwidth}
    \centering
    \includegraphics[width=\linewidth]{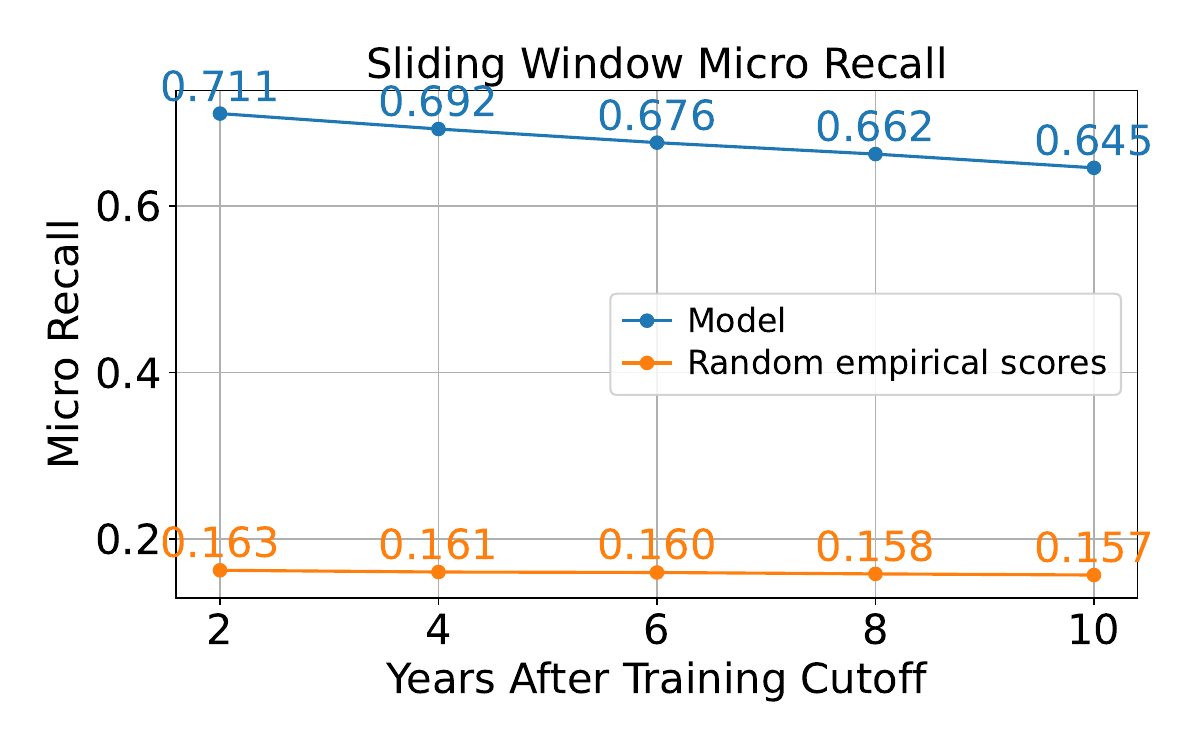}
    \caption{Interval Micro Recall for training cutoff 2010.}
    \label{fig:recall_graph_y2010}
\end{subfigure}

\vspace{0.5cm}

\begin{subfigure}[t]{0.32\textwidth}
    \centering
    \includegraphics[width=\linewidth]{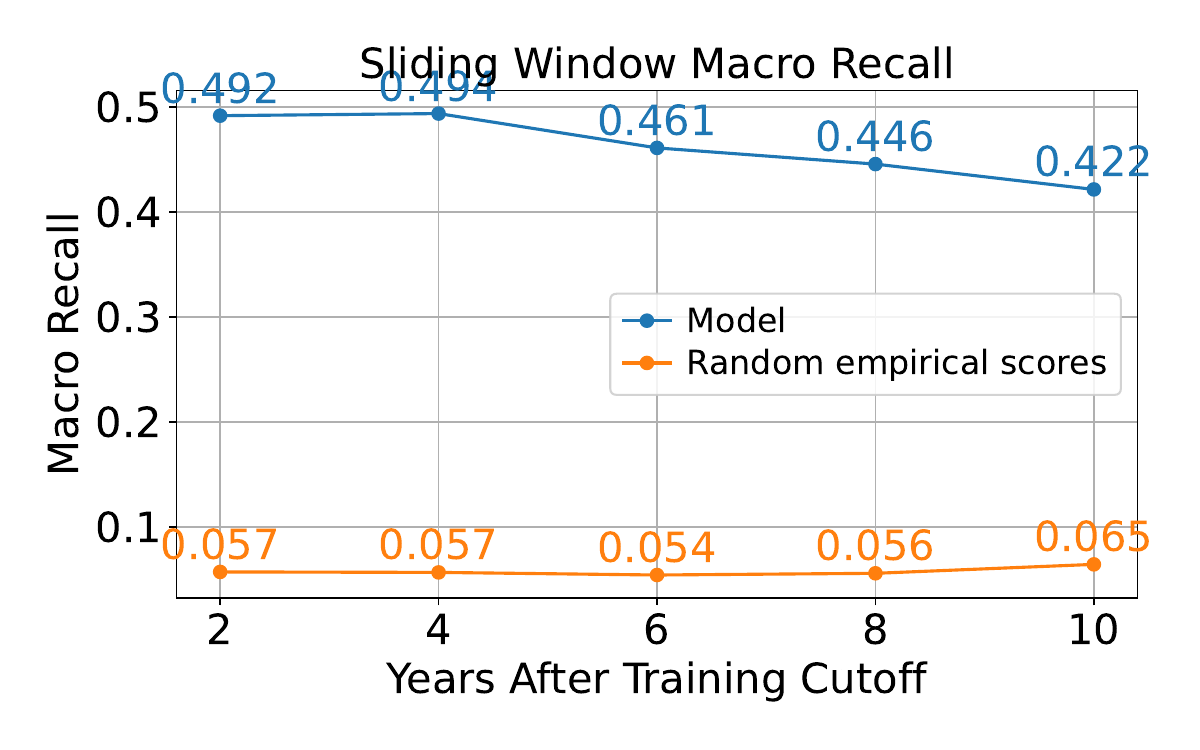}
    \caption{Interval Macro Recall for training cutoff 1990.}
    \label{fig:precision_graph_y1990}
\end{subfigure}
\hfill
\begin{subfigure}[t]{0.32\textwidth}
    \centering
    \includegraphics[width=\linewidth]{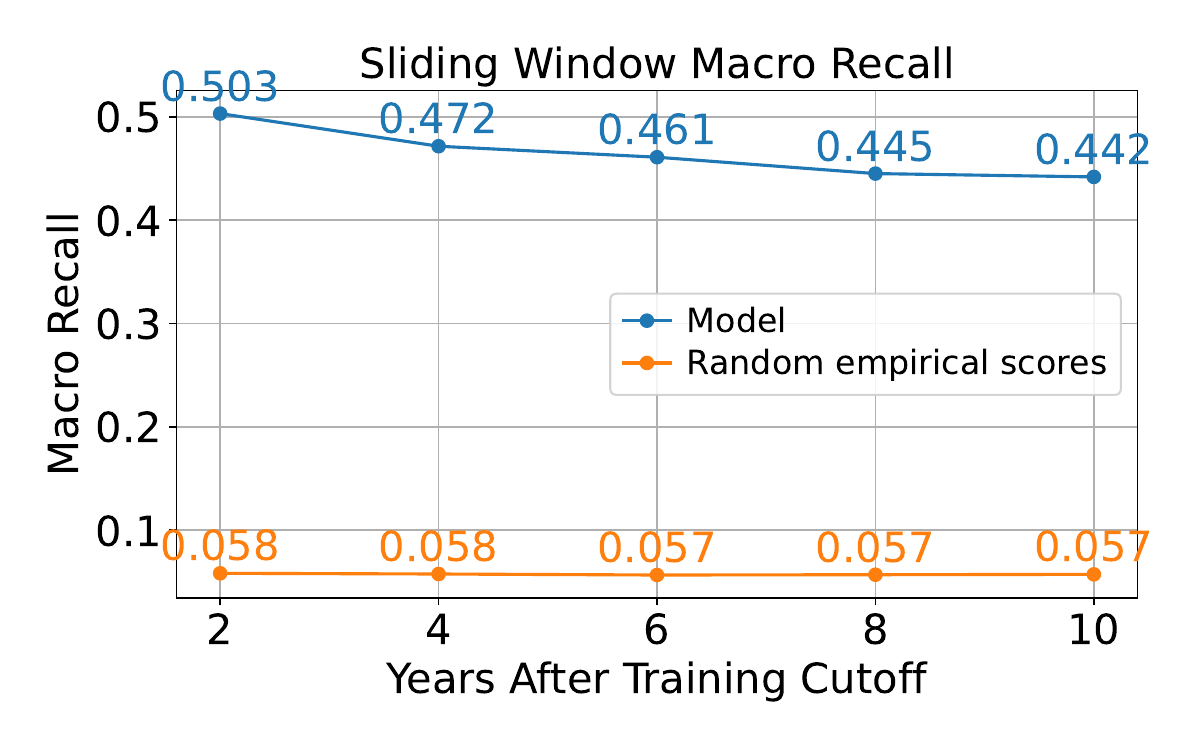}
    \caption{Interval Macro Recall for training cutoff 2000.}
    \label{fig:precision_graph_y2000}
\end{subfigure}
\hfill
\begin{subfigure}[t]{0.32\textwidth}
    \centering
    \includegraphics[width=\linewidth]{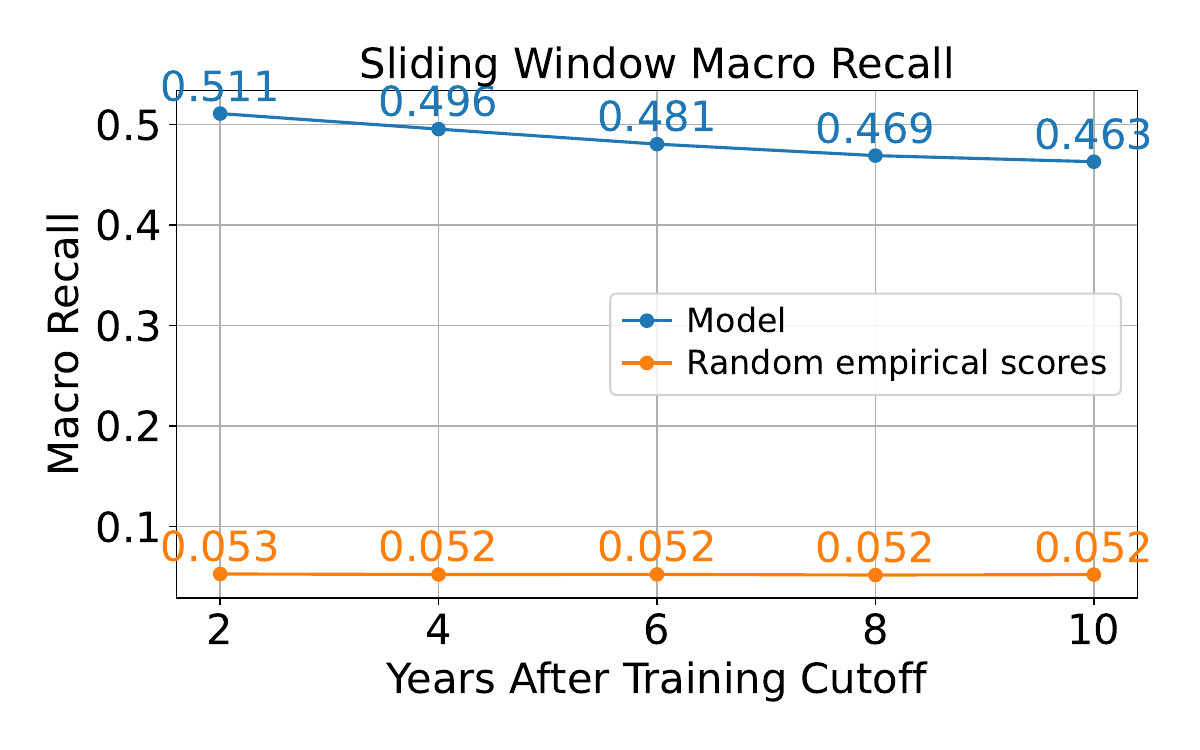}
    \caption{Interval Macro Recall for training cutoff 2010.}
    \label{fig:precision_graph_y2010}
\end{subfigure}

\vspace{0.5cm}

\begin{subfigure}[t]{0.32\textwidth}
    \centering
    \includegraphics[width=\linewidth]{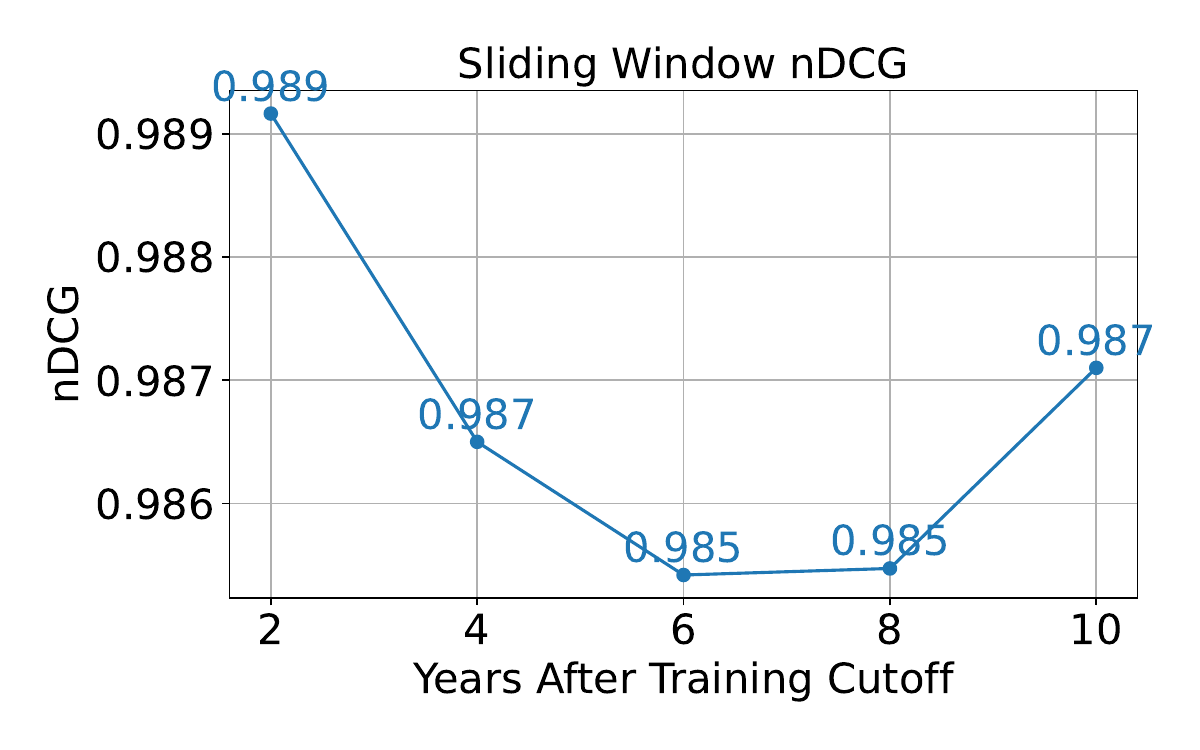}
    \caption{Mean nDCG for training cutoff 1990.}
    \label{fig:nDCG_graph_y1990}
\end{subfigure}
\hfill
\begin{subfigure}[t]{0.32\textwidth}
    \centering
    \includegraphics[width=\linewidth]{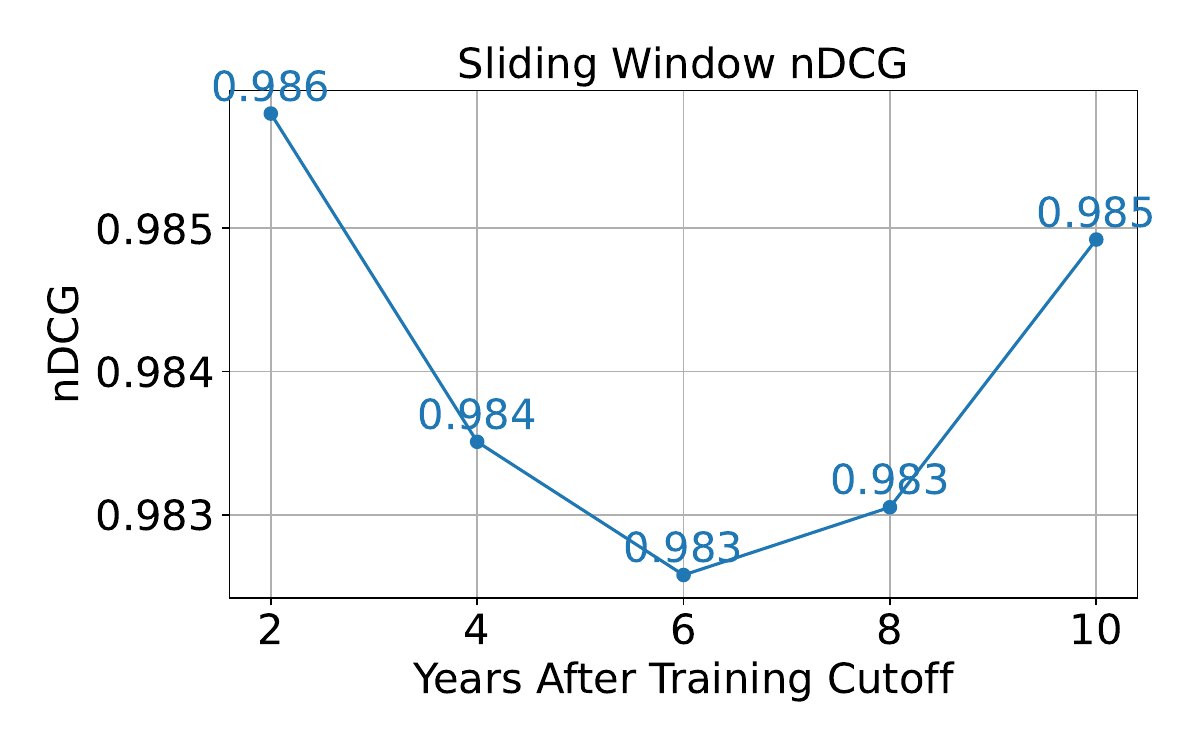}
    \caption{Mean nDCG for training cutoff 2000.}
    \label{fig:nDCG_graph_y2000}
\end{subfigure}
\hfill
\begin{subfigure}[t]{0.32\textwidth}
    \centering
    \includegraphics[width=\linewidth]{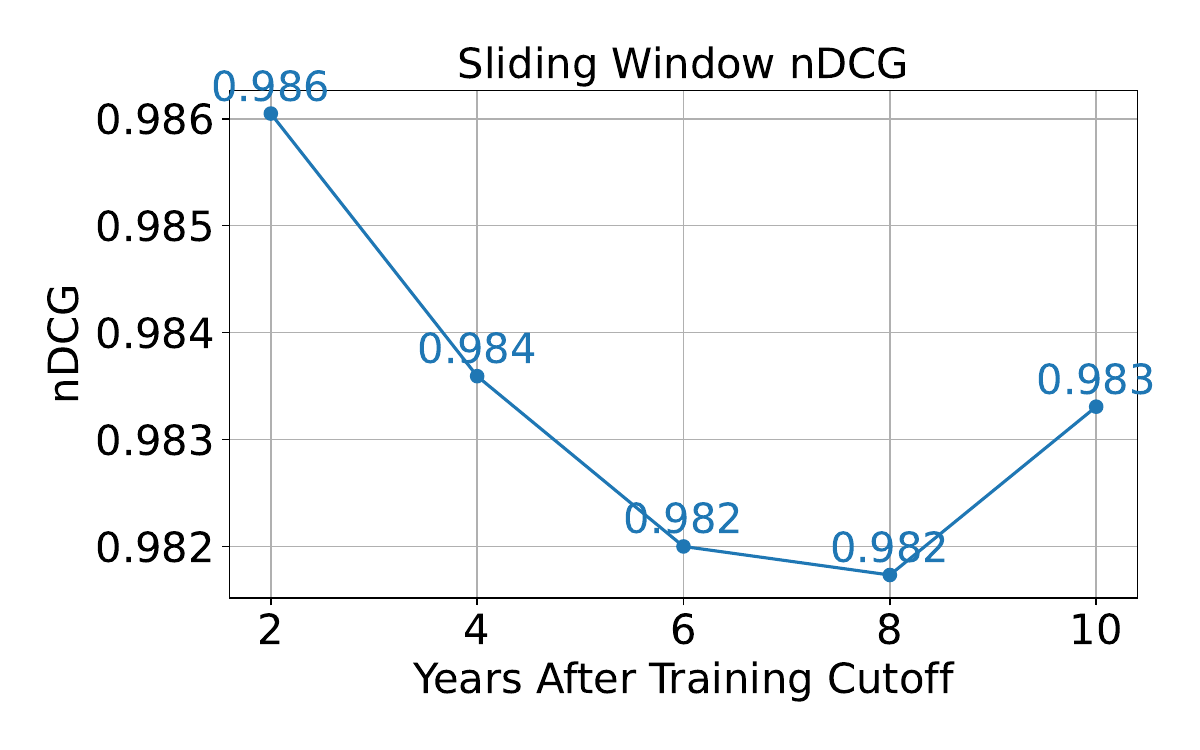}
    \caption{Mean nDCG for training cutoff 2010.}
    \label{fig:nDCG_graph_y2010}
\end{subfigure}
\caption{
Historic validation results for \llmmodel.
First row (a--c): Discovery coverage across disjoint horizons. Each curve reports micro recall for new relations that first appear in each future window.
Second row (d--e): Discovery coverage across disjoint horizons. Each curve reports macro recall for new relations that first appear in each future window.
Third row (f--g): Stability of prioritization across disjoint horizons. Mean nDCG quantifies whether relations that first appear in a given future window concentrate near the top of the ranked list.
}
\label{fig:historic_validation_graph}
\end{figure*}

In Fig.~\ref{fig:historic_validation_graph} (first line), we report the micro recall, where all relationships are treated equal, and analyze how recall evolves over time. Variation in this curve can arise from two fundamentally different sources. The first is real temporal effects, which are the primary object of interest, such as changes in the model’s ability to generalize across time, shifts in terminology, or entity drift. The second is dataset composition effects, which are artifacts caused by differences in the distribution or difficulty of relations across time periods, as well as interactions with decision thresholds.
To de-couple these effects we also report the macro recall In Fig.~\ref{fig:historic_validation_graph} (middle line). Here the recall is first computed per relation type and then averaged. The decaying trend is clear, with a drop of 8\%. This moderate decrease is consistent with the intuition that longer-term predictions are inherently more challenging, while still indicating substantial retention of predictive signal over time.

In addition, we introduce a null baseline to answer a complementary question: how well would the model perform if scores for relation $r$ came from the same overall distribution, but were randomly assigned? This baseline preserves calibration and overall score behavior while removing any input–output relationship.
We construct this baseline via random empirical sampling of prediction scores. Inputs, labels, and thresholds are kept fixed, but predicted scores are replaced with values drawn from the empirical distribution of all predictions that the model made around a certain relation type. 
Under this baseline, recall is substantially lower than the model, remaining around 0.24 in early bins and decreasing slightly in later bins. The gap between the model and the null baseline therefore reflects genuine predictive capability that persists across time, despite increasing difficulty in long-horizon prediction.

We observe that interval recall curves are nearly identical when training the model on different knowledge cut-offs (e.g., 1990, 2000, or 2010). Specifically, while the absolute set of known relations changes across these regimes, the relative pattern of performance across forecasting horizons remains highly stable: short-term predictions are easier, long-term predictions are harder, and the rate of decline from early to late bins is very similar in all cases. This suggests that the underlying structure governing predictive difficulty is largely invariant over time. In other words, even as scientific knowledge grows and the graph becomes denser, the way in which predictability degrades with increasing temporal distance remains essentially unchanged. This indicates that we are not observing regime-specific behavior tied to a particular historical period, nor a systematic shift in model difficulty across decades, but rather a stable “decay profile” of predictability that persists across different snapshots of scientific knowledge.

Finally, we analyze the stability of the model’s ranking using mean nDCG across time horizons. We observe that nDCG remains highly stable across cutoffs, horizons, and model variants. This indicates that the relative ordering of predicted hypotheses is largely preserved over time. In other words, even as recall may slightly decline, the model continues to prioritize the same relationships as most promising. This stability is critical for supporting sustained scientific exploration, where consistent prioritization matters as much as absolute accuracy.
Taken together, these temporal analyses support the practical applicability of the model. The results suggest that the scientific literature encodes persistent structural constraints that make many future discoveries predictable in advance, that the model generates hypotheses that remain coherent over time, and that it prioritizes them in a way that continues to be informative as new knowledge emerges over the following decade.

\subsection{Wet-lab Validation}
\label{subsec:wet-lab-validation}
As a gold-standard evaluation of Hakken's practical usefulness and the ability to materially contribute to the generation of new scientific knowledge, we worked with biomedical researchers and an independent Contract Research Organization (CRO) to undergo the full cycle from hypothesis generation to wet-lab validation.

As our biomedical collaborators are experts in the area of aging-related genetics, to reduce the search space and retain biological relevance, we restricted Hakken's predictions to a curated subset of entities from that space. The corresponding selection process started from a compiled list of aging-relevant genes provided by biomedical experts and continues with the generation of 1,543,297 above-confidence-threshold hypotheses. Once hypotheses had been generated, a hypothesis-selection pipeline processed the raw predictions produced by \llmmodel on the previously curated subset of nodes and yielded a refined list of mechanistically plausible hypotheses. Following completion of the pipeline, the hypotheses and \explainer's corresponding explanations were shared with the biomedical experts who were then tasked to select three predictions for experimental validation based on their subjective assessment of the hypotheses' interestingness and relevance to the research field. The following predicted biomedical relationships were then submitted to the CRO for wet-lab validation: \emph{TP53 affects the expression of BAMBI} which may link how TP53, the most commonly mutated gene in human cancer, increases cancer progression; \emph{RAF1 decreases the expression of TNF} which could provide insights into cancer-associated inflammation, immune evasion, and targeted therapies; and \emph{SOAT1 affects the transcriptional activity of STAT3} which would provide new insights into how lipid metabolism intersects with inflammation, cancer, and metabolic diseases.

This first round of experimental evaluation found confirmatory evidence for two of the three hypotheses, introducing two previously undocumented gene-gene and enzyme-gene interactions to the body of biomedical knowledge. For the suggested effect of TP53 on the expression of BAMBI, a consistent increase across initial and confirmatory assay in BAMBI expression following P53 induction---coinciding with maximal P53 protein levels---provided validating evidence for the hypothesis of a functional relationship between P53 and BAMBI. Similarly, for the hypothesized decreasing influence of RAF1 on the expression of TNF, a consistent increase across initial and confirmatory assay in TNF mRNA expression following cell treatment with a RAF1-inhibitor---coinciding with increased ERK phosphorylation indicating RAF1 dimerization and transactivation---confirmed a functional relationship between RAF1 modulation and TNF regulation. Regarding the SOAT1-STAT3 hypothesis, the results from the hypothesis testing assays showed that stimulation of SOAT1 (via LDL loading of HepG2 cells) did not demonstrate an effect on STAT3 at the transcription level or STAT3 at the protein level. These findings, thus, failed to support the hypothesis of a potential link between SOAT1 activity and STAT3 expression or activation, at least within the chosen experimental design.

The successful wet-lab validations of the relationships between TP53 and BAMBI, and between RAF1 and TNF, allow us to add two novel discoveries first suggested by Hakken to the body of biomedical knowledge. Both results have potential significant impact, particularly in the context of drug discovery and repurposing. Now that it has been confirmed that TP53 affects the expression of BAMBI, if future studies find that TP53 more specifically represses BAMBI, this could mean that P53 enhances TGF-$\beta$ tumor-suppressive effects by reducing BAMBI’s inhibitory role. Loss of TP53---which is a common event in cancer---might, thus, lead to increased BAMBI expression, suppressing TGF-$\beta$ signaling and promoting tumor progression. On the other hand, if TP53 activates BAMBI, this could suggest that P53 uses BAMBI to dampen excessive TGF-$\beta$ activity, preventing fibrosis or metastasis. As regards the confirmed regulatory relationship from RAF1 to TNF expression, this could open new therapeutic avenues for cancer, autoimmune diseases, and inflammatory disorders.

\section{Discussion}\label{sec13}

For Hakken to realize its full potential, it must be placed in the hands of researchers and integrated into their daily workflows. This requires delivering outputs in a clear, intuitive, and accessible way. To address this, we developed a Graphical User Interface (GUI) that enables researchers to: (1) create and search Hakken hypotheses using subject, object, or predicate inputs, and (2) test researcher-generated hypotheses by reviewing confidence scores produced by Hakken.
These hypotheses and confidence scores are supplemented with supporting evidence through \explainer explanations. In addition, we incorporated LLM-based contextual predictions to identify existing papers likely to contain relevant information related to a given hypothesis. The GUI also allows researchers to apply search and prediction constraints, including prioritizing recent literature (hot topics), older literature (more nascent research areas), and adjusting the preferred path length connecting subjects and objects in the knowledge graph (where longer paths may indicate less intuitive relationships).
We provided biomedical researchers from seven academic and industry organizations with access to Hakken’s GUI (for licensing reasons utilizing a version of the system trained exclusively on the PubTator3 dataset~\cite{wei2024pubtator}). The qualitative feedback was overwhelmingly positive, with users particularly highlighting the value of the explanatory evidence accompanying the hypotheses.

In general, Hakken opens up new ways to practice science. On the one hand, the new body of predicted knowledge available through Hakken is a dataset with massive utility, one that can take researchers beyond existing knowledge. On the other hand, in addition to the value of the predictions themselves, the associated confidence scores offer potentially game-changing mechanisms to researchers. As a case study, we have explored Hakken’s potential in drug development, applying its scores as a ``knowledge score'', to be used as a new means of filtering, ranking and progressing candidates (i.e., novel active substances) through the stages of discovery and development in preclinical research, alongside other mechanisms like ADMET scores~\cite{ADMET2019}. While we recognize the boundaries of a score that indicates only how strongly a hypothesis is supported by the literature, researchers in the field confirmed that the highly cost-intensive space of drug development needs all the de-risking help it can get. 

Both \llmmodel and \explainer are not bound in their functioning to any particular area of science, but have intentionally been built to be domain-general. While extensively validating Hakken in the biomedical domain, we also prototyped Hakken's application to agricultural science and to materials science among others. In agricultural science, operating on inputs extracted from 19th Century texts about market gardening, Hakken could contribute to the ideation of new opportunities for the small-scale, intensive and eco-friendly production of fruits and vegetables. 
In the case of materials science, Hakken could be used to suggest new chemical relationships.

An immediate next step to further Hakken's capabilities is to increase the expressiveness of the knowledge representation in order to overcome Hakken's current limitation to operating on concepts and their relationships only.
Improvements could be achieved by incorporating quantitative and conditional information into the knowledge graph---and, ultimately, also into the predictions---representing, for instance, quantity ratios or environmental conditions required for a relationship to hold.
Moreover, at present Hakken only considers and represents the first occurrence of a fact in the available literature.
This representation could be enriched with additional temporal and evidential metadata, such as the number of supporting papers or sentences, the most recent observation year, and the frequency of occurrence over time. 
Incorporating such signals could further improve the robustness and predictive performance of the system.
Additionally, Hakken’s method of incorporating the temporal evolution of literature points towards opportunities for deeper analysis of how science evolves relationship by relationship, and how it might evolve in the future. 
Hakken currently does not reintroduce the concept of time in its output, predicting \textit{when} a relationship might appear, and the possibility of doing this would be a valuable addition to the system's capabilities. 
Relatedly, being able to forecast the impact within the field of a predicted relationship---measured, say, by the number of follow-up discoveries it unlocks---could alter the trajectory of entire scientific disciplines.
Finally, another major expansion will be to complement Hakken’s knowledge-prediction with experimental-design generation capabilities. Feedback from working scientists suggests that, while also aiding process efficiency, receiving a suggested experimental design will further augment the trustworthiness of the suggested discovery. 

We strongly believe in a future ecosystem of AI prediction tools that wield the vast body of published scientific knowledge in a fashion similar to Hakken. Hakken is in many ways complementary to LLM-based systems such as the hypothesis-generation focused HypoGeniC~\cite{zhou2024hypothesis} and ResearchAgent~\cite{baek2025researchagent}, or to full-cycle systems such as AI Co-Scientist~\cite{gottweis2025aicoscientist} or the AI Scientist~\cite{lu2024ai}. 
LLMs often are trained on vast amounts---up to web-scale---of data of heterogeneous quality. Hakken uses a comparatively small amount of well-curated---literally peer-reviewed---data. LLMs often require almost prohibitively large amounts of compute for training, with a run taking weeks or months. \llmmodel, in the setting described in this article, has been trained on a modest cluster comprised of two machines each equipped with 8 NVIDIA H100 80GB SXM5 GPUs and 960GB memory, with the duration of a run measured in days. Finally, larger LLMs still require high-end hardware for inference, whilst Hakken runs on commodity hardware.
While this makes Hakken by itself the by far more accessible approach as regards resources, we are convinced that major added value can be obtained when both paradigms are combined. AI Co-Scientist, for instance, employs a multi-agent architecture built upon Google's Gemini 2.0 model with the intent to reproduce the scientific reasoning process. Scientists can input a research goal in natural language for the system's LLM agents to then summarize and synthesize prior literature and try to propose new hypotheses and protocols for validations. Without adding significant overhead in required resources, Hakken's predicted knowledge and corresponding explanations can serve as additional input data, guiding LLM reasoning with structured, interrogable information and expanding the reach of the system's reasoning through predicted knowledge outside of the deductive closure of the input data. At the same time, Hakken's confidence scores can provide a quantitative assessment of the degree to which existing knowledge supports LLM outputs, laying the foundations to counteract imprecision and hallucinations~\cite{xiong2025reliablescientifichypothesisgeneration}. 

\backmatter

\bmhead{Acknowledgements}

We want to thank Priyadarshini Kumari, Angel Hsing-Chi Hwang, Cliona O'Doherty, Jesus de la Fuente, Alessandro Lonardi, Laura Cabello, Carmen Martin Turrero, Donghee Choi, Anton Kratz, Ayako Yachie, Ernesto Jimenez-Ruiz, and Catia Pesquita for their input and support during different stages of Hakken's development. We further want to thank Pharmagene Discovery Services Ltd. for the positive collaboration during the wet-lab validation of hypotheses.

%
%
%
%
%
%
%
%
%
%
%
%
%
%
%
%
%
%


\begin{appendices}

\section{Open-Source Resources}~\label{appendix_open-source}
An open-source version of the Hakken infrastructure, including a prediction model trained exclusively on the NCBI's open-access PubTator3 dataset \cite{wei2024pubtator} (similar to the variant used in the context of the qualitative user studies mentioned in Section~\ref{sec13} above), is available from \url{https://github.com/SonyResearch/HakkenOSS}. 

\section{Temporal Hierarchical Graph-based Encoder Representation with LLM (\llmmodel)}
The following gives a detailed description of \llmmodel (Temporal Hierarchical Graph-based Encoder Representation with LLM).

\subsection{Problem Setup}
\label{sec:setup}
\subsubsection{Supervision as temporally indexed positive--unlabeled targets}
Each training instance is represented as
\begin{equation}
\instance{s}{o}
\end{equation}
where $\ctx_{so}$ is the text context and $\timevec_{so}\in\{0,\dots,T_{\max}\}^{C}$ stores the observed discovery time of each label. 
Here, $t_{so,c}=0$ indicates that label $c$ has not been observed for the pair, whereas $t_{so,c}>0$ indicates that label $c$ is known and was first observed at time $t_{so,c}$.
From this representation we derive the binary supervision signal

\begin{equation}
y_{so,c}=\mathbbm{1}[t_{so,c}>0].
\end{equation}

This formulation corresponds to a positive--unlabeled (PU) learning setting: observed labels constitute confirmed positives, while unobserved labels are treated as unlabeled rather than explicit negatives.

For a query time $\tau$, we additionally construct a \emph{label-history snapshot}
\begin{equation}
m^{(\tau)}_{so,c}=\mathbbm{1}[t_{so,c}>0 \;\wedge\; t_{so,c}\le \tau],
\end{equation}
which indicates which labels are already known at that time.
which indicates whether label $c$ has already been observed for the pair $(s,o)$ by time $\tau$.
During training, $\tau$ is sampled per example; during evaluation, it is fixed according to the test protocol.

\subsubsection{Model outputs}
The model produces two sets of label scores:
\begin{equation}
\ltext{so}\in\mathbb{R}^{C},
\qquad
\lgraph{so}\in\mathbb{R}^{C},
\end{equation}
corresponding to the language branch and the graph branch, respectively. 
Both branches are trained jointly.
At inference time, they are combined by a label-wise ensemble and  subsequently calibrated to obtain probability estimates.

\subsubsection{Text tokens and graph tokens}
Let the LLM hidden size be $d$ and the temporal-GNN hidden size be $d_g$.
A prompt is tokenized into a length-$L$ token ID sequence $\prompt=(\tok[1],\dots,\tok[L])$ that contains \emph{exactly one} special placeholder token \texttt{<htg>} at index $k$, where $k\in\{1,\dots,L\}$.
At runtime, this placeholder is replaced by a global node-pair graph token $G_{so}$. The temporal graph token $T_m \in \mathbb{R}^d$ is then prepended to the text token, where $T_m$ encodes information from the $m$-th temporal graph snapshot
As a result, the sequence length increases to

\begin{equation}
L' = L + M.
\end{equation}

\subsection{Model Overview}
\label{sec:overview}

Given a pair $(s,o)$ and its optional textual context $\ctx_{so}$, the temporal graph encoder constructs a sequence of $T$ pair embeddings $\{\pairembt{t}{so}\}_{t=1}^T$ using GraphSAGE-style neighborhood aggregation over $\mathcal{G}^{(t)}$ \cite{hamilton2017graphsage}. A hierarchical temporal Transformer then aggregates these $T$ embeddings into (i) a global graph vector and (ii) a time-preserving latent sequence. A lightweight \emph{temporal token mapper} downsamples the latent temporal sequence into $M$ LLM-dimensional tokens, which is then passed as token in the LLM input. The LLM produces a pooled representation for multi-label prediction, trained with a PU-safe dynamic-weight focal loss (\S\ref{sec:loss}).

\subsection{Temporal Graph Encoder}
\label{sec:tgnn}

\subsubsection{Node Features with Ontology and Domain Signals}
\label{sec:nodefeat}

Each node $i\in\nodeset$ is associated with: (a) a trainable embedding $\mathbf{e}_i\in\mathbb{R}^{d_e}$, (b) an ontology-derived auxiliary vector $\mathbf{o}_i\in\mathbb{R}^{d_o}$ (e.g., concept ontological attributes), and (c) an optional domain assignment $g(i)\in\{1,\dots,D\}$ with embedding $\mathbf{d}_{g(i)}\in\mathbb{R}^{d_d}$. We form an initial feature
\begin{equation}
\mathbf{h}^{(0)}_{i} \;=\; \mathbf{W}_{\text{comb}}
\Big[\mathbf{e}_i \,;\, \mathbf{o}_i \,;\, \mathbf{d}_{g(i)}\Big]
\in\mathbb{R}^{d_0},
\label{eq:node_init}
\end{equation}
where $\mathbf{W}_{\text{comb}} \in \R^{d_0 \times (d_e + d_o + d_d)}$ is a learned linear projection and $[\,\cdot\,;\,\cdot\,]$ denotes concatenation.

\subsubsection{Ontology-based initialization of embeddings}
\def\rdftovec{\textsc{RDF2vec}\xspace}
\def\wordtovec{\textsc{word2vec}\xspace}

We enrich the knowledge graph representation using ontological information about the entities.
The ontology is hierarchical, with child terms being more specialized than their parent terms. 
An example of a hierarchical path would be disease $\rightarrow$ infectious disease $\rightarrow$ viral infectious disease $\rightarrow$ respiratory viral disease $\rightarrow$ COVID-19.
Some concepts are entities from the knowledge graph (the ones we are interested in building representations for), while others are more general domain terms.
Unlike a strict hierarchy, a term may have more than one parent term.
This type of information is not unusual, with another common example being Gene Ontology~\cite{Ashburner2002GO}.

To build these representations, we treat the ontology as a relational graph and run the \rdftovec algorithm~\cite{Ristoski2016RDF2Vec}.
\rdftovec uses random walks on the relational graph to create sequences of RDF nodes, which are then used as input for the \wordtovec algorithm~\cite{Mikolov2013EfficientEO}.
The inductive bias of the \wordtovec algorithm, which assigns similar vector representations to words appearing in similar local contexts, is particularly appropriate for building hierarchical ontological representations. 
If concepts sit near in the hierarchy, we indeed expect them to be similar.
These representations are used as initial embeddings for the entities in the KG. 
More precisely, the initial representation of each entity consists of two components: one that is randomly initialized and optimized during the model training, and one derived from the ontological embeddings described above, which injects structured domain knowledge into the representations.
%



\subsubsection{Neighborhood Sampling Across Time}
\label{sec:sampling}

For scalability on large graphs, we compute embeddings via \emph{sampled} message passing. Let $\mathcal{N}^{(t)}_k(i)$ denote the sampled set of neighbors at hop $k$ for node $i$ within snapshot $\mathcal{G}^{(t)}$, with fixed fanout $S_k = |\mathcal{N}^{(t)}_k(i)|$. Sampling is repeated per batch and can be reshuffled each iteration to reduce bias. When available, we also sample from a negative adjacency (for ranking regularization; \S\ref{sec:rank_loss}).

\subsubsection{GraphSAGE Message Passing Per Snapshot}
\label{sec:graphsage}

For each snapshot $t$, we run $K$ layers of GraphSAGE-style propagation \cite{hamilton2017graphsage}. Let $\mathbf{h}^{(k,t)}_i\in\mathbb{R}^{d_k}$ be the node embedding of node $i$ at layer $k$ for time $t$ (with $\mathbf{h}^{(0,t)}_i=\mathbf{h}^{(0)}_i$). At layer $k\in\{1,\dots,K\}$ we compute:
\begin{align}
\mathbf{m}^{(k,t)}_i
&=
\mathrm{AGG}^{(k)}\left(\left\{\mathbf{h}^{(k-1,t)}_j:\; j\in\mathcal{N}^{(t)}_k(i)\right\}\right)\in\mathbb{R}^{d_{k-1}},
\label{eq:agg}\\
\mathbf{h}^{(k,t)}_i
&=
\phi\!\left(
\mathbf{W}^{(k)}
\left[\mathbf{h}^{(k-1,t)}_i\; ;\;\mathbf{m}^{(k,t)}_i\right]
+\mathbf{b}^{(k)}
\right)\in\mathbb{R}^{d_k},
\label{eq:sage_update}
\end{align}
where:
\begin{itemize}
\item $\mathrm{AGG}^{(k)}$ is a permutation-invariant aggregator (e.g., mean), applied over the sampled multiset of neighbor embeddings;
\item $\mathbf{W}^{(k)}\in\mathbb{R}^{d_k\times(2d_{k-1})}$ and $\mathbf{b}^{(k)}\in\mathbb{R}^{d_k}$ are learnable;
\item $\phi(\cdot)$ is a nonlinearity (e.g., SiLU/GeLU).
\end{itemize}
If edge-label features are available, the aggregator can be extended to incorporate edge label embeddings; the methodology here remains compatible by replacing the multiset in \eqref{eq:agg} with tuples $(\mathbf{h}_j,\mathbf{r}_{ij})$.

\subsubsection{Time-Resolved Pair Embeddings}
\label{sec:pair_emb}

Given the final node embeddings for pair $(s,o)$ at time $t$, the graph encoder constructs a time-specific pair representation
\begin{align}
\tilde{\mathbf{p}}^{(t)}_{so}
&=
\mathrm{Norm}\left(\mathbf{h}^{(K,t)}_s + \mathbf{h}^{(K,t)}_o\right), \\
\pairembt{t}{so}
&=
\mathbf{W}_{\mathrm{pair}}\tilde{\mathbf{p}}^{(t)}_{so}+\mathbf{b}_{\mathrm{pair}}
\in\mathbb{R}^{d_g}.
\label{eq:pair_proj}
\end{align}
The sequence
\begin{equation}
\mathbf{P}_{so}
=
\left[\mathbf{p}^{(1)}_{so},\dots,\mathbf{p}^{(T)}_{so}\right]\in\mathbb{R}^{T\times d_g}
\end{equation}
is then passed to the temporal encoder. The graph model returns a positive pair embedding $\mathbf{z}^{+}_{so}$, an optional negative-pair embedding $\mathbf{z}^{-}_{s\tilde o}$ for link ranking, a temporal latent sequence $\mathbf{Z}^{\mathrm{temp}}_{so}  = [z^+_1, ... , z^+_m]$, and a learned label-embedding matrix $\mathbf{E}^{\mathrm{label}}\in\mathbb{R}^{C\times d_g}$.

\subsubsection{Hierarchical Temporal Transformer}
\label{sec:htrans}

We model inter-window dependencies using a Transformer \cite{vaswani2017attention} applied on the temporal axis, with a \emph{hierarchical reduction} that halves sequence length after each stage.
Let $\mathbf{H}^{(0)}=\mathbf{P}_{so}\in\mathbb{R}^{T\times d_g}$.
At stage $l$, the current length is $L_l$ (initially $L_0=T$). We apply $R$ Transformer layers:
\begin{equation}
\mathbf{H}^{(l,r+1)} = \mathrm{Tx}^{(l,r)}\!\left(\mathbf{H}^{(l,r)};\mathbf{B}^{(l)}\right),
\qquad r=0,\dots,R-1,
\end{equation}
where $\mathbf{B}^{(l)}\in\mathbb{R}^{L_l\times L_l}$ is an additive attention-bias matrix (defined below). Then we merge adjacent positions:
\begin{equation}
\mathbf{H}^{(l+1,0)}_k = \mathbf{H}^{(l,R)}_{2k-1} + \mathbf{H}^{(l,R)}_{2k},
\qquad k=1,\dots,\left\lfloor\frac{L_l}{2}\right\rfloor,
\label{eq:merge}
\end{equation}
padding the last element if $L_l$ is odd. After $L=\lceil\log_2 T\rceil$ stages, we obtain a single global graph vector:
\begin{equation}
\mathbf{z}^{\mathrm{graph}}_{so} = \mathbf{H}^{(L,0)}_1\in\mathbb{R}^{d_g}.
\end{equation}

\textbf{Pair-block attention bias}
A known failure mode of hierarchical ``attend-then-merge'' temporal Transformers is a \emph{degenerate pre-merge shortcut}: because the next operation deterministically sums the fixed pair $(2k\!-\!1,2k)$, the self-attention layers immediately before merging can minimize loss by concentrating attention mass within each soon-to-be-merged pair (i.e., token $(2k\!-\!1)$ attending primarily to $(2k)$ and vice versa), effectively behaving like a local two-token mixing operator. In that regime, the model under-utilizes cross-time interactions \emph{across} merge groups, and the subsequent summation tends to collapse highly correlated representations, reducing the benefit of having a global self-attention stage.
To explicitly prevent this shortcut and enforce \emph{cross-group} temporal information flow prior to length reduction, we apply an additive attention bias that suppresses attention \emph{only} between the two tokens that will be merged together, while leaving self-attention intact.
Formally, for a temporal sequence of length $L_l$ at hierarchical stage $l$, we define $\mathbf{B}^{(l)}\in\mathbb{R}^{L_l\times L_l}$ as
\begin{equation}
\mathbf{B}^{(l)}_{ij}=
\begin{cases}
-\beta, &
\left\lfloor\frac{i-1}{2}\right\rfloor = \left\lfloor\frac{j-1}{2}\right\rfloor \;\wedge\; i\neq j,\\
0, & \text{otherwise},
\end{cases}
\qquad \beta\gg 1,
\label{eq:block_bias}
\end{equation}
where $i$ indexes the query position and $j$ indexes the key position (1-indexed). This yields a $2\times2$ within-block pattern
$\begin{bmatrix}0 & -\beta\\ -\beta & 0\end{bmatrix}$:
cross-attention between the pairmates is discouraged, so each token must draw contextual evidence from \emph{other} time steps outside its merge pair before the deterministic summation is applied. In practice, $\beta$ is implemented as a large negative constant approximating $-\infty$, making the forbidden within-pair cross-attention weights negligible after softmax.

\begin{figure}
    \centering
    \includegraphics[width=1\linewidth]{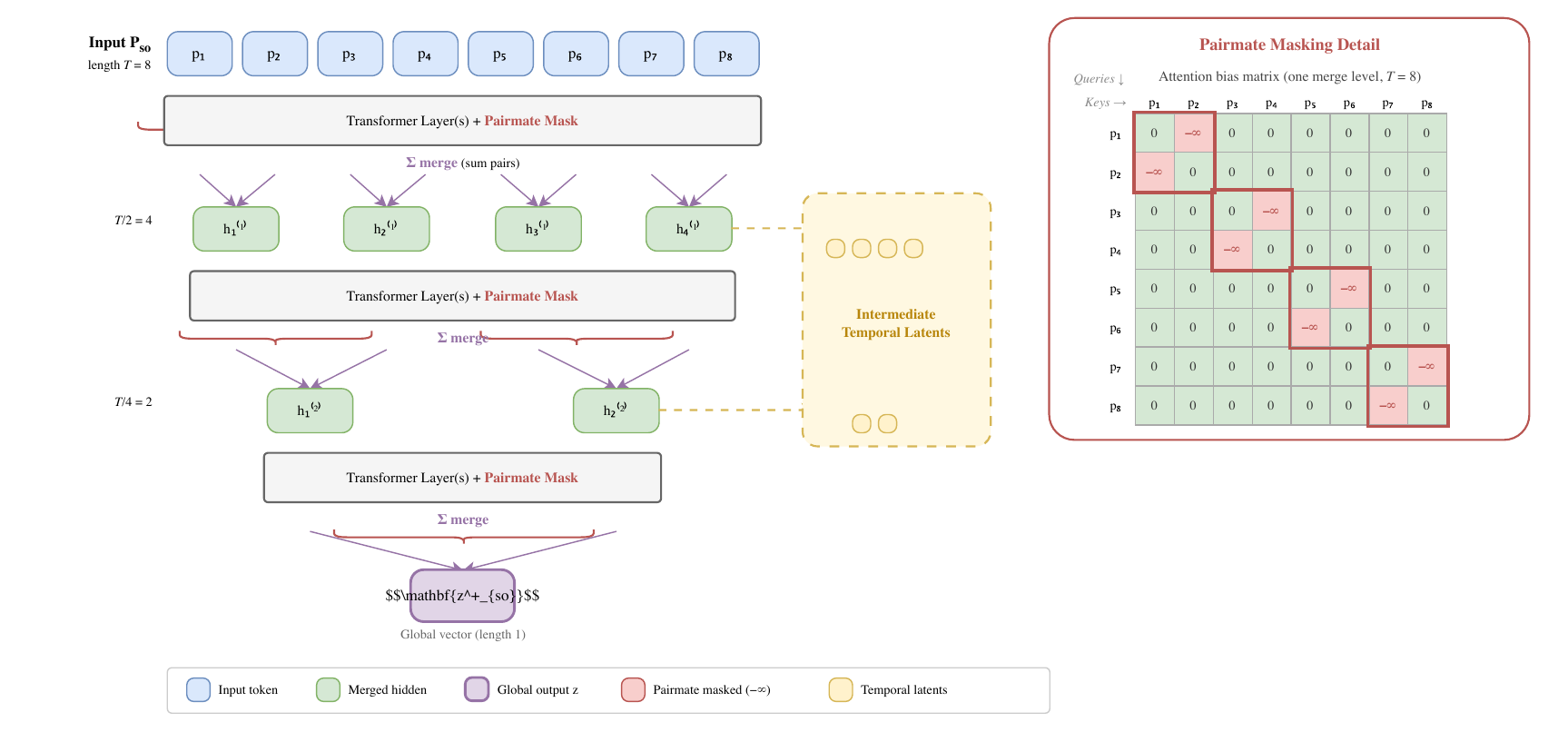}
\caption{Temporal modeling via hierarchical Transformer with pairwise merging and pairmate-only attention masking.}
\label{fig:htrans}
\end{figure}

\subsection{Graph-to-Token Mapping and LLM Injection}
\label{sec:fusion}

\subsubsection{Temporal Token Mapper}
\label{sec:mapper}

The LLM consumes embeddings in $\mathbb{R}^d$, while the temporal graph sequence lies in $\mathbb{R}^{d_g}$. Therefore, we map the temporal latent sequence into a short prefix of graph tokens using a dedicated temporal mapper.
\begin{equation}
\mathbf{Z}^{\mathrm{temp}}_{so}\in\mathbb{R}^{N\times d_g}
\end{equation}
be the temporal output of the graph encoder. The mapper first applies a reduction operator
\begin{equation}
\mathbf{R}_{so} = \mathrm{Reduce}_{M_{\max}}(\mathbf{Z}^{\mathrm{temp}}_{so}) \in \mathbb{R}^{M'\times d_g},
\qquad M' \le M_{\max},
\label{eq:reduce}
\end{equation}
where $N$ is the total number of time steps, $M'=N$ if $N\le M_{\max}$, and otherwise the sequence is downsampled to length $M_{\max}$ using adaptive average pooling or linear interpolation. Specifically, we do  \emph{not} upsample short sequences to a fixed length; short temporal sequences are kept as they are.


\subsubsection{Fourier time temporal encoding and gating}
To preserve temporal order, we add a Fourier time encoding to the temporal embedding mapper
\begin{equation}
\psi(t_m)=
\mathbf{W}_t\left[
\sin(2\pi f_1 t_m),\dots,\sin(2\pi f_F t_m);\;
\cos(2\pi f_1 t_m),\dots,\cos(2\pi f_F t_m)
\right]\in\mathbb{R}^{d},
\label{eq:fourier}
\end{equation}
with $t_m\in[0,1]$ defined on a uniform grid over the actual prefix length $M'$. The final graph token is
\begin{equation}
\mathbf{g}_m
=
\sigma(\gamma)\cdot
\mathrm{RMSNorm}\!\left(\phi(\mathbf{r}_m)+\psi(t_m)\right),
\label{eq:gate}
\end{equation}
where $\gamma$ is a learned scalar gate shared across the temporal prefix. This gate allows the model to control how strongly graph information enters the LLM.

\subsubsection{Global pair projector}
In addition to the temporal prefix, the global graph representation is projected into a single LLM-space conditioning vector
\begin{equation}
\mathbf{g}^{\mathrm{pair}}_{so}
=
\mathrm{RMSNorm}\!\left(\mathbf{W}_{\mathrm{gp}}\,\mathbf{z}^{+}_{so}\right)\in\mathbb{R}^{d}.
\end{equation}
This vector provides a compact pair-level summary that is separate from the temporal prefix.

\subsubsection{Embedding-Level Splicing at the \texttt{<htg>} Placeholder}
\label{sec:splice}

Let $\prompt=(\tok[1],\dots,\tok[L])$ denote a tokenized prompt of length $L$ containing exactly one \texttt{<htg>} token at position $k$, i.e., $\tok[k] = \texttt{<htg>}$.
Let $\mathrm{Emb}(\cdot):\text{Vocab}\rightarrow\mathbb{R}^d$ be the LLM embedding table. The text embedding sequence is:
\begin{equation}
\mathbf{E}=
\left[\mathrm{Emb}(\tok[1]),\dots,\mathrm{Emb}(\tok[k-1]),
\mathrm{Emb}(\tok[k]),
\mathrm{Emb}(\tok[k+1]),\dots,\mathrm{Emb}(\tok[L])\right]\in\mathbb{R}^{L\times d}.
\end{equation}
We remove the placeholder embedding $\mathrm{Emb}(\tok[k])$ and splice in the global node-pair graph embedding $\mathbf{G}_{so}$ and then prepend the temporal node-pair embeddings $[\mathbf{g}_1;\dots;\mathbf{g}_M]\in\mathbb{R}^{M\times d}$:
\begin{equation}
\mathbf{E}'=
\left[
\mathbf{g}_1,\dots,\mathbf{g}_M, \mathrm{Emb}(\tok[1]),\dots,\mathrm{Emb}(\tok[k-1]), \mathbf{G}_{so}, 
\mathrm{Emb}(\tok[k+1]),\dots,\mathrm{Emb}(\tok[L])
\right]\in\mathbb{R}^{(L+M)\times d}.
\label{eq:splice_main}
\end{equation}
Thus, the LLM processes a sequence of length $L'=L+M$. Graph tokens are always treated as valid tokens and are assigned indices as if the insertion had not occurred (up to an offset). This preserves the distribution of relative positions among text tokens.

\begin{figure}
    \centering
    \includegraphics[width=1\linewidth]{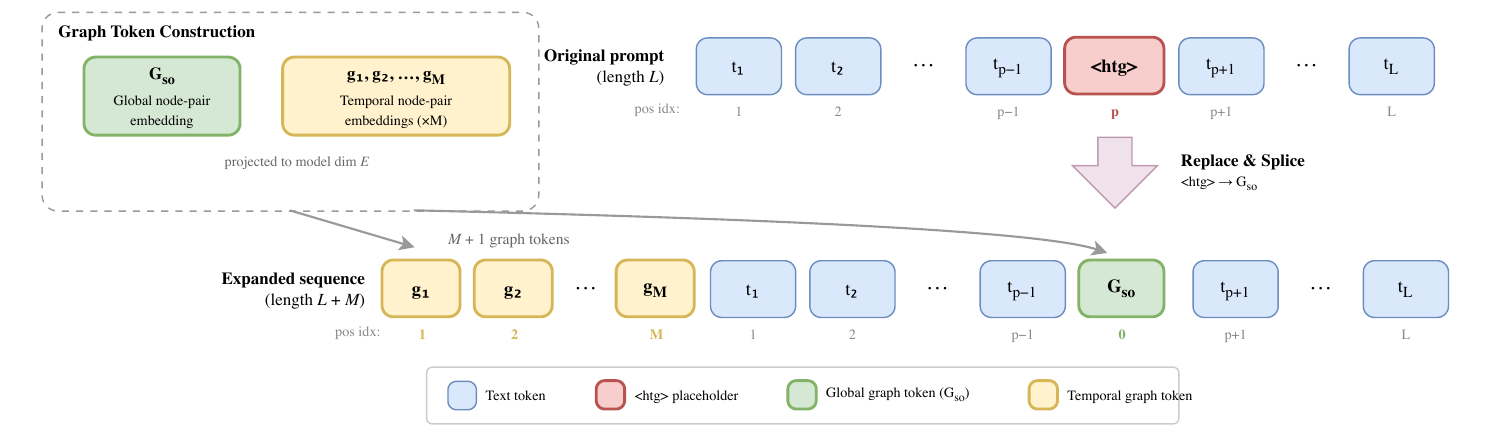}
\caption{Token-based injection by replacing \texttt{<htg>} with global node-pair embedding and prepending $M$ temporal graph tokens}
\label{fig:inject}
\end{figure}

\subsection{LLM Encoder and Multi-label Prediction}
\label{sec:llm_pred}

Let $\mathrm{LLM}_\theta$ be a decoder-only Transformer with parameters $\theta$. Given the prefixed embedding sequence $\mathbf{E}'\in\mathbb{R}^{L'\times d}$ together with the graph-pair conditioning vector $\mathbf{g}^{\mathrm{pair}}_{so}$, the language model produces hidden states
\begin{equation}
\mathbf{H}=\mathrm{LLM}_\theta(\mathbf{E}',\mathbf{a}',\boldsymbol{\pi};\mathbf{g}^{\mathrm{pair}}_{so})
\in\mathbb{R}^{L'\times d}.
\label{eq:rope_reset}
\end{equation}

For pooling, the current implementation uses entity-aware pooling whenever explicit entity markers are available. Let $e_1$ and $e_2$ denote the token positions of the two entity markers corresponding to $s$ and $o$ in the original text sequence, respectfully. After prefix insertion, the pooled text representation is 
\begin{equation}
\mathbf{h}^{\mathrm{text}}_{so}
=
\frac{1}{2}\left(
\mathbf{H}_{e_1+M'}+\mathbf{H}_{e_2+M'}
\right).
\label{eq:pool}
\end{equation}
If entity markers are unavailable, the model falls back to the last non-padding token.

\textbf{Text logits.}
A bias-free linear head first produces base text logits,
\begin{equation}
\lbase{so}
=
\mathbf{W}_{\mathrm{cls}}\,\mathbf{h}^{\mathrm{text}}_{so}
\in\mathbb{R}^{C}.
\label{eq:logits}
\end{equation}

\textbf{History-aware label correction.}
The current implementation augments these base text logits with a history-aware residual term derived from known label history. Let $\mathbf{e}_c\in\mathbb{R}^{d_g}$ denote the embedding of label $c$, and let $\mathbf{h}^{\mathrm{hist}}_{so,c}$ be the history context obtained by attending from label $c$ to the set of labels already known at query time $\tau$. The refined text logit is
\begin{equation}
\ltexte{so,c}
=
\lbasee{so,c}
+
\alpha_c\,
\Big\langle
\mathbf{W}_{\mathrm{emb}}\mathbf{e}_c
+
\mathbf{W}_{\mathrm{ctx}}\mathbf{h}^{\mathrm{hist}}_{so,c},
\;
\mathbf{W}_{\mathrm{text}}\mathbf{h}^{\mathrm{text}}_{so}
\Big\rangle,
\end{equation}
where $\alpha_c=\sigma(a_c)$ is a learned per-label merge coefficient.

\textbf{Graph logits.}
The graph branch also predicts labels directly from the global graph embedding. Using the same history context, the graph logit for label $c$ is
\begin{equation}
\lgraphe{so,c}
=
\Big\langle
\mathbf{W}_{z}\mathbf{z}^{+}_{so}
+
\mathbf{W}_{h}\mathbf{h}^{\mathrm{hist}}_{so,c}
+
\mathbf{b},
\;
\mathbf{e}_c
\Big\rangle.
\end{equation}
The model therefore outputs two complementary score vectors, $\ltext{so}$ and $\lgraph{so}$, which are supervised jointly during training.

\textbf{Inference-time ensemble and blended calibration.}
At inference time, the two branches are combined by a label-wise ensemble
\begin{equation}
\lense{so,c}
=
\alpha^{\mathrm{ens}}_{c}\,\lgraphe{so,c}
+
\left(1-\alpha^{\mathrm{ens}}_{c}\right)\ltexte{so,c},
\end{equation}
where $\alpha^{\mathrm{ens}}_{c}\in[0,1]$ is tuned per label on validation data by cross-validation. These ensemble logits are then calibrated into probabilities. In the blended calibration setting,
\begin{equation}
\hat{p}_{so,c}
=
(1-\beta_{\mathrm{blend}})\,
\mathrm{Platt}_{c}\!\left(\lense{so,c}\right)
+
\beta_{\mathrm{blend}}\,
\mathrm{Iso}_{c}\!\left(\lense{so,c}\right),
\end{equation}
where $\mathrm{Platt}_{c}$ is a per-label logistic calibrator, $\mathrm{Iso}_{c}$ is a per-label isotonic calibrator, and $\beta_{\mathrm{blend}}\in[0,1]$ is chosen on validation data. Final binary predictions are obtained using label-specific thresholds selected on the validation set, typically with a precision--recall criterion.

\subsection{PU-safe Dynamic-Weighted Focal BCE with Regularization}
\label{sec:loss}

We optimize the LLM logits $\labscore_{so}$ with a weighted loss we call Dynamically Weighted Balanced Focal loss (DWF) drawing inspiration from the Dynamically Weighted Balanced loss \cite{dynamicloss} and Focal loss \cite{lin2017focal}. This loss combines:
(i) focal reweighting \cite{lin2017focal},
(ii) dynamic class-balancing weights,
(iii) PU-safe downweighting of unlabeled negatives,
and (iv) an explicit confidence regularizer.



\subsubsection{Elementwise Definitions}
\label{sec:loss_defs}

For one entity pair example $(s,o)$ and one label $c$, let
\begin{itemize}
\item $y_{so,c}=\mathbbm{1}[t_{so,c}>0]\in\{0,1\}$ be the binary target derived from the temporal label annotation;
\item $\ltexte{so,c}$ and $\lgraphe{so,c}$ be the text and graph logits;
\item $\tilde y_{so,c}\in[0,1]$ be the \emph{soft target} used in training.
\end{itemize}
Soft targets are obtained by annealing the unlabeled entries:
\begin{equation}
\tilde y_{so,c}
=
y_{so,c}
+
(1-y_{so,c})\,\eta_c(i),
\end{equation}
where $i$ is the current training step and
\begin{equation}
\eta_c(i)
=
\mathrm{clip}\!\left(
\hat q_c\,\eta_{\max}\,r(i),\;
0,\;
\eta_{\mathrm{cap}}\,r(i)
\right).
\end{equation}
Here, $\hat q_c$ is a label-specific prior estimate, $r(i)$ is a warm-up schedule, and $\eta_{\max}$ and $\eta_{\mathrm{cap}}$ are user-defined constants. This construction makes the supervision less brittle in the positive--unlabeled regime.

Define the probability of the correct class:
\begin{equation}
p_t = y\,p + (1-y)(1-p).
\label{eq:pt}
\end{equation}

\subsubsection{Weighted BCE Terms (with PU-safe unlabeled negatives)}
\label{sec:bce}

We write BCE in separated positive/negative terms:
\begin{align}
\mathrm{BCE}^+(z,y) &= -y\,\log(p+\varepsilon), \label{eq:bce_pos}\\
\mathrm{BCE}^-(z,y) &= -(1-y)\,\log(1-p+\varepsilon). \label{eq:bce_neg}
\end{align}
In PU learning, the $y=0$ entries are unlabeled; treating them as negatives is biased. We mitigate this by scaling the negative term:
\begin{equation}
\widetilde{\mathrm{BCE}}(z,y)=
w^+\,\mathrm{BCE}^+(z,y)
+
w^-\,\lambda_u\,\mathrm{BCE}^-(z,y).
\label{eq:pu_bce}
\end{equation}
Thus unlabeled negatives contribute less to gradients than confirmed positives.

\subsubsection{Focal and Dynamic Class-Balancing Factors}
\label{sec:focal_dyn}

\textbf{Focal factor.}
The focal factor downweights easy examples ($p_t\approx 1$) \cite{lin2017focal}:
\begin{equation}
\mathrm{Focal}(p_t)=(1-p_t)^\gamma.
\label{eq:focal}
\end{equation}

\textbf{Dynamic weight.}
We define a dynamic multiplicative factor that depends on $p_t$:
\begin{equation}
\mathrm{Dyn}(p_t;c_w)=c_w^{\,1-p_t}.
\label{eq:dyn}
\end{equation}
Interpretation: if the example is hard ($p_t$ small), the effective weight approaches $c_w$; if it is easy ($p_t$ near 1), it approaches 1. This provides a smooth, probability-dependent form of class balancing.

\subsubsection{Confidence Regularization}
\label{sec:reg}

We include a bounded regularizer based on Bernoulli variance:
\begin{equation}
\mathcal{R}(p)=p(1-p)\in\left[0,\frac14\right].
\label{eq:var}
\end{equation}
The implemented loss subtracts $\lambda_{\mathrm{reg}}\mathcal{R}(p)$:
\begin{equation}
-\lambda_{\mathrm{reg}}\,\mathcal{R}(p)\ge -\frac{\lambda_{\mathrm{reg}}}{4},
\end{equation}
so it cannot make the objective unbounded below. This term encourages confident predictions (away from $p=0.5$) once the focal/dynamic weighting drives learning on informative examples.

\subsubsection{DWF: Per-entry and Batch-normalized Form}
\label{sec:dwf_final}

\textbf{Per-entry loss.}
Combining Eq.~\ref{eq:pu_bce}, Eq.~\ref{eq:focal}, Eq.~\ref{eq:dyn}, and Eq.~\ref{eq:var}, the per-entry loss is:
\begin{equation}
\mathcal{L}_{\mathrm{DWF}}(z,y)
=
\mathrm{Dyn}(p_t;c_w)\cdot \mathrm{Focal}(p_t)\cdot \widetilde{\mathrm{BCE}}(z,y)
-\lambda_{\mathrm{reg}}\,p(1-p).
\label{eq:dwf}
\end{equation}

\textbf{Effective-mass normalization.}
Multi-label PU data exhibit highly variable positive density across batches. To stabilize the overall scale, we normalize by an ``effective mass'':
\begin{equation}
\mathrm{\gamma}
=
\sum \mathrm{Dyn}(p_t;c_w)\cdot\Big(w^+y + w^-(1-y)\lambda_u\Big)
+\varepsilon,
\label{eq:mass}
\end{equation}
where the sum is over all examples in the batch and all labels (broadcasting as needed). The final batch loss is:
\begin{equation}
\mathcal{L}_{\text{batch}}
=
\frac{\sum \mathcal{L}_{\mathrm{DWF}}(z,y)}{\mathrm{\gamma}}.
\label{eq:dwf_norm}
\end{equation}
This makes optimization less sensitive to label sparsity and the magnitude of $c_w,w^+,w^-$.

\subsection{Optional Auxiliary Graph Ranking Regularizer}
\label{sec:rank_loss}

When negative pairs are available, we add a pairwise ranking loss that encourages the model to score the observed pair higher than a corrupted pair. Let $\mathbf{z}^{\mathrm{graph}}_{so}$ be the global graph embedding for the positive pair and $\mathbf{z}^{\mathrm{graph}}_{s\tilde o}$ for a negative pair $(s,\tilde o)$ created by negative sampling. Define scalar scores:
\begin{equation}
a^+=\mathbf{w}_r^\top \mathbf{z}^{\mathrm{graph}}_{so},
\qquad
a^-=\mathbf{w}_r^\top \mathbf{z}^{\mathrm{graph}}_{s\tilde p},
\end{equation}
with trainable $\mathbf{w}_r\in\mathbb{R}^{d_g}$. The ranking loss is:
\begin{equation}
\mathcal{L}_{\mathrm{rank}}=\log\!\left(1+\exp\!\big(-(a^+-a^-)\big)\right).
\label{eq:rank}
\end{equation}
This term complements the PU classification objective by explicitly separating true pairs from sampled negatives in graph embedding space.

\subsection{Overall Optimization Objective}
\label{sec:objective}

Let $\Theta$ denote all trainable parameters (graph encoder, temporal Transformer, token mapper, and LLM classifier a subset of LLM parameters). For a batch $\mathcal{B}$, we minimize:
\begin{equation}
\min_{\Theta}\;\;
\mathbb{E}_{\supinstance{s}{o}\sim\mathcal{B}}
\left[
\mathcal{L}_{\text{batch}}(\labscore_{so},\mathbf{y}_{so})
+\lambda_{\mathrm{rank}}\mathcal{L}_{\mathrm{rank}}
\right],
\label{eq:overall_obj}
\end{equation}
where $\lambda_{\mathrm{rank}}\ge 0$ controls the auxiliary ranking term.


\begin{algorithm}[t]
\caption{Forward pass with token-based temporal graph injection}
\label{alg:forward}
\begin{algorithmic}[1]
\Require Pair $(s, o)$, text context $\ctx_{so}$, snapshots $\{\mathcal{G}^{(t)}\}_{t=1}^T$, placeholder index $k$, token budget $M$.
\State \textbf{(1) Temporal GNN: node embeddings}
\For{$t=1$ to $T$} 
    \State Sample neighborhoods $\{\mathcal{N}^{(t)}_k(\cdot)\}_{k=1}^K$ with fanouts $\{S_k\}$.
    \State Compute $\{\mathbf{h}^{(K,t)}_i\}$ by Eq.~\ref{eq:agg}--\ref{eq:sage_update}.
    \State Build $\pairembt{t}{so}$ via Eq.~\ref{eq:pair_proj}.
\EndFor
\State Stack $\mathbf{P}_{so}=[\pairembt{1}{so},\dots, \pairembt{T}{so}]\in\mathbb{R}^{T\times d_g}$.
\State \textbf{(2) Temporal Transformer}
\State Compute $\mathbf{z}^{\mathrm{graph}}_{so}$ and temporal latents via Eq.~\ref{eq:merge}--\ref{eq:block_bias}.
\State \textbf{(3) Token mapper}
\State Reduce to $M$ and project to LLM space: $\mathbf{G}_{so}\in\mathbb{R}^{M\times d}$ by Eq.~\ref{eq:reduce}--\ref{eq:gate}.
\State \textbf{(4) Prompt construction}
\State Tokenize prompt containing exactly one \texttt{<htg>} token at position $k$.
\State \textbf{(5) Token injection}
\State Replace \texttt{<htg>} with $\mathbf{G}_{so}$ using Eq.~\ref{eq:splice_main}; build $\mathbf{a}'$ and RoPE indices $\boldsymbol{\pi}$ via Eq.~\ref{eq:rope_reset}.
\State \textbf{(6) LLM forward pass}
\State $\mathbf{H}=\mathrm{LLM}_\theta(\mathbf{E}',\mathbf{a}',\boldsymbol{\pi})$ and pool $\mathbf{h}_{so}$ via Eq.~\ref{eq:pool}.
\State \textbf{(7) Logits computation} 
\State Compute $\boldsymbol{\ell}_{so}$ via Eq.~\ref{eq:logits}.\\
\Return Logits $\boldsymbol{\ell}_{so}$.
\end{algorithmic}
\end{algorithm}

\begin{algorithm}[t]
\caption{Training step with DWF (PU-safe) and ranking Loss}
\label{alg:train}
\begin{algorithmic}[1]
\Require Batch $\{\supinstance{s_i}{o_i}\}_{i=1}^B$; $\gamma,\lambda_u,\lambda_{\mathrm{reg}}$; $c_w,w^+,w^-$; $\lambda_{\mathrm{rank}}$.
\For{$i=1 \to B$}
    \State $\labscore_i \leftarrow \textsc{Forward}(s_i,o_i,\ctx_{s_i o_i})$ using Algorithm~\ref{alg:forward}.
\EndFor
\State \textbf{(Classification loss)} Compute $\mathcal{L}_{\text{batch}}$ via Eq.~\ref{eq:dwf}--\ref{eq:dwf_norm} over all $(i,c)$.
\If{$\lambda_{\mathrm{rank}}>0$}
    \State Sample negative pairs $(s_i,\tilde o_i)$ and compute $a_i^+,a_i^-$; compute $\mathcal{L}_{\mathrm{rank}}$ via Eq.~\ref{eq:rank}.
    \State $\mathcal{L}\leftarrow \mathcal{L}_{\text{batch}} + \lambda_{\mathrm{rank}}\cdot \frac{1}{B}\sum_{i=1}^B \mathcal{L}_{\mathrm{rank}}^{(i)}$.
\Else
    \State $\mathcal{L}\leftarrow \mathcal{L}_{\text{batch}}$.
\EndIf
\State Backpropagate $\nabla_{\Theta}\mathcal{L}$ and update parameters with an optimizer step.
\end{algorithmic}
\end{algorithm}

\subsection{Complexity Discussion (Precise Token/Graph Contributions)}
\label{sec:complexity}

Let $L$ be the original prompt length and $L'=L+M-1$ the spliced length.
\textbf{LLM cost.}
For a decoder-only Transformer with $d$ dimension, $N$ layers, the dominant attention cost per example is:
\begin{equation}
\mathcal{O}\left(N\cdot {L'}^2 \cdot d\right),
\end{equation}
so the overhead of graph injection is controlled by $M$ and scales quadratically in $L'$.

\textbf{Temporal GNN cost.}
With fanouts $\{S_k\}_{k=1}^K$ and $T$ snapshots, message passing is approximately:
\begin{equation}
\mathcal{O}\left(
T \cdot B \cdot \sum_{k=1}^{K}\left(\prod_{j=1}^{k} S_j\right)\cdot d_g
\right),
\end{equation}
for neighborhood expansion and aggregation (constants depend on the aggregator and implementation). The hierarchical temporal Transformer over length $T$ reduces cost relative to a single full-attention layer by merging; worst-case is $\mathcal{O}(T^2 d_g)$ at the first stage, but subsequent stages shrink quadratically as length halves.

\section{Benchmarking \llmmodel}~\label{appendix_benchmarks}
We evaluated \llmmodel against a diverse set of baseline models for multi-label relation prediction under a temporally separated protocol designed to prevent training-time information leakage. This evaluation assesses model performance on temporally held-out data, testing its ability to identify relations that emerge in later years. 
Experiments are conducted using multiple temporal cut-off years: 1990, 2000, 2010, and 2020. For each cut-off year $t$, all relation instances represented as triples $(s,r,o)$ first observed on or before $t$ are used for model training. Relation instances first observed after $t$ are reserved for evaluation.
For each cut-off year, we randomly sample a validation subset of 100,000 subject--object entity pairs from those that first appear after the cut-off year, and exclude them from the final test set.

Models perform multi-label edge prediction by assigning one or more candidate relation types to a given subject--object entity pair. Each pair is evaluated over the full relation space. For evaluation at cut-off $t$, a predicted relation type is considered a true positive if the corresponding relation instance is first observed after $t$. Relation types not observed after $t$ are treated as negative instances for metric computation within the evaluation period.

\subsection{Baselines}
The selected baseline models represent complementary modeling paradigms, enabling systematic comparison across heuristic, similarity-based, supervised learning, structural knowledge graph embedding, temporal graph modeling, and literature-based hypothesis generation approaches. 
A \emph{random baseline} assigns relations randomly among candidate relation types, serving as a lower-bound reference for evaluating whether meaningful structure can be recovered beyond chance.
\emph{KNN} (k-nearest neighbors), as a similarity-based predictor operating on subject--object representations, evaluates whether local proximity in the representation space alone can recover emerging relations without explicit relational modeling.
A graph-only, \emph{rule-based} multi-label classifier learns per-node label profiles and label-specific neighbor sets from training edges, then scores each label for a pair $(s,o)$ using label-wise common neighbors plus a weighted popularity term. The core score components are taken from Common Neighbors (CN) and Preferential Attachment (PA)~\cite{zeng2016link}.
An \emph{MLP} (multi-layer perceptron) supervised multi-label classifier trained on observed relation
instances isolates general supervised learning
capacity from explicit structural modeling of knowledge graphs.
A structural knowledge graph model based on \emph{ComplEx}~\cite{trouillon_complex_2016} captures relational structure through complex-valued representations.
A temporal graph modeling baseline based on \emph{tNodeEmbed}~\cite{singer_node_2019} captures the evolution of graph structure over time. In our implementation, sequences of entity representations over time are processed using an LSTM-based sequence encoder. Node representations are provided by the shared embedding space used across models, instead of learning time-evolving node embeddings.
Our previous model \emph{\graphmodel}~\cite{akujuobi_link_2024} is a time-aware knowledge graph–based predictor that integrates temporal and structural signals to rank plausible relations.
Finally, we here also include a literature-based hypothesis generation approach incorporating selected components of \emph{\agatha}~\cite{sybrandt_agatha_2020}, including transformer-based representations and its negative sampling strategy.
In our implementation, we do not perform graph embedding learning as in the original model; instead, node representations are derived from the shared embedding space used across models.

Several baseline methods (e.g., \emph{ComplEx} and \agatha) were originally proposed for binary link prediction or association scoring tasks. To enable comparison under our prediction setting, their prediction layers are reformulated to perform multi-label relation prediction over candidate relation types. Similarly, where applicable, input representations are constructed using the same ontology-based embeddings as used by \llmmodel. This common representation space ensures that performance differences reflect modeling capabilities rather than differences in underlying entity representations.

All models are evaluated using the same temporal split and validation protocol for each cut-off year. Due to \agatha's high resource demands and limitations on the computational infrastructure available for benchmarking, results are reported only for the 1990 and 2000 cut-offs.
Mean nDCG (averaged over $k \in \{1,3,5,10\}$) serves to assess ranking quality independently of threshold selection. Ranking metrics are computed over the full candidate set. 

\subsection{Performance across temporal cut-offs}
\label{subsec:temporal-cut-offs}
\begin{figure}[t]
    \centering
    \includegraphics[width=0.65\linewidth]{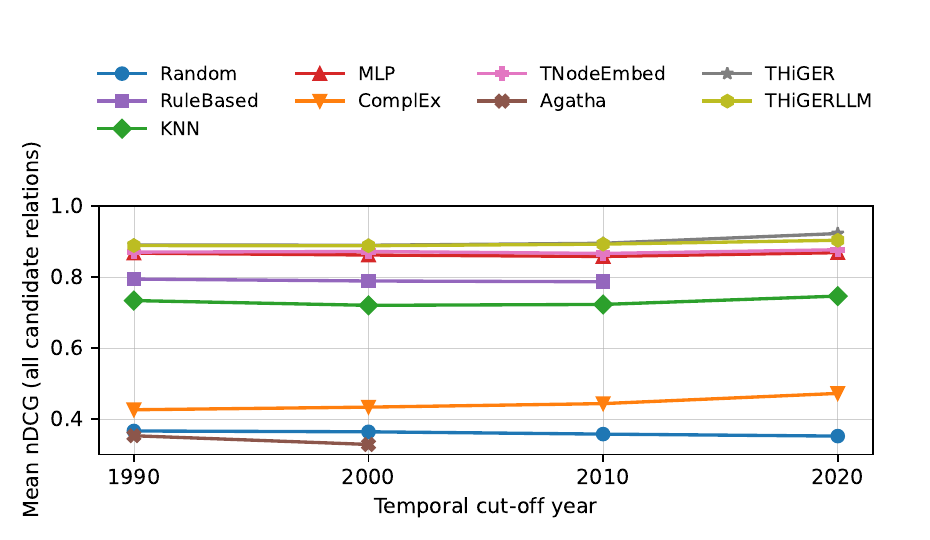}
    \caption{
    \textbf{Temporal evaluation across historical cut-offs (ranking over all candidates).}
    Mean nDCG computed over the full candidate set (including both transductive and inductive cases).
    Scores are reported for temporal cut-offs at 1990, 2000, 2010, and 2020.
    }
    \label{fig:appendix_temporal_ndcg_all}
\end{figure}

\begin{figure}[h]
    \centering

    \includegraphics[width=0.98\linewidth]{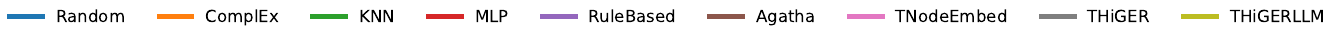}

    \vspace{0.5em}

    \includegraphics[width=0.99\linewidth]{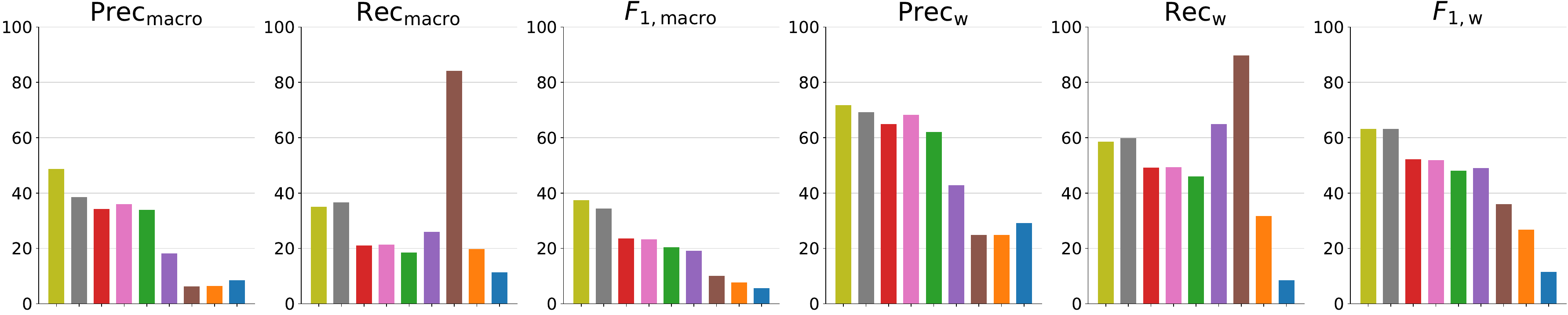}
    \includegraphics[width=0.99\linewidth]{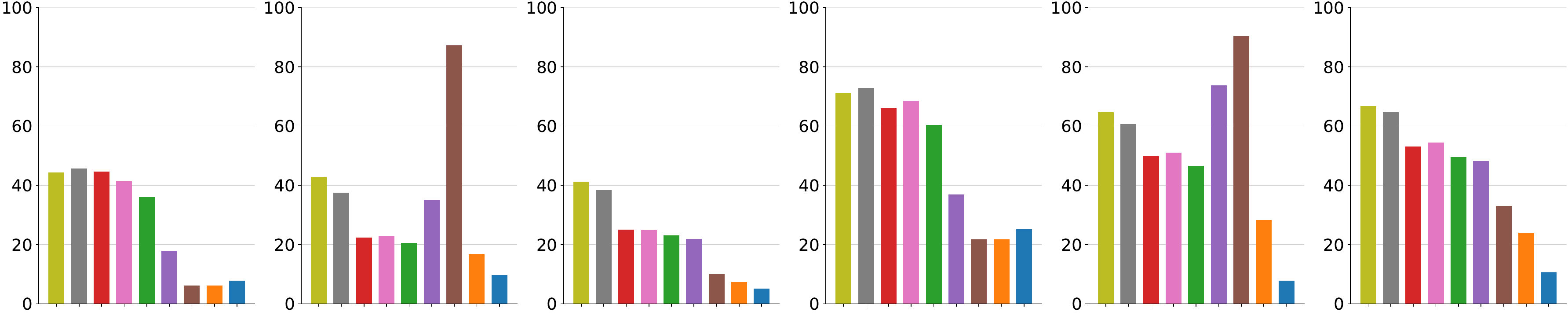}
    \includegraphics[width=0.99\linewidth]{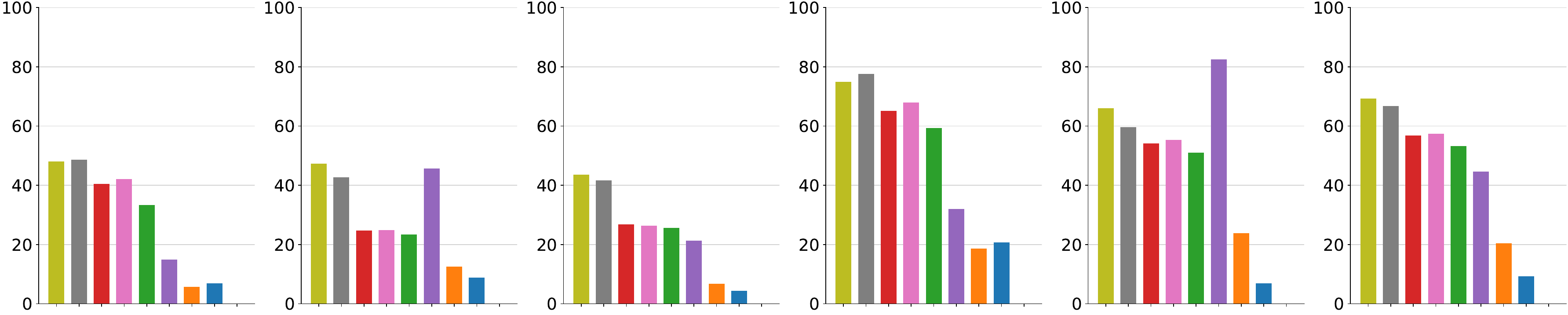}
    \includegraphics[width=0.99\linewidth]{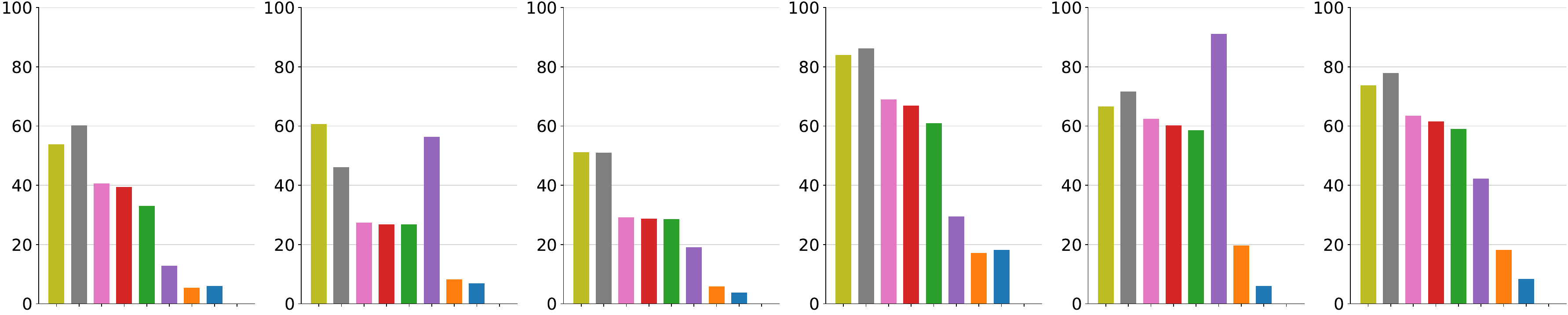}

    \caption{Benchmark performance of models trained incrementally up to different cutoff years. Rows correspond to training cut-offs at 1990, 2000, 2010, and 2020 (top to bottom). Each model is evaluated on the following decade, except the model trained up to 2020, which is evaluated on data through 2023. Columns show different metrics.}
    \label{fig:appendix_benchmark-bars}
\end{figure}

Fig.~\ref{fig:appendix_temporal_ndcg_all} shows mean nDCG over all candidate relations, again indicating consistently strong performance of \llmmodel and \graphmodel across temporal splits. 
Fig.~\ref{fig:appendix_benchmark-bars} provides a summary of model performance across different temporal cut-offs (1990, 2000, 2010, and 2020).
Overall, the trends observed in the main text remain consistent across all training horizons. 
In particular, \llmmodel maintains an advantage in $\text{F}_{\text{1,macro}}$, driven primarily by improvements in macro recall.
Also, \llmmodel achieves higher $\text{Rec}_{\text{macro}}$ than \graphmodel for all cut-off periods except 1990, indicating that the benefit of incorporating textual information becomes more pronounced as more temporal data is available.

\subsection{Per-relation-type evaluation}
\label{subsec:per-relation-type-eval}
\begin{figure}[h]
    \centering

    \includegraphics[width=0.45\linewidth]{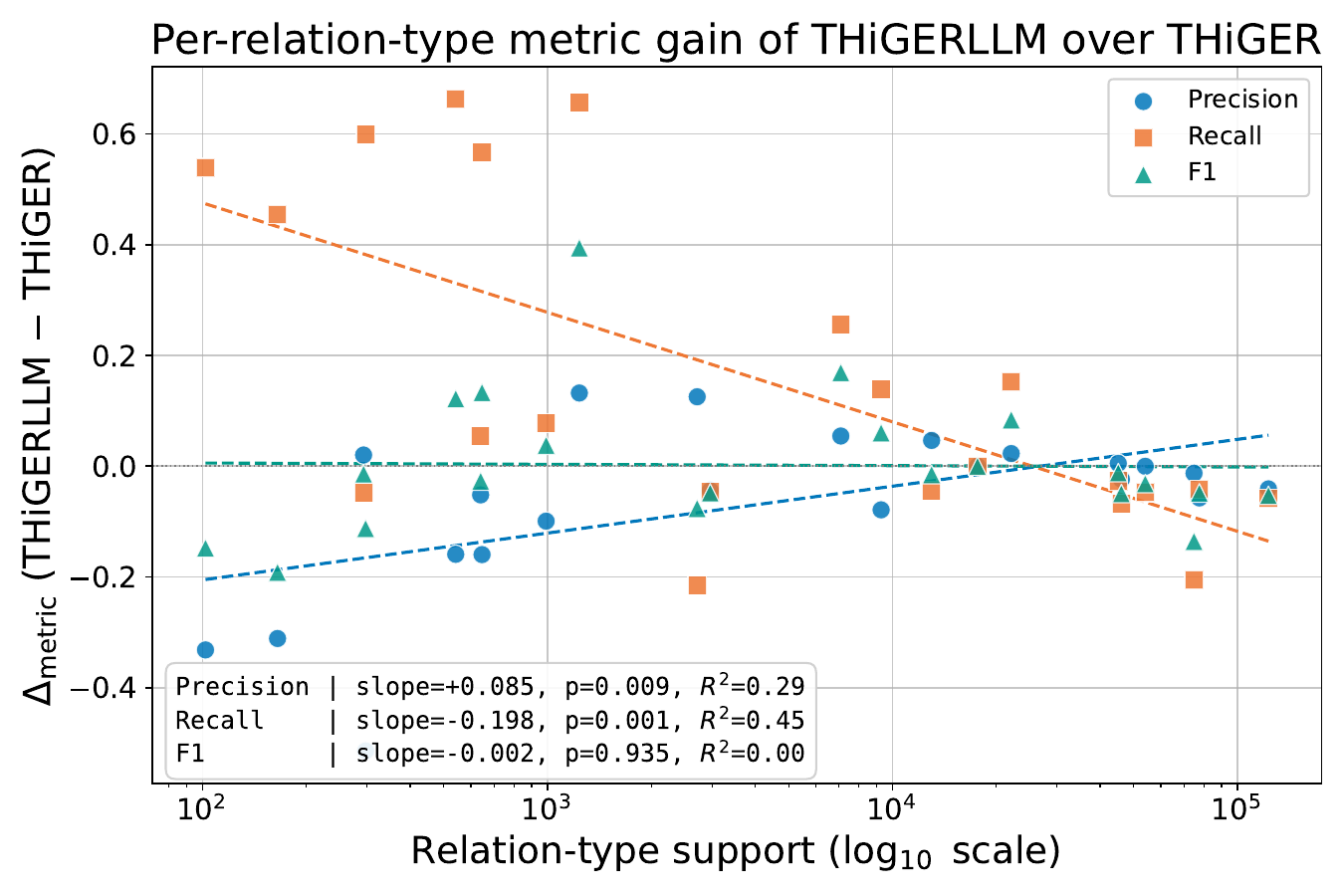}
    \includegraphics[width=0.42\linewidth]{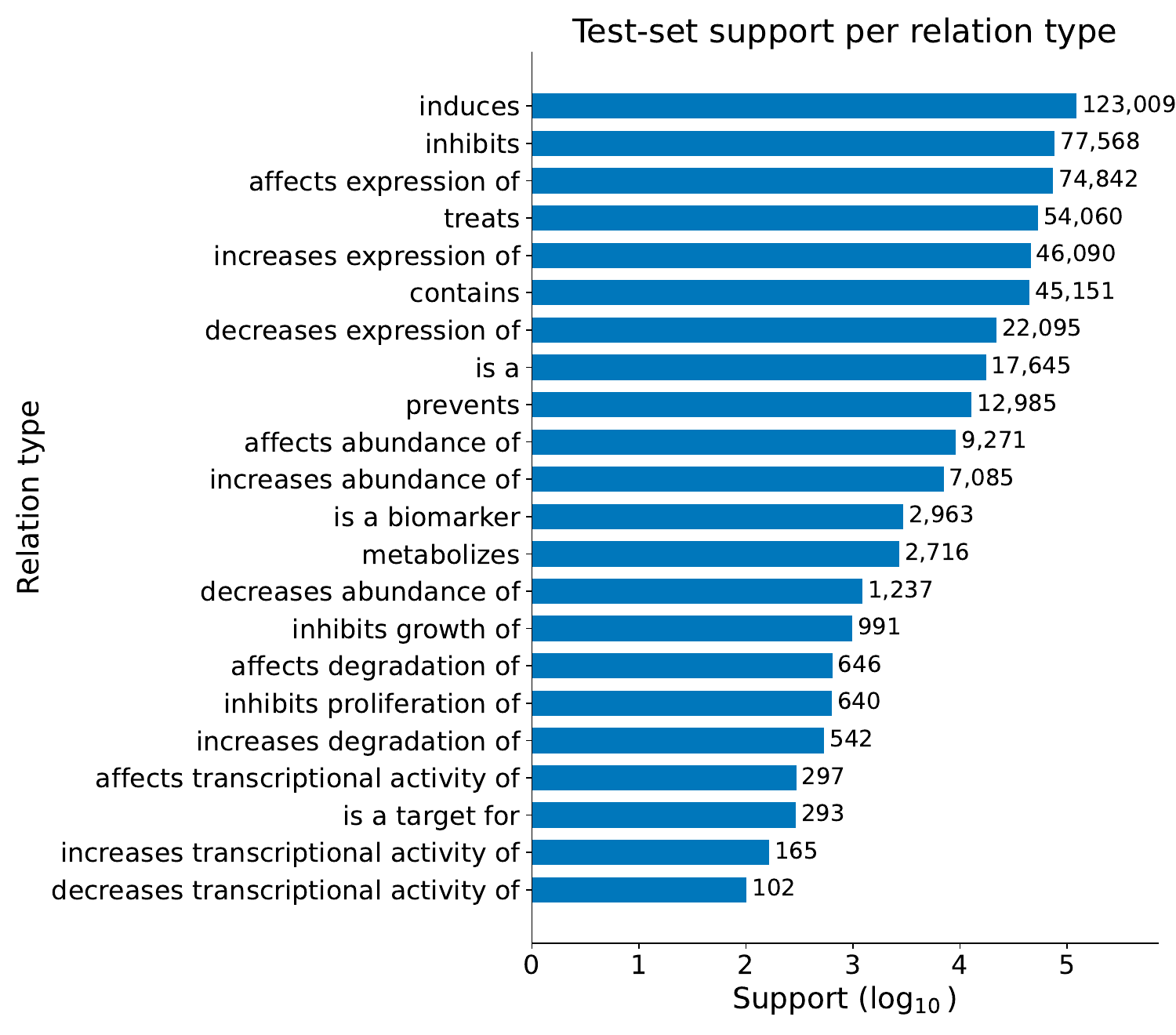}

    \caption{Per-relation-type performance gain of \llmmodel over \graphmodel as a function of relation-type  test support. \textbf{Left:} For each relation type, the change in Precision, Recall, and F1 is plotted against the number of test facts ($\log_{10}$-scaled). Each point represents one relation type. Dashed lines show ordinary least-squares fits on log-scaled support. The inset reports the slope, p-value, and $R^2$ for each metric. \textbf{Right:} Distribution of test-set support across relation types, shown on a $\log_{10}$ scale. This figure illustrates how the relative performance gain of \llmmodel varies with relation frequency.}
    \label{figapp:relation-type-gain}
\end{figure}

To understand whether the advantage of \llmmodel over \graphmodel depends on the frequency of a relation type, we computed per-relation-type differences in precision, recall, and F1 and regressed each against the log-transformed support (Figure \ref{figapp:relation-type-gain}).
The analysis reveals two significant and opposing trends.
\llmmodel achieves substantially higher recall on low-support relation types. This trend is statistically significant and indicates that the LLM-augmented model is better able to recover rare edges that \graphmodel misses. 
Conversely, this recall gain comes at the cost of precision, which is lower for low-support types, suggesting that \llmmodel produces more false positives in the long tail of the relation-type distribution. 
Importantly, these two effects cancel out in terms of F1, whose slope is not significantly different from zero. 
This indicates that the overall F1 of \llmmodel and \graphmodel is uniform across relation types regardless of their support, and is not driven by a few highly populated classes.
Taken together, these results suggest that the LLM component primarily improves the ability of the model to recall underrepresented relation types---a desirable property for knowledge graph completion tasks where the long tail of rare relations is often the most informative---while maintaining a stable F1 trade-off across the full support spectrum.

\subsection{Top-$m$ evaluation under a fixed prediction budget}
\label{subsec:top-m-eval}

In addition to threshold-based evaluation, we report a complementary top-$m$ analysis in Table \ref{tabapp:baseline-benchmark-m}, where for each entity pair only the top-$m$ predicted relation labels are retained as positive predictions.
We use $m \in \{1,3, 5\}$.
This protocol controls the prediction budget across models and therefore reduces sensitivity to threshold calibration. 
As such, it provides a useful alternative view of model behavior in the multi-label setting.

\begin{table}[h!]
    \caption{Baseline comparison for time-aware multi-label relation prediction under a 2020 cut-off using top-$m$ evaluation ($m=1,3,5$). For each entity pair, only the top-$m$ predicted relation labels are treated as positive predictions. Macro- and weighted precision, recall, and F1 are reported in percent. Unlike threshold-based evaluation, this setting compares models under a fixed prediction budget.}
    \centering
    \begin{tabular}{lccccccc}
    \toprule
    \textbf{Model} & $\mathrm{Prec}_{\mathrm{macro}}^{m}$ & $\mathrm{Rec}_{\mathrm{macro}}^{m}$ & $F_{1,\mathrm{macro}}^{m}$ & $\mathrm{Prec}_{\mathrm{w}}^{m}$ & $\mathrm{Rec}_{\mathrm{w}}^{m}$ & $F_{1,\mathrm{w}}^{m}$ \\
    \midrule
    \multicolumn{7}{c}{$m=1$} \\
    \midrule
    Random & 6.04 & 4.99 & 3.19 & 19.12 & 5.01 & 7.34 \\
    ComplEx & 5.34 & 4.14 & 3.95 & 17.15 & 9.76 & 12.29 \\
    KNN & 33.13 & 23.54 & 24.34 & 61.64 & 52.16 & 53.19 \\
    MLP & 37.51 & 25.54 & 26.43 & 65.50 & 58.46 & 57.74 \\
    RuleBased & 38.92 & 18.53 & 19.78 & 58.37 & 48.84 & 47.02 \\
    Agatha & - & - & - & - & - & - \\
    TNodeEmbed & 36.70 & 25.98 & 26.76 & 67.37 & 60.24 & 59.31 \\
    THiGER & $\mathbf{47.18}$ & $\mathbf{32.63}$ & 34.48 & 70.20 & $\mathbf{67.52}$ & $\mathbf{68.40}$ \\
    THiGERLLM & 46.85 & 32.21 & $\mathbf{35.94}$ & $\mathbf{74.26}$ & 63.64 & 66.87 \\
    \midrule
     \multicolumn{7}{c}{$m=3$} \\
    \midrule
    Random & 5.48 & 11.47 & 4.44 & 16.33 & 19.08 & 10.76 \\
    ComplEx & 5.34 & 12.52 & 6.98 & 17.15 & 30.74 & 21.90 \\
    KNN & 29.61 & 33.89 & 26.73 & $\mathbf{53.16}$ & 65.11 & 53.85 \\
    MLP & 22.07 & 47.72 & 27.14 & 44.44 & 87.05 & 56.93 \\
    RuleBased & 21.20 & 35.00 & 22.66 & 37.39 & 78.44 & 48.29 \\
    Agatha & - & - & - & - & - & - \\
    TNodeEmbed & 22.47 & 45.73 & 26.86 & 44.90 & 87.50 & 57.28 \\
    THiGER & 30.37 & $\mathbf{61.47}$ & 37.71 & 46.98 & $\mathbf{95.28}$ & 61.22 \\
    THiGERLLM & $\mathbf{38.97}$ & 59.40 & $\mathbf{44.68}$ & 50.25 & 95.06 & $\mathbf{62.66}$ \\
    \midrule
    \multicolumn{7}{c}{$m=5$} \\
    \midrule
    Random & 6.12 & 22.48 & 4.00 & 18.24 & 10.10 & 8.54 \\
    ComplEx & 5.33 & 21.12 & 8.07 & 17.12 & 48.73 & 25.16 \\
    KNN & 25.74 & 38.47 & 22.41 & $\mathbf{57.89}$ & 59.44 & $\mathbf{56.12}$ \\
    MLP & 14.87 & 63.69 & 21.48 & 34.40 & 95.45 & 48.62 \\
    RuleBased & 14.48 & 50.48 & 18.80 & 31.19 & 88.73 & 44.11 \\
    Agatha & - & - & - & - & - & - \\
    TNodeEmbed & 13.32 & 65.18 & 20.47 & 30.05 & 95.60 & 44.70 \\
    THiGER & 18.04 & 71.95 & 25.95 & 31.75 & 98.49 & 46.09 \\
    THiGERLLM & $\mathbf{29.44}$ & $\mathbf{72.34}$ & $\mathbf{38.26}$ & 38.51 & $\mathbf{98.51}$ & 51.21 \\
    \bottomrule
    \end{tabular}
    \label{tabapp:baseline-benchmark-m}
\end{table}

The results further support the conclusions drawn from the threshold-based evaluation in the main text.
Across all prediction budgets, \llmmodel achieves the strongest macro-averaged $F_1$ score, indicating that it is particularly effective when performance is evaluated uniformly across relation types rather than being dominated by frequent relations.
This is consistent with the main results, where \llmmodel improves macro recall and macro $F_1$, suggesting better coverage of less frequent relation classes.

At $m=1$, \llmmodel and \graphmodel perform very similarly, with only minor differences across metrics.
This indicates that both models produce strong top-ranked predictions and that there is no decisive advantage for either model when only a single relation label is retained per entity pair.
As the prediction budget increases from $m=1$ to $m=3$, performance generally improves, especially in terms of recall, as expected.
Under this setting, \llmmodel begins to show a clearer advantage over \graphmodel, particularly for macro-level metrics.

At $m=5$, recall becomes very high while precision decreases, reflecting the expected trade-off introduced by retaining more predicted labels: more true relations are recovered, but more false positives are also introduced.
Under this larger prediction budget, \llmmodel becomes consistently stronger, especially on macro-averaged metrics.
This reinforces the interpretation from the main text that the textual component helps recover correct triples across a broader range of relation types.

Traditional baselines such as KNN, MLP, and RuleBased remain competitive in some settings, particularly for weighted precision or weighted recall.
However, their performance is less consistent across prediction budgets and evaluation metrics than that of \graphmodel and \llmmodel.

Overall, the top-$m$ analysis confirms that \llmmodel is preferable when balanced performance across relation types is the primary objective, whereas \graphmodel remains a strong alternative and can be preferable under weighted metrics, where frequent relation types have greater influence.

\section{Predicted Hypothesis Elucidation with Literature-Inferred Explanations (\explainer)}
\llmmodel is capable of generating novel, high-confidence hypotheses in the form of $(s,r,o)$ triples.
However, these predictions are not inherently interpretable to domain experts. 
In practice, a score alone rarely provides insight into why the model expects a particular relation to hold, what evidence supports the prediction, or how it relates to existing scientific knowledge.

In scientific discovery, interpretability is essential not only for establishing trust but also for enabling meaningful action. 
Scientists must be able to understand and evaluate the rationale behind a proposed hypothesis; knowing that a connection is likely is insufficient without an explanation of the underlying factors that support it.
Prior work has shown that user trust on AI systems depend strongly on perceived understanding of model decisions, particularly when explanations are interactive or counterfactual in nature. 
This requirement is especially pronounced in high-stakes domains such as biomedicine, where scientific validation demands transparent and scientifically justifiable reasoning rather than black-box predictions.


Existing explanation methods for KG link prediction suffer from several important limitations.
They are often computationally inefficient, tied to specific model architectures, and narrow in scope, for example, restricting explanations to facts directly connected to the target triple.
These constraints substantially limit their general applicability in scientific domains.

A substantial body of prior work has focused on explaining  GNN-based predictors through various paradigms, including feature attribution methods \cite{Huang2020GraphLIMELI}, predictive subgraph identification \cite{Ying2019GNNExplainerGE,Luo2020ParameterizedEF,yuan2021on}, and counterfactual explanations \cite{Lucic2021CFGNNExplainerCE, Bajaj2021RobustCE, Kosan2022GlobalCE}.
However, these approaches are typically tailored to GNNs and do not readily generalize to other classes of KG predictors.

Another line of work targets KGE models specifically \cite{Zhao2023KEXTS}.
Complementary research has explored logic-based explanations by mining symbolic rules from KGs to identify predictive paths that support model decisions \cite{betz2022adversarial,arakelyan2021complex,sadeghian2019drum,zhang2010iteratively}
Although interpretable, such approaches can struggle to scale.
A separate stream of work examines the influence of training data on KGE predictions, estimating how perturbations to the training set affect model scores using techniques such as Taylor approximations \cite{pezeshkpour2019investigating,zhang2019data}, instance attribution methods \cite{gu2023iae,bhardwaj2021adversarial}, or model retraining \cite{rossi2022explaining,barile2025addititve}.
Additionally, several methods seek to distill KGE model scores into surrogate GNNs to extract explanatory subgraphs, often under relatively loose structural constraints\cite{ma2024kgexplainer,chang2024path,huang2024foundation,zhang2023page}

To overcome these limitations, we introduce PHELInE (Predicted Hypothesis Elucidation with Literature-Inferred Explanations), a fast, model-agnostic explanation framework that provides  contextual information to support the predictions of \predmodel, while remaining scalable and broadly applicable across KG prediction architectures.

\subsection{Notation}

We assume that our knowledge graph $\gG$ is partitioned into disjoint subsets $\gG = \Gtrain \cup \Gunseen $, where $\Gunseen$ usually comprises validation and test triples.
Finally, we consider a trained predictive model with parameters $\theta$ (e.g., \llmmodel) and a scoring function $\scorefn[\theta]{s,r,o} \in \sR$, which assigns a real-valued score to any triple $(s, r, o)$.
Refer to \ref{sec:setup} for further notation.

\subsection{Overview of the Explainer}

The input to the explainer is a hypotheses $(s, r, o)$ together with its model score $\scorefn[\theta]{s,r,o}$.
We begin by defining the search space $\gY := \{(s, r, o) \in \Omega  | C_k(s, r, o) = \true \forall k = 1, \dots, K\}$, which contains all candidate facts that satisfy a set of $K$ boolean constraints.
Candidate explanations $\gX$  are a subset of $\gY$.
This formulation allows us to restrict the search space as needed. For example, setting $C_1(s,r,o) = (s,r,o) \in \gG$ yields $\gY = \gG$, i.e., explanations composed solely of observed facts.

To quantify the influence of a candidate explanation $\gX$ on the prediction of the trained model, we employ a pseudo-retraining procedure $\retrfn$.
This algorithm approximates how modifying the training context of the model---by adding or removing the facts in $\gX$---would change the score assigned by $\scoref[\theta]$.

Based on this mechanism, we define an explanation scoring function $\Psi(\gX; s, r, o, \retrfn, \theta) \in \sR$, which measures the effectiveness of $\gX$ in supporting or influcencing the prediction for $(s, r, o)$.
The output of the framework is one or several explanations $\gX = \{s_i, r_i, o_i\}_{i=1}^k$, i.e., a set of contextual facts whose presence (or absence) significantly affects the score  $\scorefn[\theta]{s,r,o}$.

In general terms, the framework solves the following optimization problem:

\begin{equation}
\label{eq:explainability}
\{\gX^\star_1, \dots, \gX^\star_K\}
=
\operatorname{TopK}_{\gX \subseteq \mathcal{Y} \setminus \varnothing}
\,\Psi(\gX; s, r, o, \retrfn, \theta).
\end{equation}

This problem depends on the specification of the search space, the underlying prediction model, and several core components.
First, a pseudo-retraining algorithm estimates how each candidate explanation affects the score of the model.
Next, a candidate finder identifies the most promising explanation sets within the constrained search space.
These candidates are then evaluated by an explanation scoring function, which measures their relevance or influence on the prediction.
Finally, a reranking strategy organizes the top candidates according to user-defined requirements.

The framework is model-agnostic, requiring only query access to the scoring function $\scorefn[\theta]{s,r,o}$;
it does not rely on model weights or training details, making it applicable even to black-box predictors.
It also offers a flexible candidate search mechanism, allowing different strategies for exploring the search space---for example, searching over graph-derived structures such as paths.
Moreover, the scoring mechanism accommodates multiple explanation objectives, such as sufficiency and necessity, enabling users to tailor explanations to their preferred interpretive lens.

\begin{algorithm}[t]
\caption{Model-agnostic Explanation Framework}
\label{alg:explanation-framework}
\begin{algorithmic}[1]
\Require
  hypothesis $(s, r, o)$; trained model with scoring function $\scorefn[\theta]{\cdot}$; search space $\gY$; pseudo-retraining procedure $\retrfn$; candidate finder $\mathsf{FindCandidates}$; scoring function $\Psi$; reranking strategy $\mathsf{Rerank}$; number of explanations $K$
\Ensure
  explanations $\{\mathcal{X}^\star_1, \dots, \mathcal{X}^\star_K\}$ for $(s,r,o)$

\Statex

\Statex \textbf{Generate candidate explanations}
\State $\mathcal{C} \gets \mathsf{FindCandidates}(\mathcal{Y}, s, r, o)$
\Comment{$\mathcal{C} \subseteq 2^{\mathcal{Y}} \setminus \{\varnothing\}$}

\Statex

\Statex \textbf{Score candidate explanations}
\State Initialize list $\mathcal{L} \gets [\,]$
\ForAll{$\mathcal{X} \in \mathcal{C}$}
  \State $\psi \gets \Psi(\mathcal{X}; s, r, o, \retrfn, \theta)$
  \Comment{influence of $\mathcal{X}$ on $\scorefn[\theta]{s,r,o}$ via $\retrfn$}
  \State append $(\mathcal{X}, \psi)$ to $\mathcal{L}$
\EndFor

\Statex

\Statex \textbf{Select top-$K$ explanations}
\State $\{\mathcal{X}_1, \dots, \mathcal{X}_K\}
  \gets \operatorname{TopK}_{(\mathcal{X}, \psi) \in \mathcal{L}}(\psi, K)$

\Statex
\Statex \textbf{Rerank according to user requirements}

\State $\{\mathcal{X}^\star_1, \dots, \mathcal{X}^\star_K\}
  \gets \mathsf{Rerank}(\{\mathcal{X}_1, \dots, \mathcal{X}_K\}, s, r, o)$ 

\State \Return $\{\mathcal{X}^\star_1, \dots, \mathcal{X}^\star_K\}$
\end{algorithmic}
\end{algorithm}

\subsection{Candidate finder}

The goal of the candidate finder is to identify the the subset of $\gY$ that is most relevant for optimizing the objective in Eq. \ref{eq:explainability}.
We propose two complementary modes for deriving these candidate sets, depending on whether the user prefers to rely strictly on observed graph structure or to incorporate plausible but unobserved relations.

\begin{algorithm}[t]
\caption{Latent Path Candidate Finder}
\label{alg:plausible-path-finder}
\begin{algorithmic}[1]
\Require
  hypothesis $(s, r, o)$; scoring function $\scorefn[\theta]{\cdot}$; entity set $\nodeset$; relation set $\relset$; path length $k$; score threshold $\tau$; number of candidate paths $N$; maximum trials per step $T$
\Ensure
  set of plausible paths $\mathcal{C}_{\mathrm{path}}(s,r,o)$
\State $\mathcal{C}_{\mathrm{path}}(s,r,o) \gets \varnothing$

\For{$n = 1$ to $N$}
    \State $e_{0} \gets s$; \quad $P \gets [\,]$
    \For{$i = 1$ to $k$}
        \State \textit{accepted} $\gets$ false
        \For{$t = 1$ to $T$}
            \State sample $r_i \sim \mathrm{Unif}(\relset)$
            \If{$i < k$}
                \State sample $e_i \sim \mathrm{Unif}(\nodeset)$
            \Else
                \State $e_i \gets o$ \Comment{final hop must end at $o$}
            \EndIf
            \State $\psi \gets \scorefn[\theta]{e_{i-1}, r_i, e_i}$
            \If{$\psi \ge \tau$}
                \State append $(e_{i-1}, r_i, e_i)$ to $P$
                \State \textit{accepted} $\gets$ true
                \State \textbf{break}
            \EndIf
        \EndFor
        \If{not \textit{accepted}}
            \State \textbf{break} \Comment{no plausible continuation}
        \EndIf
    \EndFor
    \If{$|P| = k$}
        \State $\mathcal{C}_{\mathrm{path}}(s,r,o) \gets \mathcal{C}_{\mathrm{path}}(s,r,o) \cup \{P\}$
    \EndIf
\EndFor

\State \Return $\mathcal{C}_{\mathrm{path}}(s,r,o)$
\end{algorithmic}
\end{algorithm}

\subsubsection{Corpus Path Candidate Finder} 

Given a hypothesis $(s,r,o)$, this approach enumerates all relational paths of length $k$ in the knowledge graph $\gG$ (corpus) that connect the source entity $s$ to the target entity $o$.
All triples occurring along these paths are collected as candidate explanations.
This method leverages the intuition that explicit graph connectivity often captures meaningful relational structure: paths encode multi-hop logical patterns that frequently provide useful context for the predicted relation.

\subsubsection{Latent Path Candidate Finder} 
The second mode extends the search beyond the observed knowledge graph. Given a $(s,r,o)$, the method seeks paths of length $k$ connecting $s$ and $o$ where intermediate triples may be either observed in $\gG$ or \emph{plausible} according to the prediction model.
Such triples are not required to appear in the current graph but must received sufficiently high scores under $\scoref[\theta]$.
This enables the framework to incorporate relational structure that the model considers plausible, even if it is not explicitly present in the graph.

For example, for a hypothesis $(\text{drugA}, \text{treats}, \text{diseaseB})$, a plausible path might include $(\text{drugA}, \text{interactsWith}, \text{proteinC})$ and 
$(\text{proteinC}, \text{involvedIn}, \text{diseaseB})$ even if the first triple is not observed in the knowledge graph but is highly plausible according to the model.
This mode allows explanations to capture latent or emerging relationships that the model relies on but that the graph has not yet recorded.

\subsection{Pseudo-retraining Procedure}
The goal of the pseudo-retraining procedure is to approximate how the prediction model $\scorefn[\theta]{s,r,o}$ would behave if it were retrained on a modified training graph $\Gtrain[\prime] \neq \Gtrain$, resulting in updated parameters $\theta^\prime$.
In principle, to evaluate the influence of a candidate explanation $\gX$ on a hypothesis $(s,r, o)$,  one would need to retrain the predictive model on a graph in which the triples in $\gX$ have been added or removed.
However, performing an actual retraining for every candidate explanation is computationally infeasible.

To circumvent this, we train a surrogate model  $\surro[\phi]{s,r,o \mid \gS}$ that estimates how the original score of the model would change under different training graphs.
The surrogate is trained only once and then used to approximate the behavior of $\scorefn[\theta]{\cdot}$ under perturbations of the training graph.

Concretely, the surrogate is fitted to satisfy

\begin{equation}
\surro[\phi]{s,r,o ; \Gtrain} \approx  \scorefn[\theta]{s,r,o}
\end{equation}

for both positive and negative triples. The training objective is

\begin{equation}
\mathcal{L}(\phi) = \sum_{(s,r,o) \in \Gtrain[\star]} \left( \surro[\phi]{s,r,o ; \Gtrain} - \scorefn[\theta]{s,r,o} \right)^2
\end{equation}

where $\Gtrain[\star]$ denotes the union of observed training triples and sampled negative triples. The surrogate model must satisfy two key requirements:

\textbf{Graph-conditional scoring:} It must allow efficient evaluation of scores under a modified graph $\Gtrain[\prime]$, enabling us to simulate the effect of adding or removing explanation triples.

\textbf{Computational efficiency:} Score evaluation on different graphs should be achievable with a single (or very few) forward passes. This is essential to compute explanations at scale—for large knowledge graphs and thousands of candidate explanations—in practical runtimes (on the order of seconds).

By using the surrogate as an approximation to full retraining, our framework achieves a tractable method for estimating the influence of candidate explanations while avoiding the prohibitive cost of repeatedly retraining the prediction model. In our experiments, we use GraphSAGE \cite{hamilton2017inductive} as the surrogate model architecture.

\subsection{Explanation Scoring and Types}

To evaluate the quality of a candidate explanation $\gX$ for a hypotehsis $(s,r,o)$, we define two complementary scoring objectives: sufficiency and necessity.
These scores quantify how strongly the triples in $\gX$ influence the output of the model and allow the framework to support different explanation goals.

\subsubsection{Sufficiency}
The sufficiency score measures how well the triples in $\gX$ alone can reproduce the model’s score for $(s,r,o)$.
Intuitively, an explanation is sufficient if adding only the explanation triples to an otherwise empty (or minimally informative) graph yields a high predicted score.
Formally, using the surrogate model $\phi$,
\begin{equation}
    \suff[\gX] = \surro[\phi]{s,r,o \mid \gX} - \scorefn[\theta]{s,r,o},
\end{equation}
which quantifies how strongly $\gX$ \emph{supports} the hypothesis.

\subsubsection{Necessity}
The necessity score measures how much the model’s prediction deteriorates when the triples in $\gX$ are removed from the training graph. 
An explanation is necessary if its removal causes a strong drop in the predicted score.
Let $\Gtrain \setminus \gX$ denote the graph with the explanation removed. The necessity score is then defined as

\begin{equation}
    \nece[\gX] = \scorefn[\theta]{s,r,o} - \surro[\phi]{s,r,o \mid \Gtrain \setminus \gX},
\end{equation}

which captures how \emph{critical} $\gX$ is to the original prediction of the model.

By selecting one or combining both, users can tailor explanations to emphasize the most supportive or indispensable facts behind a given prediction.

\subsection{Reranking}

After candidate explanations have been generated and scored, the final step of the framework is to rerank the top candidates according to user-defined preferences.
The reranking stage allows the framework to produce explanations that satisfy desiderata that go beyond the raw optimization objective in Eq. \ref{eq:explainability}.
In our implementation, we prioritize unique pathways, meaning that we rank higher those explanations that correspond to distinct relational paths—i.e., paths that involve different intermediate entities. 
This ensures that the user is presented with diverse reasoning patterns rather than multiple variants of the same underlying path.

However, the methodology is deliberately general. The reranking function $\rerankfn$ can implement a wide range of criteria, including:
\begin{itemize}
    \item Domain-specific constraints, such as prioritizing biomedical pathways, causal directionality, or temporal consistency.
    \item Faithfulness-oriented reranking, emphasizing explanations with the strongest sufficiency or necessity scores.
\end{itemize}

The output of the reranking step is a collection of explanation subgraphs or sets of contextual triples $\{\gX^\star_1, \dots, \gX^\star_K\}$, representing the top explanations under the chosen criterion.

\subsection{Final Thoughts}

\subsubsection{Model-agnostic design} 
A key strength of the proposed framework is its model-agnostic nature. 
Because it interacts with the predictive model solely through queries to the scoring function $\scorefn[\theta]{s,r,o}$, the approach applies to any knowledge graph model—embedding-based, rule-based, or hybrid—without requiring access to model parameters, gradients, or training procedures.
At the same time, if any of these internal components are available, such as model weights, intermediate embeddings, or training metadata, the framework can leverage them to enhance the pseudo-retraining procedure, yielding more accurate or more efficient explanation estimates. 
This flexibility allows the method to function in fully black-box settings while still benefiting from richer model access when possible.

\subsubsection{Scalability} 
The framework is designed with scalability in mind.
The candidate finder restricts the search space to promising graph neighborhoods or plausible paths; the surrogate model provides rapid score estimates; and the reranking step operates on a small subset of top candidates. 
Together, these components make it possible to generate explanations in seconds even for large graphs containing millions of triples.

\subsubsection{Limitations and future directions} 
Despite its strengths, several challenges remain.
The fidelity of the explanations depends on the ability of the surrogate model---GraphSAGE in our experiments---to accurately approximate the behavior of the original predictive model under graph perturbations.
Improving surrogate fidelity or incorporating uncertainty estimates would strengthen trust in the explanations.

Additionally, while our scoring functions capture sufficiency and necessity, they do not explicitly address causal notions of influence; extending the framework with causal or counterfactual reasoning remains an important direction for future work.

\section{Biomedical Data Provenance, Extraction \& Cleaning}~\label{appendix_data}
The primary dataset used for the evaluation of Hakken in biomedical science was made up of information derived from PMC Open Access (full text articles) and MEDLINE (titles and abstracts), together with select licensed data. We describe the relevant extraction and normalization process utilized as part of the processing pipeline at a high level below.

\subsection{Relation extraction}
Relations are extracted from scientific publications using a rule-based relation extraction framework, which transforms unstructured text into normalized semantic triples of the form (head entity, relation type, tail entity). Each triple links two ontology-grounded entities via a predefined semantic relation, with directionality determined by the rule patterns applied.

The framework relies on three core components: domain ontologies, rule-based NLP, and multi-level syntactic simplification. First, named entities are identified and mapped to ontology concepts during document indexing (entity linking). The annotated text is then converted into structured phrase tokens, enabling the application of rule-based patterns to detect candidate relations.

These candidates are refined through multiple processing stages, including sentence simplification, resolution of nested expressions and enumerations, and selection of the most specific applicable relation type. Relation arguments are normalized to canonical concept identifiers and semantic roles. Rules are designed to reduce ambiguity, and heuristic filtering is applied to discard structurally incomplete or low-confidence matches, where confidence is computed as a weighted average of the matched pattern reliability and the confidence scores of the participating entities. Negation and speculative statements are currently not included in the data and are acknowledged as a current limitation of this extraction procedure.

\subsection{Ontology construction and versioning}
The ontologies used are created through a practical, iterative process that combines trusted external knowledge sources, automation, and expert curation. For each domain, the scope and intended use (such as text mining or semantic search) is defined first, then authoritative sources are selected as starting points—for example MeSH and MONDO for diseases, UniProt and HGNC for genes and proteins, or 
ChEBI for materials and substances. Concepts, names, synonyms, and existing hierarchies are automatically ingested and merged where possible, such as merging disease concepts that share the same preferred name across sources, or— in the Chemistry ontology—automatically assigning millions of compounds to chemical classes like steroids or monoterpenes using structure-based algorithms. 
A high-level hierarchy is then established to organize concepts from broad categories (e.g. disease of anatomical entity) down to specific entries (e.g. Parkinson’s disease), after which ontology experts manually review and refine the content to resolve ambiguities, remove redundancies, and enrich synonym coverage (for example adding common literature terms or brand names). Where needed, additional semantic relationships beyond simple hierarchies are modeled, such as linking a drug trade name to its active ingredient (e.g. Tylenol → acetaminophen) or connecting a gene to its protein product (e.g. TP53 → p53). The ontologies are continuously quality-checked, versioned, and updated through regular incremental releases and annual major releases, and are delivered in multiple standard formats so they can be directly applied in text mining, semantic search, and knowledge graph applications.

\subsection{Quality control}
The quality of the underlying extraction pipeline is assessed through a multi-stage evaluation process. For named entity recognition (NER), large-scale statistics are computed over a benchmark of approximately 100k PubMed Central Open Access subset (PMC) articles, complemented by targeted manual validation on randomly sampled annotations to verify entity recognition and normalization accuracy.

For relation extraction, evaluation is performed in two stages. First, a curated development set of 230 sentences from selected PMC Open Access articles is used for detailed precision analysis. Second, large-scale validation is conducted on a corpus of approximately 10k PMC Open Access articles, where aggregate statistics are computed and compared across successive pipeline versions to monitor consistency and improvements over time.

As this extraction framework is an established component provided externally and evaluated independently of the present study, we do not include an additional manual audit of sampled triples here. Instead, we rely on the provider’s standardized evaluation procedures and continuous quality monitoring to ensure the consistency and reliability of the dataset used in our experiments.

\subsection{Polarity Conflicts in Source Data}

The data provided showed some polarity conflicts when dealing with relationships containing the keywords AFFECTS, INCREASES, DECREASES. Indeed, several relation-
type triplets appeared
in conflicting or inconsistent patterns. Expert-guided rules were implemented to
reconcile these cases. Illustrative examples and resulting decisions are provided in Table~\ref{tab:affect_rules}.

\begin{table}[h]
\centering
\caption{Examples and rules used to resolve inconsistencies among \texttt{AFFECTS}, \texttt{INCREASES}, and \texttt{DECREASES}.}
\label{tab:affect_rules}
\begin{tabular}{p{3.8cm} p{5.2cm} p{5.2cm}}
\toprule
\textbf{Scenario} & \textbf{Issue} & \textbf{Resolution} \\
\midrule

Only \texttt{AFFECTS} present  
(e.g., A affects B @ t1)
& No inconsistency; base relation already present. 
& \textbf{Do nothing.} AFFECTS is kept. \\

Only \texttt{INCREASES} present  
(e.g., A increases B @ t1)
& Missing the more general parent relation. 
& \textbf{Add AFFECTS with same timestamp} (t1); keep INCREASES. \\

Only \texttt{DECREASES} present  
(e.g., A decreases B @ t2)
& Missing the more general parent relation. 
& \textbf{Add AFFECTS with same timestamp} (t2); keep DECREASES. \\

Both \texttt{INCREASES} and \texttt{AFFECTS} present  
(e.g., A affects B @ t1; A increases B @ t1)
& No inconsistency; parent + child relation already aligned. 
& \textbf{Do nothing.} Both are kept. \\

Both \texttt{DECREASES} and \texttt{AFFECTS} present  
(e.g., A affects B @ t3; A decreases B @ t3)
& No inconsistency; parent + child relation already aligned. 
& \textbf{Do nothing.} Both are kept. \\

Both \texttt{INCREASES} and \texttt{DECREASES} present  
(e.g., A increases B @ t1; A decreases B @ t4)
& Mutually contradictory child relations.  
A general relation (AFFECTS) is missing.
& \textbf{Remove INCREASES and DECREASES},  
\textbf{add AFFECTS with the earlier timestamp} (t1). \\

\bottomrule
\end{tabular}
\end{table}

Our data-cleaning strategy comprises four major phases: preliminary expert analysis, domain search, edge cleaning, and node cleaning. A panel of biomedical experts was involved in examining the relation types present in the dataset and prioritizing them---in preparation for the eventual wet-lab validation of hypotheses described below---according to their relevance for drug discovery and genetics/genomics. This analysis was also used to identify opportunities for simplifying relation semantics and to recommend the merging or renaming of low-frequency or ambiguous relation types. Based on this expert review, we restricted the set of possible relation types occurring in our biomedical dataset and renamed or merged some relations for semantic alignment. The expert selected relations and transformation rules are summarized in Table~\ref{tab:relation_transformations}. 

\begin{table}[!h]
\centering
\caption{Relation-type renaming, merging, or removal based on expert review.}
\label{tab:relation_transformations}
\begin{tabularx}{1.2\textwidth}{cXlX}
\toprule
\textbf{\#} & \textbf{Original relation} & \textbf{Action} & \textbf{Transformed relation} \\
\midrule
1  & INDUCES                                     & keep                        & -- \\
2  & IS\_A\_BIOMARKER\_FOR                       & keep                       & -- \\
3  & OF                                          & merge into                        & IS\_A\_BIOMARKER\_FOR \\
4  & INHIBITS                                    & keep                        & -- \\
5  & TREATS                                      & keep                        & -- \\
6  & IS\_TREATED\_BY                             & swap (s$\leftrightarrow$o)   & TREATS \\
7  & USED\_FOR\_TREATMENT\_OF                    & rename to                         & TREATS \\
8  & IS\_DIFFERENTIATED\_BY                      & keep                        & -- \\
9  & IS\_A\_TARGET\_FOR                          & keep                        & -- \\
10 & IS\_A                                       & keep                        & IS\_A \\
11 & METABOLIZES                                 & keep                        & METABOLIZES \\
12 & IS\_METABOLIZED\_BY                         & swap (s$\leftrightarrow$o)   & METABOLIZES \\
13 & IS\_METABOLITE\_OF                          & rename to                         & METABOLIZES \\
14 & CONTAINS                                    & keep                        & -- \\
15 & PREVENTS                                    & keep                      & -- \\
16 & INCREASES\_DEGRADATION\_OF                  & keep                      & -- \\
17 & DECREASES\_TRANSCRIPTIONAL\_ACTIVITY\_OF    & keep                     & -- \\
18 & INCREASES\_TRANSCRIPTIONAL\_ACTIVITY\_OF    & keep                      & -- \\
19 & AFFECTS\_TRANSCRIPTIONAL\_ACTIVITY\_OF      & keep                      & -- \\
20 & INCREASES\_EXPRESSION\_OF                   & keep                      & -- \\
21 & EXPRESSION\_IS\_INCREASED\_IN               & swap (s$\leftrightarrow$o)   & INCREASES\_EXPRESSION\_OF \\
22 & DECREASES\_EXPRESSION\_OF                   & keep                       & -- \\
23 & EXPRESSION\_IS\_DECREASED\_IN               & swap (s$\leftrightarrow$o)   & DECREASES\_EXPRESSION\_OF \\
24 & AFFECTS\_EXPRESSION\_OF                     & keep                       & -- \\
25 & EXPRESSION\_IS\_AFFECTED\_IN                & swap (s$\leftrightarrow$o)   & AFFECTS\_EXPRESSION\_OF \\
26 & INHIBITS\_GROWTH\_OF                        & keep                      & -- \\
27 & INHIBITS\_PROLIFERATION\_OF                 & keep                       & -- \\
28 & DECREASES\_ABUNDANCE\_OF                    & keep                     & -- \\
29 & AFFECTS\_ABUNDANCE\_OF                      & keep                        & -- \\
30 & INCREASES\_ABUNDANCE\_OF                    & keep                      & -- \\
31 & AFFECTS\_DEGRADATION\_OF                    & keep                       & -- \\
\bottomrule
\end{tabularx}

\end{table}

For the domain search, we used the provided ontology information to obtain more biologically interpretable macro-domains by recursively tracing each entity’s ancestry until the highest-level domain was identified. This procedure resulted in 22 candidate macro-domains.
Subsequent edge cleaning included the removal of exact duplicate entries, duplicate triplets with later timestamps (only retaining the earliest occurrence of a triplet), invalid and null entries, and of extremely infrequent relations. Also, based on expert guidance, we resolved relationships appearing in inconsistent or contradictory patterns. These are polarity conflicts such as increases vs decreases. In Appendix \ref{appendix_data} we report an example of such cases. 
The final node cleaning assigns valid domains to entities with missing or inconsistent annotations using the role tag provided with the data. Where an entity's role corresponded to multiple macro-domains we performed a mapping to the dominant macro-domain based on incoming and outgoing edges.

This results in a cleaned dataset containing \num{254806} entities and \num{7127960} triples across 23 relation types and 20 macro-domains (as two macro-domains were found to be empty after cleaning).

\section{Wet-lab validation of hypotheses}
\subsection{Wet-lab hypothesis generation}

Generating all possible hypotheses encoded in the knowledge graph would be computationally prohibitive and would hinder the prioritization of candidates for a specific topic to send for wet-lab investigation. Preliminary estimates indicate that exhaustive exploration of all candidate triples would require orders of magnitude more computational time than is practical for iterative experimental cycles. To reduce the search space and retain biological relevance, we restricted predictions to a curated subset of entities.
We chose to focus on the topic of `aging' and identify promising hypothesis around this topic. The selection process  starts from a compiled list of genes provided by the biomedical experts and associated with the chosen topic.



 This curated list is available in the Supplementary Information. After receiving the list, we aligned all identifiers to the entity vocabulary used in the knowledge graph to ensure seamless integration. This produced our set of seed nodes.

We applied \llmmodel to predict all potential relations among all pairwise combinations of these nodes. For a total of \num{1386} entity nodes, this corresponds to \num{959805} unordered entity pairs. The resulting set consisted of \num{1543297} above-confidence-threshold hypotheses, which were subsequently prioritised through the selection strategy described below, for expert review and wet-lab evaluation.




\subsection{Wet-lab hypothesis selection}

The hypothesis-selection pipeline processes the raw predictions produced by the \llmmodel model and yields a refined list of mechanistically plausible hypotheses. Following completion of the pipeline and addition of corresponding explanations from PHELInE, three hypotheses were selected by biomedical experts for wet-lab validation and submitted to an independent Contract Research Organization (CRO).




The pipeline consists of five principal stages, each implementing a specific filtering or characterization step, designed to progressively reduce noise and increase the chance to spot promising and novel routes:

\begin{enumerate}
    \item \textbf{Preparation:}
    This stage standardizes model output and removes any hypotheses corresponding to edges already present in the reference knowledge graph.
    \begin{itemize}
        \item Converts raw predictions to a uniform tabular schema.
        \item Excludes entity pairs for which a relation is already recorded in the graph, ensuring that only novel hypotheses are considered.
    \end{itemize}

    \item \textbf{Recency characterization:}
    For each entity, we compute a recency score defined as the median publication year associated with that node. The result is used as additional discriminator when reviewed by experts.  


    \item \textbf{Top-$k$ entity filtering:}
    To limit the dominance of highly connected entities and ensure a more diverse hypothesis set, we retain only the top-$k$ highest-confidence predictions associated with each entity. In the final pipeline, $k=3$ was used.

    \item \textbf{Path-length characterization:}
    For each candidate prediction, we compute the shortest path between the two entities in the reference graph (the graph prior to the predictions). This metric gives an idea of the complexity of the proposed hypothesis and can help an expert eye prioritize a prediction over another.


    \item \textbf{Finalization:}
    The final stage formats the selected hypotheses into standardized output files. Separating this step allows flexible formatting or visualization changes without altering the core logic.
\end{enumerate}


Table~\ref{tab:filtering_stats} summarizes the number of hypotheses retained after each filtering stage.

\begin{table}[h!]
\centering
\caption{Filtering statistics for hyothesis selection pipeline.}
\label{tab:filtering_stats}
\begin{tabular}{lccc}
\toprule
\textbf{Step} & \textbf{Input hypotheses} & \textbf{Output hypotheses} & \textbf{Notes} \\
\midrule
prepare & 1,543,297 & 1,524,046 & Removes existing graph relations. \\
filter\_topk\_entities & 1,524,046 & 2,804 & Top-3 filter applied. \\
\bottomrule
\end{tabular}
\end{table}

The distribution of shortest path lengths among the final hypotheses is shown in Table~\ref{tab:path_lengths}.

\begin{table}[h!]
\centering
\caption{Shortest-path-length distribution among final Batch~2 hypotheses.}
\label{tab:path_lengths}
\begin{tabular}{lc}
\toprule
\textbf{Path length} & \textbf{Count} \\
\midrule
3 & 104 \\
4 & 2,068 \\
5 & 631 \\
6 & 1 \\
\bottomrule
\end{tabular}
\end{table}


Out of the almost \num{3000} prioritized hypotheses, the biomedical experts chose the three hypotheses with confidence above $0.8$, that are reported in Table \ref{tab:selected_hypotheses}.

\begin{table}[h!]
\centering
\caption{List of the selected hypotheses for wet-lab validation. E1=entity 1 in the relation, E2=entity 2 in the relation, MPL=minimum path length exisiting between the two entitites.}
\label{tab:selected_hypotheses}
\begin{tabular}{lllllll}
\toprule
\textbf{E1} & \textbf{E2} & \textbf{Relation} &
\textbf{Recency E1} & \textbf{Recency E2} & \textbf{MPL} & \textbf{Confidence} \\
\midrule
SOAT1 & STAT3 & affects\_transcriptional\_activity\_of & 2021 & 2021 & -- & 0.84183 \\
RAF1  & TNF   & decreases\_expression\_of              & 2021 & 2011 & -- & 0.83114 \\
TP53  & BAMBI & affects\_expression\_of                & 2021 & 2018 & -- & 0.85391 \\
\bottomrule
\end{tabular}
\end{table}




\subsection{Wet-lab hypothesis validation}
We performed validation experiments of the hypotheses using human cell lines. The three selected hypotheses involved relationships between SOAT1-STAT3, RAF1-TNF, and TP53-BAMBI (Table~\ref{tab:wetlabresults}).\\

\begin{table}[h!]
\caption{List of the selected hypotheses for wet-lab validation. E1=entity 1 in the relation, E2=entity 2 in the relation.}
\label{tab:wetlabresults}
\begin{tabular}{llll}
\toprule
\textbf{E1} & \textbf{E2} & \textbf{Relation} & \textbf{Validation Summary} \\
\midrule
SOAT1 & STAT3 & affects\_transcriptional\_activity\_of & Denied \\
RAF1  & TNF   & decreases\_expression\_of              & Supported \\
TP53  & BAMBI & affects\_expression\_of                & Supported \\
\bottomrule
\end{tabular}
\end{table}

\paragraph{SOAT1-STAT3 hypothesis}
The hypothesis that SOAT1 activity affects STAT3 transcriptional activity was tested in HepG2 cells. LDL was used to activate SOAT1 activity, and downstream effects on STAT3 phosphorylation and STAT3 mRNA expression were assessed. Cholesterol ester (CE) quantification served as an internal control to ensure that SOAT1 activity was effectively modulated.

Exposure to LDL in both the absence and presence of serum showed a concentration-dependent increase in CE levels (1\% serum: 10.5, 14.2 $\mu$M and 0\% serum: 6.7, 10.7 $\mu$M with 100 and 200 $\mu$g/mL LDL respectively), consistent with SOAT1 activation. Treatment with LDL caused an increase in CE formation which was inhibited by the SOAT1 inhibitor nevanimibe by up to 50\% at both 4 and 24 hours. Cholesterol ester levels were slightly higher after 4 hours than 24 hours, with basal CE levels of approximately 5 $\mu$M. LDL (100 $\mu$g/mL) increased CE approximately 2-fold at both timepoints, and the presence of nevanimibe caused similar inhibition at 300 nM (approximately 20\%) for both timepoints.

\subparagraph{STAT3 protein expression under SOAT1 induction}

\begin{figure}
    \centering
    \includegraphics[width=1\linewidth]{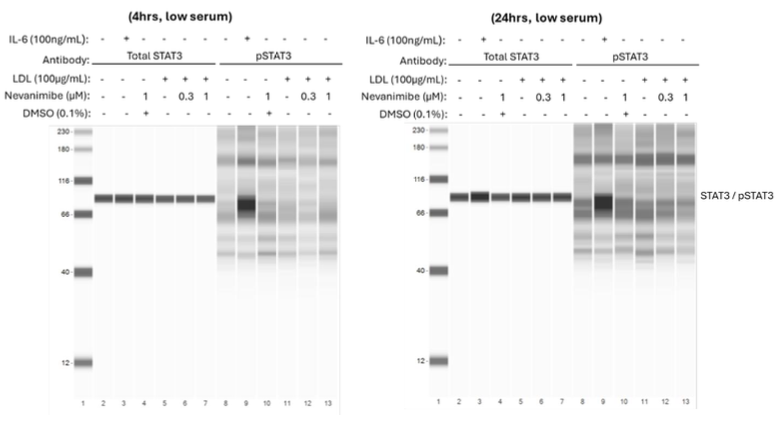}
    \caption{Jess (Simple Western) of STAT3/pSTAT3 under induction of SOAT1. pSTAT3 induction in IL-6 or LDL treated HepG2 lysates at 4 and 24 hours under low serum conditions. HepG2 cells were pre-treated with nevanimibe (0, 0.3, 1 $\mu$M) and then stimulated with IL-6 (100 ng/mL) for 4 or 24 hours. Total STAT3 was detected at the expected molecular weight in all treatment conditions. The induction of pSTAT3 was only observed in the presence of IL-6 and not induced by any of the LDL treatment conditions.}
    \label{fig:soat1-stat3-jess}
\end{figure}

To assess STAT3/pSTAT3 protein levels, Jess (Simple Western) was performed using cell lysates generated with cells grown and treated in low serum conditions (1\%). For total STAT3, clear peaks were detected at the expected molecular weight (approximately 90 kDa) in all samples at both 4 and 24 hours, with peak areas ranging from 260,000--500,000. In contrast, pSTAT3 only showed a peak with the positive control (IL-6 stimulation) at both timepoints, which was at the expected molecular weight (79 kDa). Peak area was approximately 1,000,000, representing a 3.2- and 2.7-fold increase over the corresponding basal control for 4 hours and 24 hours respectively. No clear peaks were observed with any of the other treatment conditions (Figure~\ref{fig:soat1-stat3-jess}).

\subparagraph{STAT3 mRNA expression under SOAT1 induction}

\begin{figure}
    \centering
    \includegraphics[width=0.75\linewidth]{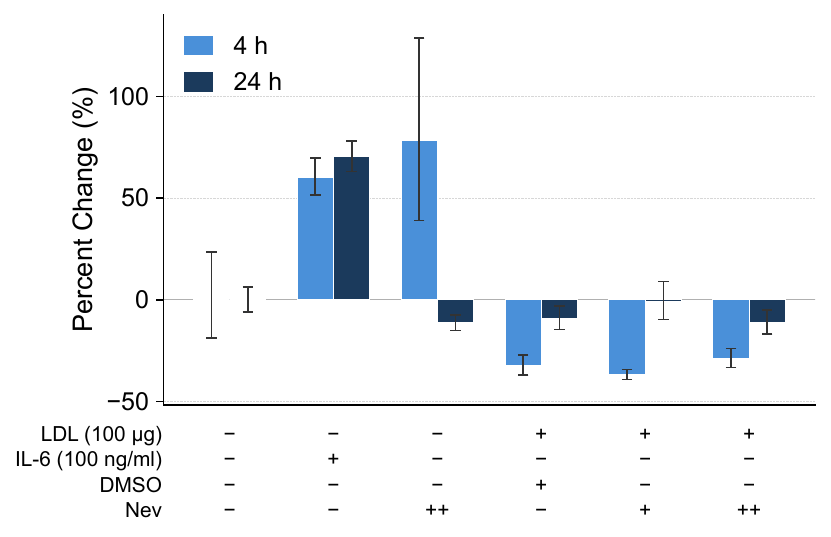}
    \caption{RT-qPCR of STAT3 under SOAT1 induction. Time-dependent STAT3 expression changes in response to LDL and nevanimibe treatment in HepG2 cells (4 and 24 hours, 1\% serum). STAT3 expression levels were compared as a percent change relative to the vehicle control baseline (medium only). IL-6 (100 ng/mL) was included as a positive control for STAT3 induction and showed increased expression at both timepoints. Treatment with LDL failed to significantly alter STAT3 expression levels at either timepoint. Nev: + (300 nM), ++ (1000 nM)}
    \label{fig:soat1-stat3-qpcr}
\end{figure}

To assess any effects of SOAT1 activation on STAT3 at the gene level, STAT3 mRNA expression was evaluated by RT-qPCR. Primer/probe sets for ACTB (selected housekeeping gene) and STAT3 targets were validated using RNA reference standards. Control HepG2 lysate samples exhibited robust expression of ACTB and STAT3, with minimal variation between replicates. Exposure to IL-6 (100 ng/mL) induced an increase in normalized STAT3 expression, with a percentage change of STAT3/ACTB of approximately 60\% at 4 hours and 70\% at 24 hours. Treatment with LDL at both concentrations and timepoints failed to significantly alter STAT3 expression levels (Figure~\ref{fig:soat1-stat3-qpcr}).

The results from the hypothesis testing assays demonstrated that stimulation of SOAT1 (via LDL loading of HepG2 cells) failed to demonstrate an effect on STAT3 at the transcription level (no significant change in STAT3 mRNA detected by RT-qPCR) or STAT3 at the protein level (measured by STAT3/pSTAT3 by Jess). In summary, these findings failed to support the hypothesis of a potential link between SOAT1 activity and STAT3 expression or activation (phosphorylation), at least within the current experimental design with HepG2 cells \cite{bairos2024}.

\paragraph{RAF1-TNF hypothesis}
The hypothesis that RAF1 modulation affects TNF expression was tested in THP-1 cells. RAF1 activity was assessed indirectly using an assay that measured ERK1/2 phosphorylation (pERK1/2), as ERK activation occurs downstream of RAF1 in the MAPK pathway \cite{molina2006}. TNF mRNA expression relative to ACTB was quantified by RT-qPCR, with lipopolysaccharide (LPS) used as a positive control to induce TNF expression and secretion. All experiments were conducted under low serum conditions (1\% FBS) to maintain cell viability throughout the assays.

\begin{figure}[h!]
\centering
    \includegraphics[width=0.75\linewidth]{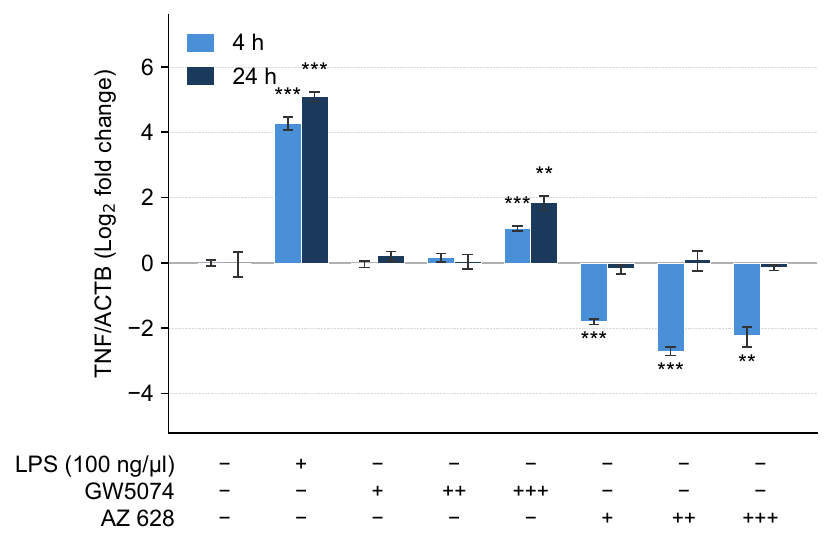}
\caption{TNF expression in THP-1 cells following treatment with RAF1 inhibitors GW5074 and AZ 628. THP-1 cells (8 $\times$ 10$^4$ per well) were treated for 4 hours or 24 hours with 0.2\% DMSO (vehicle control), 100 ng/mL LPS (positive control), or the indicated concentrations of GW5074 (+:10 nM, ++: 100 nM, +++: 1000 nM) or AZ 628 (+:30 nM, ++: 300 nM, +++: 3000 nM) RAF1 inhibitors. TNF expression was normalized to ACTB and presented as Log$_2$ fold change. Each condition represents n=4 biological replicates; bars represent mean $\pm$ SEM. Statistically significant differences compared to the vehicle control are indicated (*p $\leq$ 0.05, **p $\leq$ 0.01, ***p $\leq$ 0.001).}
\label{fig:raf1-tnf-stage3}
\end{figure}

The TNF expression was analyzed following treatment with GW5074, a selective inhibitor of RAF1 \cite{tocris2016}, and AZ 628, a pan-RAF kinase inhibitor \cite{medchemexpress2025a}. Exposure to 100 ng/mL LPS (positive control) caused a significant increase in normalized TNF expression, with a Log$_2$ fold change in TNF/ACTB of 4.2 at 4 hours and 5.0 at 24 hours. Treatment with GW5074 at a concentration of 1000 nM caused a smaller, but significant increase in TNF expression at both 4 hour and 24 hour timepoints, with a Log$_2$ fold change of 1.1 and 1.8, respectively. In contrast, treatment with AZ 628 caused a decrease in TNF expression at the 4 hour timepoint (range: $-1.8$ to $-2.7$) which had normalized relative to the baseline control at 24 hours (Figure~\ref{fig:raf1-tnf-stage3}).

Conditioned supernatants from the same assay plates were analyzed for any changes in secreted TNF protein levels. Consistent with the mRNA data, stimulation of the THP-1 cells with LPS caused a robust increase in TNF (658 pg/mL). However, the two RAF1 inhibitors did not impact TNF secretion, with all readings below the lower limit of quantification (LLOQ = 15.6 pg/mL).

The findings with GW5074 (selective RAF1 inhibitor) supported the proposed hypothesis of a functional relationship between RAF1 modulation and TNF. However, given that the changes in TNF gene expression were only observed at a single GW5074 concentration and were not mirrored at the (secreted) protein level, a confirmatory assay was conducted.

\begin{figure}[h!]
\centering
    \includegraphics[width=0.75\linewidth]{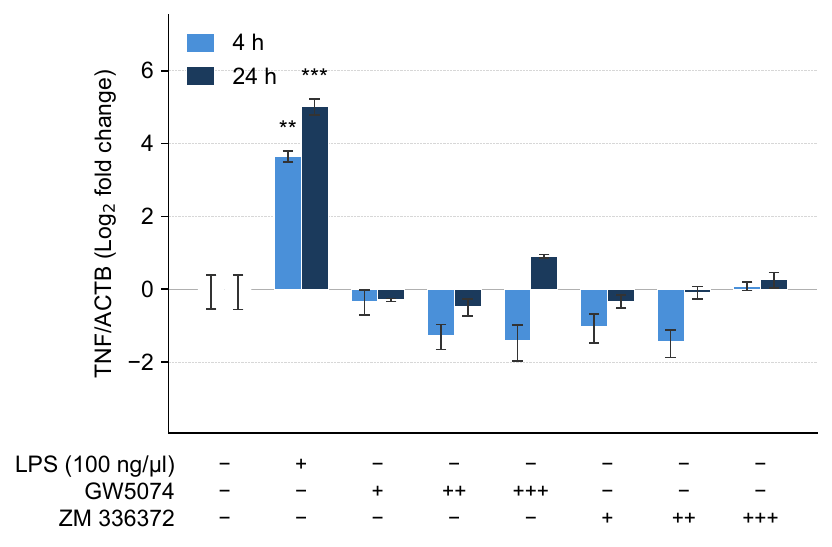}
\caption{TNF expression in THP-1 cells following GW5074 and ZM336372 treatment (confirmatory assay). THP-1 cells (8 $\times$ 10$^4$ per well) were treated for 4 hours or 24 hours with 0.2\% DMSO (vehicle control), 100 ng/mL LPS (positive control), or the indicated concentrations of GW5074  (+:10 nM, ++: 100 nM, +++: 1000 nM) or ZM336372  (+:100 nM, ++: 300 nM, +++: 1000 nM) RAF1 inhibitors. Consistent with the initial Stage 3 assay, a modest increase in normalized TNF expression was observed at 24 hours after treatment with 1000 nM GW5074. Each condition represents n=4 biological replicates; bars represent mean $\pm$ SEM. Statistically significant differences compared to the vehicle control are indicated (*p $\leq$ 0.05, **p $\leq$ 0.01, ***p $\leq$ 0.001).}
\label{fig:raf1-confirmatory}
\end{figure}

In the confirmatory assay, TNF expression was evaluated following the same assay design, but replacing AZ 628 with an alternative RAF1 inhibitor, ZM336372, reported to have greater selectivity for RAF1 \cite{medchemexpress2025b}. Exposure to 100 ng/mL LPS (positive control) resulted in increased normalized TNF expression at both timepoints (Log$_2$ fold change of 3.7 and 5.0 at 4 hours and 24 hours, respectively). Consistent with Stage 3 hypothesis assay data, a modest increase (Log$_2$ fold change of 0.9) in normalized TNF expression was observed at 24 hours after treatment with 1000 nM GW5074, with no change in secreted TNF levels. Analysis of treatment-matched lysates using Jess confirmed that 1000 nM GW5074 was causing an increase in phosphorylated ERK1/2 protein. Interestingly, a marked increase in pERK1/2 protein levels was also observed with all concentrations of ZM336372 (Figure~\ref{fig:raf1-confirmatory}).

The increase in ERK phosphorylation upon RAF1 inhibition is consistent with the phenomenon of paradoxical ERK activation, which occurs when inhibitor binding promotes RAF1 dimerization and transactivation, leading to downstream MEK and ERK activation \cite{callahan2014}. These findings support the hypothesis of a potential link between RAF1 signaling and TNF expression. The consistent increase in TNF mRNA expression with GW5074 treatment (1000 nM) at 24 hours across both assays, coinciding with increased ERK phosphorylation, suggests a functional relationship between RAF1 modulation and TNF regulation \cite{parameswaran2010}.

\paragraph{TP53-BAMBI hypothesis}
The hypothesis that TP53 affects BAMBI expression was tested in HepG2 cells. Nutlin-3, a small-molecule inhibitor of MDM2, was used to induce P53 protein levels, and downstream effects on BAMBI mRNA expression were assessed. TGF$\beta$1 was included as a positive control, as BAMBI functions as a non-signaling pseudoreceptor for TGF$\beta$1 \cite{sekiya2004}.

\begin{figure}[h!]
\centering
\includegraphics[width=0.75\linewidth]{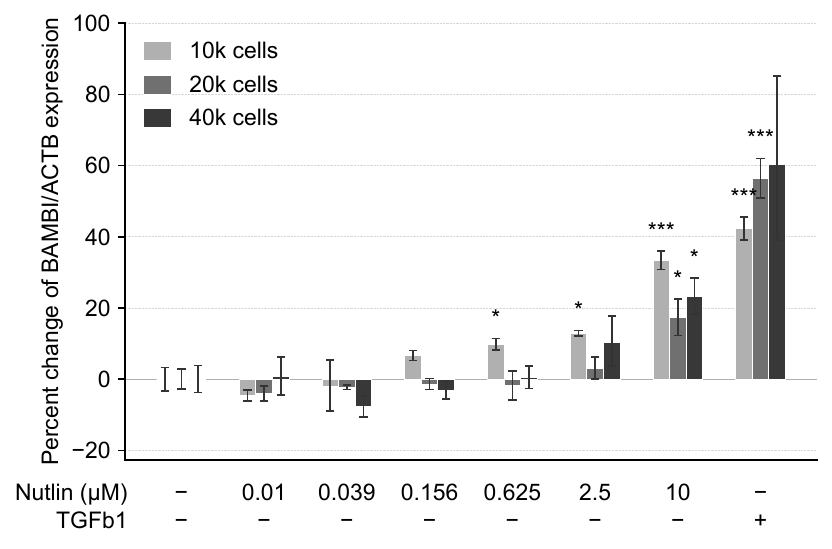}
\caption{BAMBI expression changes in response to Nutlin-3 treatment at 24 hours. HepG2 cells were treated with the indicated concentrations of Nutlin-3 for 24 hours. Expression was quantified using the $\Delta\Delta$Ct method and normalized to ACTB. Data are presented as percentage change relative to the vehicle control baseline (0.2\% DMSO). Three initial seeding densities (1, 2, or 4 $\times$ 10$^4$ cells per well) are indicated by bar colors. Each condition represents n=3--4 biological replicates; bars show mean $\pm$ standard error. TGF$\beta$1 (10 ng/mL) was included as a positive control for BAMBI induction. Statistically significant changes compared to the vehicle control are indicated (*p $\leq$ 0.05, **p $\leq$ 0.01, ***p $\leq$ 0.001).}
\label{fig:tp53-bambi-stage3}
\end{figure}

In the initial Stage 3 hypothesis test, BAMBI expression was assessed following Nutlin-3 treatment (0.01--10 $\mu$M) at a single 24 hour timepoint. Exposure to 10 ng/mL TGF$\beta$1 (positive control) resulted in an increase in normalized BAMBI expression at all cell densities (42\%, 57\%, and 65\% at seeding densities of 1 $\times$ 10$^4$, 2 $\times$ 10$^4$, and 4 $\times$ 10$^4$ cells, respectively). An increase in normalized BAMBI expression was also observed with Nutlin-3 treatment. The greatest increase in BAMBI expression occurred with 10 $\mu$M Nutlin-3, with relative changes of 33\%, 17\%, and 23\% observed at seeding densities of 1 $\times$ 10$^4$, 2 $\times$ 10$^4$, and 4 $\times$ 10$^4$ cells, respectively (Figure~\ref{fig:tp53-bambi-stage3}).

Analysis of P53 protein expression in treatment-matched lysates confirmed induction by Nutlin-3, with the highest increase (8.7-fold increase in normalized peak area) also observed at 10 $\mu$M. These findings supported the proposed hypothesis of a functional relationship between P53 and BAMBI. However, given that the changes in BAMBI expression were modest (17--33\%), a Stage 3 confirmatory assay was conducted to further evaluate BAMBI expression dynamics over a time-course of Nutlin-3 treatment.

\begin{figure}[h!]
\centering
\includegraphics[width=0.75\linewidth]{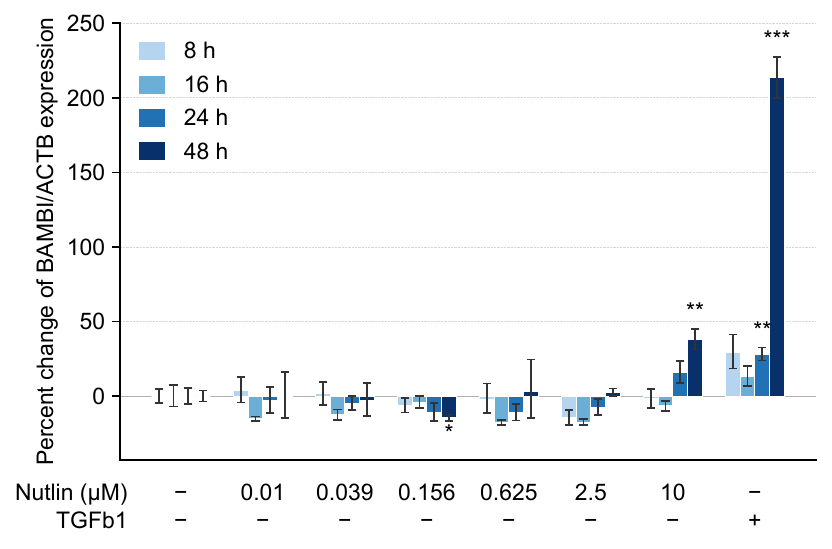}
\caption{Time-dependent BAMBI expression changes with Nutlin-3 treatment. HepG2 cells were treated with the indicated concentrations of Nutlin-3 for 8, 16, 24, or 48 hours. BAMBI expression was quantified using the $\Delta\Delta$Ct method and normalized to ACTB. Data are shown as percent change relative to the vehicle control baseline (0.2\% DMSO). TGF$\beta$1 (10 ng/mL) was included as a positive control for BAMBI induction. Each condition represents n=4 biological replicates; bars show mean $\pm$ standard error. The 48-hour TGF$\beta$1 treatment exceeded the y-axis range, with a mean percent change of 214\% ($\pm$14\%). Statistically significant changes compared to the vehicle control are indicated (*p $\leq$ 0.05, **p $\leq$ 0.01, ***p $\leq$ 0.001).}
\label{fig:tp53-bambi-timecourse}
\end{figure}

In the confirmatory assay, BAMBI expression was evaluated following Nutlin-3 treatment over a time-course with sampling at 8, 16, 24, and 48 hours. Exposure to 10 ng/mL TGF$\beta$1 (positive control) resulted in increased normalized BAMBI expression at all timepoints (30\%, 13\%, 28\%, and 213\% at 8 hours, 16 hours, 24 hours, and 48 hours, respectively). Consistent with Stage 3 hypothesis assay data, a modest increase (16\%) in normalized BAMBI expression was observed at 24 hours after treatment with 10 $\mu$M Nutlin-3. This effect became more pronounced at 48 hours, where normalized BAMBI expression increased by 38\% compared to the vehicle control (Figure~\ref{fig:tp53-bambi-timecourse}).

Analysis of treatment-matched lysates using Jess confirmed robust induction of P53 protein, with the highest induction (11-fold increase) occurring with 10 $\mu$M Nutlin-3. The time-course analysis also revealed modest decreases in BAMBI expression of up to 18\% relative to baseline when cells were exposed to lower Nutlin-3 concentrations (0.156--2.5 $\mu$M).

The consistent increase in BAMBI expression with 10 $\mu$M Nutlin-3 treatment at 24 hours across both assays, coinciding with maximal P53 protein induction, supports the hypothesis of a functional relationship between P53 and BAMBI \cite{solyakov2009}. The time-dependent increase in BAMBI expression, becoming more pronounced at 48 hours, suggests that the effect may be mediated through transcriptional regulation. However, the modest magnitude of the changes (16--38\%) and the observation of decreased BAMBI expression at lower Nutlin-3 concentrations suggest a subtle and context-dependent relationship that may require further investigation to fully elucidate.
\subsection{Experimental protocols}

For each selected hypothesis, we designed human cell line experiments to test the hypothesis.

\subsubsection{Equipment}

All equipment used for experimental validation is listed in Table~\ref{tab:equipment}.

\begin{table}[h!]
\centering
\caption{List of equipment used for wet-lab validation.}
\label{tab:equipment}
\begin{tabular}{ll}
\toprule
\textbf{Equipment/Software Name} & \textbf{Cat No./ID No./Serial No./Version} \\
\midrule
Vac-Man 96 Vacuum Manifold & A2291 \\
MiniAmp Thermal Cycler & A37834 \\
Quantstudio 6 Flex Real-Time PCR System & 4485689 \\
Agilent 50 TS Plate Washer & S/N 2302141D \\
BioTek Cytation5 & S/N 21041316 \\
Gen5 software & 3.13.15 \\
Jess Simple Western system & JS4787 \\
Compass for Simple Western software & 7.0.0 \\
\bottomrule
\end{tabular}
\end{table}

\subsubsection{Cell Lines}

All cell lines used for experimental validation are shown in Table~\ref{tab:celllines}.

\begin{table}[h!]
\centering
\caption{List of cell lines used for wet-lab validation.}
\label{tab:celllines}
\small
\begin{tabular}{p{2cm}p{1.5cm}p{1cm}p{1.5cm}p{2cm}p{2cm}p{2.5cm}}
\toprule
\textbf{Hypothesis} & \textbf{Cell Line} & \textbf{Vendor} & \textbf{Catalog No.} & \textbf{Morphology} & \textbf{Tissue} & \textbf{Disease} \\
\midrule
SOAT1-STAT3 & HepG2 & ATCC & HB-8065 & Epithelial-like & Liver & Hepatocellular carcinoma \\
TP53-BAMBI & HepG2 & ATCC & HB-8065 & Epithelial-like & Liver & Hepatocellular carcinoma \\
RAF1-TNF & THP-1 & ATCC & TIB-202 & Monocyte & Peripheral blood & Acute monocytic leukemia \\
\bottomrule
\end{tabular}
\end{table}

\subsubsection{Cell Culture Reagents}

Cell culture reagents for the SOAT1-STAT3 hypothesis validation are listed in Table~\ref{tab:reagents-soat1}.

\begin{table}[p!]
\centering
\caption{List of reagents used for SOAT1-STAT3 hypothesis validation.}
\label{tab:reagents-soat1}
\scriptsize
\begin{tabular}{p{2cm}p{6cm}p{3.5cm}p{2cm}}
\toprule
\textbf{Category} & \textbf{Reagent Description} & \textbf{Vendor} & \textbf{Catalog No.} \\
\midrule
Cell culture & Eagle's minimum essential medium (EMEM) & ATCC & 30-2003 \\
Cell culture & FBS & ATCC & 30-2020 \\
Cell culture & Ca$^{2+}$/Mg$^{2+}$ free HBSS & Merck (Sigma Aldrich) & H6648 \\
Cell culture & Ca$^{2+}$/Mg$^{2+}$ free PBS & Merck (Sigma Aldrich) & D8537 \\
Cell culture & Penicillin-Streptomycin (100x) & Merck (Sigma Aldrich) & P4333 \\
Cell culture & Recombinant Human IL-6 & Fisher Scientific & 200-06-50UG \\
Cell culture & RIPA Lysis and Extraction Buffer & ThermoFisher & 89900 \\
Cell culture & Halt Phosphatase Inhibitor Single-Use Cocktail & ThermoFisher & 78420 \\
Cell culture & Halt Protease Inhibitor Cocktail (100X) & ThermoFisher & 87786 \\
Cell culture & BCA protein assay kit & ThermoFisher (Pierce) & 23225 \\
Cell culture & Human LDL & Kalen Biomedical & 770200-4 \\
Cell culture & Cholesterol/Cholesterol Ester-Glo Assay & Promega & J3190 \\
Cell culture & Nevanimibe Hydrochloride & MedChemExpress & HY-10399A \\
Antibody & Human Phospho-STAT3(Y705) & Bio-Techne (Novus biologicals) & AF4607 \\
Antibody & Stat3 (D1B2J) & Cell Signalling Technologies & 30835 \\
Antibody & ERK1 primary antibody (positive control) & Bio-Techne (ProteinSimple) & 042-486 \\
Antibody & Anti-Mouse Secondary HRP antibody & Bio-Techne (ProteinSimple) & 042-205 \\
Antibody & Anti-Rabbit Secondary HRP antibody & Bio-Techne (ProteinSimple) & 042-206 \\
Antibody & Streptavidin-HRP (for detection of ladder) & Bio-Techne (ProteinSimple) & 042-414 \\
Jess & 12-230 KDa Separation Module & Bio-Techne (ProteinSimple) & SM-W004 \\
Jess & 12-230 KDa pre-filled microplate & Bio-Techne (ProteinSimple) & 30-2020 \\
Jess & 10x sample buffer & Bio-Techne (ProteinSimple) & 042-195 \\
Jess & Biotinylated ladder (12-230 KDa) & Bio-Techne (ProteinSimple) & PS-ST01EZ-8 \\
Jess & DTT & Bio-Techne (ProteinSimple) & PS-ST01EZ-8 \\
Jess & Fluorescent 5X Master Mix & Bio-Techne (ProteinSimple) & PS-ST01EZ-8 \\
Jess & Hela lysate (assay control) & Bio-Techne (ProteinSimple) & 042-488 \\
Jess & Luminol & Bio-Techne (ProteinSimple) & 043-311 \\
Jess & Peroxide & Bio-Techne (ProteinSimple) & 043-379 \\
Jess & Antibody Diluent 2 & Bio-Techne (ProteinSimple) & 042-203 \\
Jess & Wash buffer & Bio-Techne (ProteinSimple) & 042-202 \\
Genomics & SV 96 Total RNA Isolation System & Promega & Z3500 \\
Genomics & DNase 1 & ThermoFisher & 18068015 \\
Genomics & High-Capacity cDNA Reverse Transcription Kit & ThermoFisher & 4368813 \\
Genomics & TaqMan Universal PCR Master & ThermoFisher (Applied Biosystems) & 4304437 \\
Genomics & ACTB ($\beta$-actin) Primer/Probe set & ThermoFisher & Hs01060665\_g1 \\
Genomics & STAT3 Primer/Probe set & ThermoFisher & Hs00374280\_m1 \\
Genomics & Universal Human Reference RNA & ThermoFisher & QS0639 \\
Genomics & TE Buffer (1x TE solution, pH 8.0) & IDT & 11-05-01-13 \\
\bottomrule
\end{tabular}
\end{table}

Cell culture reagents for the TP53-BAMBI hypothesis validation are listed in Table~\ref{tab:reagents-tp53}.

\begin{table}[p!]
\centering
\caption{List of reagents used for TP53-BAMBI hypothesis validation.}
\label{tab:reagents-tp53}
\scriptsize
\begin{tabular}{p{2cm}p{6cm}p{3.5cm}p{2cm}}
\toprule
\textbf{Category} & \textbf{Reagent Description} & \textbf{Vendor} & \textbf{Catalog No.} \\
\midrule
Cell culture & Eagle's minimum essential medium (EMEM) & ATCC & 30-2003 \\
Cell culture & FBS & ATCC & 30-2020 \\
Cell culture & Ca$^{2+}$/Mg$^{2+}$ free HBSS & Merck (Sigma Aldrich) & H6648 \\
Cell culture & Ca$^{2+}$/Mg$^{2+}$ free PBS & Merck (Sigma Aldrich) & D8537 \\
Cell culture & Penicillin-Streptomycin (100x) & Merck (Sigma Aldrich) & P4333 \\
Cell culture & Transforming growth factor-$\beta$1 & R\&D systems & 240-B-002 \\
Cell culture & Nutlin-3 & Bio-Techne & 3984 \\
Cell culture & RIPA Lysis and Extraction Buffer & ThermoFisher & 89900 \\
Cell culture & Halt Phosphatase Inhibitor Single-Use Cocktail & ThermoFisher & 78420 \\
Cell culture & Halt Protease Inhibitor Cocktail (100X) & ThermoFisher & 87786 \\
Cell culture & BCA protein assay kit & ThermoFisher (Pierce) & 23225 \\
Antibody & P53 antibody & Bio-Techne (Novus biologicals) & NBP2-29453 \\
Antibody & ERK1 primary antibody (positive control) & Bio-Techne (ProteinSimple) & 042-486 \\
Antibody & Anti-Mouse Secondary HRP antibody & Bio-Techne (ProteinSimple) & 042-205 \\
Antibody & Anti-Rabbit Secondary HRP antibody & Bio-Techne (ProteinSimple) & 042-206 \\
Antibody & Streptavidin-HRP (for detection of ladder) & Bio-Techne (ProteinSimple) & 042-414 \\
Jess & 12-230 KDa Separation Module & Bio-Techne (ProteinSimple) & SM-W004 \\
Jess & 12-230 KDa pre-filled microplate & Bio-Techne (ProteinSimple) & 30-2020 \\
Jess & 10x sample buffer & Bio-Techne (ProteinSimple) & 042-195 \\
Jess & Biotinylated ladder (12-230 KDa) & Bio-Techne (ProteinSimple) & PS-ST01EZ-8 \\
Jess & DTT & Bio-Techne (ProteinSimple) & PS-ST01EZ-8 \\
Jess & Fluorescent 5X Master Mix & Bio-Techne (ProteinSimple) & PS-ST01EZ-8 \\
Jess & Hela lysate (assay control) & Bio-Techne (ProteinSimple) & 042-488 \\
Jess & Luminol & Bio-Techne (ProteinSimple) & 043-311 \\
Jess & Peroxide & Bio-Techne (ProteinSimple) & 043-379 \\
Jess & Antibody Diluent 2 & Bio-Techne (ProteinSimple) & 042-203 \\
Jess & Wash buffer & Bio-Techne (ProteinSimple) & 042-202 \\
Jess & Protein normalisation module & Bio-Techne (ProteinSimple) & DM-PN02 \\
Genomics & SV 96 Total RNA Isolation System & Promega & Z3500 \\
Genomics & DNase 1 & ThermoFisher & 18068015 \\
Genomics & High-Capacity cDNA Reverse Transcription Kit & ThermoFisher & 4368813 \\
Genomics & TaqMan Universal PCR Master & ThermoFisher (Applied Biosystems) & 4304437 \\
Genomics & ACTB ($\beta$-actin) Primer/Probe set & ThermoFisher & Hs01060665\_g1 \\
Genomics & BAMBI Primer/Probe set & ThermoFisher & Hs03044164\_m1 \\
Genomics & Universal Human Reference RNA & ThermoFisher & QS0639 \\
Genomics & TE Buffer (1x TE solution, pH 8.0) & IDT & 11-05-01-13 \\
\bottomrule
\end{tabular}
\end{table}

Cell culture reagents for the RAF1-TNF hypothesis validation are listed in Table~\ref{tab:reagents-raf1}.

\begin{table}[h!]
\centering
\caption{List of reagents used for RAF1-TNF hypothesis validation.}
\label{tab:reagents-raf1}
\scriptsize
\begin{tabular}{p{2cm}p{6cm}p{3.5cm}p{2cm}}
\toprule
\textbf{Category} & \textbf{Reagent Description} & \textbf{Vendor} & \textbf{Catalog No.} \\
\midrule
Cell culture & RPMI-1640 medium & ATCC & 30-2001 \\
Cell culture & FBS & ATCC & 30-2020 \\
Cell culture & 2-Mercaptoethanol & Fisher & 11508916 \\
Cell culture & Ca$^{2+}$/Mg$^{2+}$ free HBSS & Merck (Sigma Aldrich) & H6648 \\
Cell culture & Penicillin-Streptomycin (100x) & Merck (Sigma Aldrich) & P4333 \\
Cell culture & Lipopolysaccharides from \textit{Escherichia coli} O127:B8 (LPS) & Merck & L4516 \\
Cell culture & GW5074 & Biotechne & 1381 \\
Cell culture & AZ 628 & Biotechne & 4836 \\
Cell culture & ZM336372 & Fisher & HY13343 \\
Cell culture & RIPA Lysis and Extraction Buffer & ThermoFisher & 89900 \\
Cell culture & Halt Phosphatase Inhibitor Single-Use Cocktail & ThermoFisher & 78420 \\
Cell culture & Halt Protease Inhibitor Cocktail (100X) & ThermoFisher & 87786 \\
Cell culture & BCA protein assay kit & ThermoFisher (Pierce) & 23225 \\
Genomics & SV 96 Total RNA Isolation System & Promega & Z3500 \\
Genomics & DNase 1 & ThermoFisher & 18068015 \\
Genomics & High-Capacity cDNA Reverse Transcription Kit & ThermoFisher & 4368813 \\
Genomics & TaqMan Universal PCR Master & ThermoFisher (Applied Biosystems) & 4304437 \\
Genomics & ACTB ($\beta$-actin) Primer/Probe set & ThermoFisher & Hs01060665\_g1 \\
Genomics & TNF$\alpha$ Primer/Probe set & ThermoFisher & Hs00174128\_m1 \\
Genomics & Universal Human Reference RNA & ThermoFisher & QS0639 \\
Genomics & TE Buffer (1x TE solution, pH 8.0) & IDT & 11-05-01-13 \\
\bottomrule
\end{tabular}
\end{table}

\subsubsection{Methods}

\paragraph{SOAT1-STAT3 Hypothesis}

\subparagraph{Cell Culture}
HepG2 cells were cultured in Eagle's Minimum Essential Medium (EMEM) supplemented with 10\% FBS and 1\% Penicillin-Streptomycin (PS) according to ATCC recommendations.

\subparagraph{Compound Preparation}
Human LDL was stored at 2--8°C and further diluted as required in low-serum medium.

Stock solutions of nevanimibe (10mM) were prepared in 100\% DMSO on the first day of use and then stored at -20°C for subsequent assays. Stocks were further diluted in 100\% DMSO and low serum medium.

Recombinant IL-6 was reconstituted at 500$\mu$g/mL in sterile water, aliquoted and stored at -20°C. On the day of the assay, an aliquot was thawed and diluted as required in low-serum medium.

\subparagraph{Hypothesis Test Assay}
HepG2 cells were plated in 96-well plates (CE bioassay, RT-qPCR) at a single cell density (4 $\times$ 10$^4$ cells per well) or 6-well plates (Jess) (1 $\times$ 10$^6$ cells per well) in low-serum medium (EMEM supplemented with 1\% FBS and 1\% PS) and incubated (37°C and 5\% CO$_2$) overnight.

Cells were treated with LDL (100$\mu$g/mL) $\pm$ nevanimibe (300, 1000nM), IL-6 (100ng/mL), vehicle control (0.1\% DMSO), or medium only for 4 or 24 hours (37°C and 5\% CO$_2$).

\subparagraph{Cholesterol/Cholesterol Ester-Glo Assay}
The production of cholesterol and cholesterol ester (CE) was measured using a Cholesterol/CE-Glo kit (Promega) according to the manufacturer's standard protocol. For measuring CE, the assay includes a cholesterol esterase that removes the fatty acid from cholesterol esters to produce a single molecule of cholesterol per molecule of ester. The amount of CE was determined from the difference of cholesterol measured in the absence (free cholesterol) and presence (total cholesterol) of esterase.

Samples were analyzed neat or at a single dilution (1:2) in duplicate or triplicate wells for total and free cholesterol.

The protocol used is summarized as follows:
\begin{enumerate}
\item Medium removed and cells washed twice with 100$\mu$L PBS
\item Cholesterol lysis solution (50$\mu$L) added per well. Plate shaken briefly and incubated for 30 minutes at 37°C
\item A cholesterol standard curve was prepared using the kit provided 20mM cholesterol standard (0$\mu$M--80$\mu$M)
\item If needed, samples were diluted 1:2 with cholesterol lysis solution
\item Samples, standards or background controls (50$\mu$L) transferred to a 96-well white-walled plate
\item Cholesterol detection reagent (50$\mu$L) with or without esterase added to all wells. Plate shaken briefly for 30--60 seconds at a low rpm on a plate shaker and then incubated at room temperature for 1 hour
\item Luminescence measured using a plate reader (Cytation5)
\item Free and total cholesterol was calculated by comparison of the luminescence of samples and standards assayed under the same conditions
\end{enumerate}

\paragraph{TP53-BAMBI Hypothesis}

\subparagraph{Cell Culture}
HepG2 cells were cultured in Eagle's Minimum Essential Medium (EMEM) supplemented with 10\% FBS and 1\% Penicillin-Streptomycin (PS) according to ATCC recommendations.

\subparagraph{Compound Preparation}
Transforming growth factor-beta1 (TGF$\beta$1) was reconstituted at 20$\mu$g/mL in 4mM HCl + 1mg/mL BSA, aliquoted and stored at -20°C. On the day of the assay, an aliquot was thawed and diluted as required in low-serum medium.

A 10mM stock of Nutlin-3 in 100\% DMSO was prepared fresh on the day of each assay and diluted as required in low serum medium.

\subparagraph{Hypothesis Test Assay}
HepG2 cells were plated in 96-well plates (RT-qPCR) (4 $\times$ 10$^4$ cells per well) or 6-well plates (Jess) (1 $\times$ 10$^6$ cells per well) in low-serum medium (EMEM supplemented with 1\% FBS and 1\% PS) and incubated (37°C and 5\% CO$_2$) overnight.

Cells were then treated with Nutlin-3 (0.01--10$\mu$M), TGF$\beta$1 (10ng/mL), vehicle control (0.2\% DMSO), or medium only (6-well plates only).

The 96-well plates were incubated for 8, 16, 24 or 48 hours, and the 6-well plates for 24 hours (37°C and 5\% CO$_2$).

\paragraph{RAF1-TNF Hypothesis}

\subparagraph{Cell Culture}
THP1 cells were cultured in RPMI-1640 medium supplemented with 10\% FBS and 1\% Penicillin-Streptomycin (PS) and 0.05mM beta-mercaptoethanol (BME) according to ATCC recommendations.

\subparagraph{Compound Preparation}
Lipopolysaccharide (LPS) was reconstituted at 1mg/mL in sterile PBS, aliquoted and stored at -20°C. On the day of the assay, an aliquot was thawed and diluted as required in low-serum medium.

Stock solutions of GW5074 (10mM), ZM336372 (30mM), and AZ628 (10mM) were prepared in 100\% DMSO fresh on the day of each assay and diluted as required in low serum medium.

\subparagraph{Hypothesis Test Assay}
THP-1 were plated in 96-well plates (RT-qPCR) (8 $\times$ 10$^4$ cells per well) or transferred to 15mL falcon tubes (Jess) (4 $\times$ 10$^6$ cells per tube) in low-serum medium (RPMI-1640 + 1\% FBS + 1\% PS + 0.05mM BME) and allowed to equilibrate for 4 hours (37°C and 5\% CO$_2$).

Cells were treated with GW5074 (10, 100, or 1000nM), ZM336372 (100, 300, or 1000nM), LPS (100ng/mL), or vehicle control (0.2\% DMSO).

The 96-well plates were incubated for 4 or 24 hours, and the tubes for 30 minutes (37°C and 5\% CO$_2$).

\paragraph{Common Assay Methods}

\subparagraph{Lysate Preparation}
Cells were washed twice with ice-cold PBS (0.5mL per well for 6-well plates, 1mL per well for THP-1 experiments).

Cold RIPA buffer (100$\mu$L per well) supplemented with Halt protease and phosphatase inhibitors was then added and the plates kept on ice for 10 minutes (HepG2) or 15 minutes (THP-1).

The lysed cells were transferred to a microcentrifuge tube and centrifuged at 14,000 $\times$ g for 10 minutes at 4°C to pellet cell debris.

The supernatant was then transferred to a new tube and stored at -80°C pending analysis.

\subparagraph{BCA Assay}
Total protein concentration was determined using the Pierce\textsuperscript{TM} BCA protein assay kit and the microplate procedure.

All samples were analyzed at a single (1:10) dilution in triplicate wells.

Absorbance was measured at 562nm using a BioTek Cytation5 cell imaging multi-mode reader and blank-corrected absorbance measurements of the unknowns interpolated from the standard curve.

\subparagraph{Jess Capillary Electrophoresis}

All lysate samples were analyzed on 12-230 KDa Separation Modules under denaturing conditions with chemiluminescence detection.

\textit{SOAT1-STAT3 Hypothesis:} The expression of STAT3 and pSTAT3 (Y705) was determined by capillary electrophoresis (Bio-techne Simple Western, Jess). An optimal protein concentration (total STAT3 = 0.08mg/mL, pSTAT3 = 2mg/mL) and antibody concentration (total STAT3 = 1.4$\mu$g/mL; pSTAT3 = 20$\mu$g/mL) were used.

\textit{TP53-BAMBI Hypothesis:} The expression of P53 was determined by capillary electrophoresis (Bio-techne Simple Western, Jess). An optimal protein concentration (1.2mg/mL) and P53 antibody concentration (10$\mu$g/mL) were used.

Assay and antibody controls were included for all Jess runs (e.g., Biotinylated ladder, Hela lysate + ERK1 antibody, no primary, and no sample controls).

The Jess protocol used is summarized in brief below:
\begin{enumerate}
\item Prepare Standard Pack Reagents (DTT, Fluorescent 5X Master Mix, \& Biotinylated ladder)
\item Prepare cell lysate dilutions using 0.1X Sample Buffer
\item Add Fluorescent 5X Master Mix to each lysate sample
\item Prepare a no sample control (Fluorescent 5X Master Mix + 0.1X Sample Buffer only)
\item Prepare the antibody dilutions using Antibody Diluent 2
\item Denature the lysate samples @ 95°C for 5 minutes, vortex, spin, and store on ice
\item Stage 3 only (TP53-BAMBI)--prepare the protein normalization reagent (50$\mu$L stock + 250$\mu$L reconstitution agent)
\item Dispense the samples, reagents and wash buffer into the microplate
\item Centrifuge the plate for 5 minutes @ 1000 $\times$ g at RT
\item Load plate into the Jess and start assay run (primary antibody incubation time: 30 minutes)
\end{enumerate}

The peak areas of named peaks were calculated using the Jess Compass for Simple Western Software (peak fit = dropped lines (manual) and baseline type = spline). For Stage 3 assays with protein normalization (TP53-BAMBI), peak area data were corrected for total protein loading.

\subparagraph{Primer Probe Validation}

Primer/probe sets for ACTB ($\beta$-actin) and target genes (BAMBI for TP53-BAMBI, STAT3 for SOAT1-STAT3, TNF$\alpha$ for RAF1-TNF) were validated by RT-qPCR using 10 ng and 100 ng of Universal Human Reference RNA as input.

ACTB and TNF$\alpha$ (RAF1-TNF only) were further evaluated by qPCR analysis using in-house oligonucleotide standards corresponding to the ACTB and TNF$\alpha$ sequence regions shown below. A 10-fold serial dilution was prepared in 1$\times$ TE buffer (10 mM Tris, pH 8.0; 0.1 M EDTA), covering an input range from 10$^8$ copies to 10$^1$ copies.\\

\noindent ACTB oligonucleotide:

\noindent\small\texttt{CCTCGCCTTTGCCGATCCGCCGCCCGTCCACACCCGCCGCCAGCTCACCATGGATGATGAT}\\
\texttt{ATCGCCGCGCTCGTCGTCGACAACGGCTCCGGCATGTGCAAGGCCGGCTTCGCGGGCGACG}\\
\texttt{ATGCCCCCCGGGCCGTCTTCCCCTCCATCGTGGGGCGCCCCAGGCACCAGGGCGTGATGGT}\\
\texttt{GGGCATGGGTCAGAAG}\\

\noindent TNF$\alpha$ oligonucleotide:

\noindent\small\texttt{CTCGAACCCCGAGTGACAAGCCTGTAGCCCATGTTGTAGCAAACCCTCAAGCTGAGGGGCA}\\
\texttt{GCTCCAGTGGCTGAACCGCCGGGCCAATGCCCTCCTGGCCAATGGCGTGGAGCTGAGAGAT}\\
\texttt{AACCAGCTGGTGGTGCCATCAGAGGGCCTGTACCTCATCTACTCCCAGGTCCTCTTCAAGG}\\
\texttt{GCC}\\

\subparagraph{RNA Extraction}
Cells were washed (1$\times$ HBSS), lysed with RLA buffer + 1\% beta-mercaptoethanol (100$\mu$L per well) and the plate was stored at -80°C pending RNA extraction.

RNA was extracted according to the manufacturer's instructions using the Promega SV 96 Total RNA Isolation System in combination with a Vac-Man 96 Vacuum Manifold. Purified RNA was eluted in nuclease-free water (100$\mu$L).

\subparagraph{RT-qPCR}
Purified RNA was treated with DNase I according to the manufacturer's protocol and heat-inactivated at 65°C for 10 minutes using a MiniAmp Thermal Cycler.

Reverse transcription was performed using the High-Capacity cDNA Reverse Transcription Kit on a MiniAmp Thermal Cycler with the following program: 25°C for 10 minutes, 37°C for 2 hours, 85°C for 5 minutes, and 4°C for 10 minutes.

qPCR reactions were prepared using the TaqMan Universal PCR Master Mix with primer/probe sets for ACTB and BAMBI (TP53-BAMBI), STAT3 (SOAT1-STAT3), or TNF$\alpha$ (RAF1-TNF). Amplification and detection were performed on a QuantStudio 6 Flex Real-Time PCR System under the following thermocycling conditions: 50°C for 2 minutes, 95°C for 10 minutes; 40 cycles of: 95°C for 15 seconds and 60°C for 1 minute.

\textit{TP53-BAMBI Hypothesis:} Relative expression of BAMBI was calculated against the ACTB housekeeping gene using the $\Delta\Delta$Ct method. Outliers were identified using Grubbs' test at $\alpha$ = 0.05. A replicate was removed if its G statistic exceeded the critical value (1.15 for n = 3 or 1.48 for n = 4). Treatment conditions for Stage 3 qPCR assays were statistically compared using Student's t-test.

\textit{SOAT1-STAT3 Hypothesis:} Relative expression of STAT3 was calculated against the ACTB housekeeping gene using the $\Delta\Delta$Ct method. Outliers were identified using Grubbs' test at $\alpha$ = 0.05. A replicate was removed if its G statistic exceeded the critical value (1.15 for n = 3 or 1.48 for n = 4). Treatment conditions for Stage 3 qPCR assays were statistically compared using Student's t-test.

\textit{RAF1-TNF Hypothesis:} TNF$\alpha$ and ACTB transcript levels were quantified by RT-qPCR using reference standards for each target. Fold changes in TNF$\alpha$ expression normalized relative to ACTB were calculated for each condition. Outliers were identified using Grubbs' test at $\alpha$ = 0.05. A replicate was removed if its G statistic exceeded the critical value (1.15 for n = 3 or 1.48 for n = 4). Treatment conditions for qPCR assays were statistically compared using Student's t-test.




\end{appendices}


\bibliography{sn-bibliography}

@article{wei2024pubtator,
  title={PubTator 3.0: an AI-powered literature resource for unlocking biomedical knowledge},
  author={Wei, Chih-Hsuan and Allot, Alexis and Lai, Po-Ting and Leaman, Robert and Tian, Shubo and Luo, Ling and Jin, Qiao and Wang, Zhizheng and Chen, Qingyu and Lu, Zhiyong},
  journal={Nucleic Acids Research},
  volume={52},
  number={W1},
  pages={W540--W546},
  year={2024},
  publisher={Oxford University Press}
}

@article{jumper2021highly,
  title={Highly accurate protein structure prediction with AlphaFold},
  author={Jumper, John and Evans, Richard and Pritzel, Alexander and Green, Tim and Figurnov, Michael and Ronneberger, Olaf and Tunyasuvunakool, Kathryn and Bates, Russ and {\v{Z}}{\'\i}dek, Augustin and Potapenko, Anna and others},
  journal={nature},
  volume={596},
  number={7873},
  pages={583--589},
  year={2021},
  publisher={Nature Publishing Group UK London}
}

@article{merchant2023scaling,
  title={Scaling deep learning for materials discovery},
  author={Merchant, Amil and Batzner, Simon and Schoenholz, Samuel S and Aykol, Muratahan and Cheon, Gowoon and Cubuk, Ekin Dogus},
  journal={Nature},
  volume={624},
  number={7990},
  pages={80--85},
  year={2023},
  publisher={Nature Publishing Group UK London}
}

@inproceedings{zhou2024hypothesis,
  title={Hypothesis generation with large language models},
  author={Zhou, Yangqiaoyu and Liu, Haokun and Srivastava, Tejes and Mei, Hongyuan and Tan, Chenhao},
  booktitle={Proceedings of the 1st Workshop on NLP for Science (NLP4Science)},
  pages={117--139},
  year={2024}
}

@inproceedings{baek2025researchagent,
  title={Researchagent: Iterative research idea generation over scientific literature with large language models},
  author={Baek, Jinheon and Jauhar, Sujay Kumar and Cucerzan, Silviu and Hwang, Sung Ju},
  booktitle={Proceedings of the 2025 Conference of the Nations of the Americas Chapter of the Association for Computational Linguistics: Human Language Technologies (Volume 1: Long Papers)},
  pages={6709--6738},
  year={2025}
}

@article{lu2024ai,
  title={The AI scientist: Towards fully automated open-ended scientific discovery},
  author={Lu, Chris and Lu, Cong and Lange, Robert Tjarko and Foerster, Jakob and Clune, Jeff and Ha, David},
  journal={arXiv preprint arXiv:2408.06292},
  year={2024}
}

@article{longa2023GNNs,
  author={Antonio Longa and Veronica Lachi and Gabriele Santin and Monica Bianchini and Bruno Lepri and Pietro Lio and Franco Scarselli and Andrea Passerini},
  title={Graph Neural Networks for Temporal Graphs: State of the Art, Open Challenges, and Opportunities},
  year={2023},
  cdate={1672531200000},
  journal={Trans. Mach. Learn. Res.},
  volume={2023},
  url={https://openreview.net/forum?id=pHCdMat0gI}
}

@article{ADMET2019,
title = {ADMET modeling approaches in drug discovery},
journal = {Drug Discovery Today},
volume = {24},
number = {5},
pages = {1157-1165},
year = {2019},
issn = {1359-6446},
doi = {https://doi.org/10.1016/j.drudis.2019.03.015},
url = {https://www.sciencedirect.com/science/article/pii/S1359644618303301},
author = {Leonardo L.G. Ferreira and Adriano D. Andricopulo}
}

@article{singhal2023large,
  title={Large language models encode clinical knowledge},
  author={Singhal, Karan and Azizi, Shekoofeh and Tu, Tao and Mahdavi, S Sara and Wei, Jason and Chung, Hyung Won and Scales, Nathan and Tanwani, Ajay and Cole-Lewis, Heather and Pfohl, Stephen and others},
  journal={Nature},
  volume={620},
  number={7972},
  pages={172--180},
  year={2023},
  publisher={Nature Publishing Group UK London}
}

@article{saab2024capabilities,
  title={Capabilities of gemini models in medicine},
  author={Saab, Khaled and Tu, Tao and Weng, Wei-Hung and Tanno, Ryutaro and Stutz, David and Wulczyn, Ellery and Zhang, Fan and Strother, Tim and Park, Chunjong and Vedadi, Elahe and others},
  journal={arXiv preprint arXiv:2404.18416},
  year={2024}
}

@article{tian2024opportunities,
  title={Opportunities and challenges for ChatGPT and large language models in biomedicine and health},
  author={Tian, Shubo and Jin, Qiao and Yeganova, Lana and Lai, Po-Ting and Zhu, Qingqing and Chen, Xiuying and Yang, Yifan and Chen, Qingyu and Kim, Won and Comeau, Donald C and others},
  journal={Briefings in Bioinformatics},
  volume={25},
  number={1},
  pages={bbad493},
  year={2024},
  publisher={Oxford University Press}
}

@misc{xiong2025reliablescientifichypothesisgeneration,
      title={Toward Reliable Scientific Hypothesis Generation: Evaluating Truthfulness and Hallucination in Large Language Models}, 
      author={Guangzhi Xiong and Eric Xie and Corey Williams and Myles Kim and Amir Hassan Shariatmadari and Sikun Guo and Stefan Bekiranov and Aidong Zhang},
      year={2025},
      eprint={2505.14599},
      archivePrefix={arXiv},
      primaryClass={cs.CL},
      url={https://arxiv.org/abs/2505.14599}, 
}

@misc{nsfPublicationsOutput,
	author = {Karen White},
	title = {{P}ublications {O}utput: {U}.{S}. {T}rends and {I}nternational {C}omparisons | {N}{S}{F} - {N}ational {S}cience {F}oundation --- ncses.nsf.gov},
	howpublished = {\url{https://ncses.nsf.gov/pubs/nsb20214}},
	year = {2021},
	note = {[Accessed 12-March-2026]},
}

@inproceedings{betz2022adversarial,
  title     = {Adversarial Explanations for Knowledge Graph Embeddings},
  author    = {Betz, Patrick and Meilicke, Christian and Stuckenschmidt, Heiner},
  booktitle = {Proceedings of the Thirty-First International Joint Conference on
               Artificial Intelligence, {IJCAI-22}},
  publisher = {International Joint Conferences on Artificial Intelligence Organization},
  editor    = {Lud De Raedt},
  pages     = {2820--2826},
  year      = {2022},
  month     = {7},
  note      = {Main Track},
  doi       = {10.24963/ijcai.2022/391},
  url       = {https://doi.org/10.24963/ijcai.2022/391},
}

@misc{alkan2025surveyhypothesisgenerationscientific,
      title={A Survey on Hypothesis Generation for Scientific Discovery in the Era of Large Language Models}, 
      author={Atilla Kaan Alkan and Shashwat Sourav and Maja Jablonska and Simone Astarita and Rishabh Chakrabarty and Nikhil Garuda and Pranav Khetarpal and Maciej Pióro and Dimitrios Tanoglidis and Kartheik G. Iyer and Mugdha S. Polimera and Michael J. Smith and Tirthankar Ghosal and Marc Huertas-Company and Sandor Kruk and Kevin Schawinski and Ioana Ciucă},
      year={2025},
      eprint={2504.05496},
      archivePrefix={arXiv},
      primaryClass={cs.CL},
      url={https://arxiv.org/abs/2504.05496}, 
}

@misc{bhasuran2023literaturebaseddiscoverylbd,
      title={Literature Based Discovery (LBD): Towards Hypothesis Generation and Knowledge Discovery in Biomedical Text Mining}, 
      author={Balu Bhasuran and Gurusamy Murugesan and Jeyakumar Natarajan},
      year={2023},
      eprint={2310.03766},
      archivePrefix={arXiv},
      primaryClass={cs.IR},
      url={https://arxiv.org/abs/2310.03766}, 
}

@misc{gottweis2025aicoscientist,
      title={Towards an AI co-scientist}, 
      author={Juraj Gottweis and Wei-Hung Weng and Alexander Daryin and Tao Tu and Anil Palepu and Petar Sirkovic and Artiom Myaskovsky and Felix Weissenberger and Keran Rong and Ryutaro Tanno and Khaled Saab and Dan Popovici and Jacob Blum and Fan Zhang and Katherine Chou and Avinatan Hassidim and Burak Gokturk and Amin Vahdat and Pushmeet Kohli and Yossi Matias and Andrew Carroll and Kavita Kulkarni and Nenad Tomasev and Yuan Guan and Vikram Dhillon and Eeshit Dhaval Vaishnav and Byron Lee and Tiago R D Costa and José R Penadés and Gary Peltz and Yunhan Xu and Annalisa Pawlosky and Alan Karthikesalingam and Vivek Natarajan},
      year={2025},
      eprint={2502.18864},
      archivePrefix={arXiv},
      primaryClass={cs.AI},
      url={https://arxiv.org/abs/2502.18864}, 
}

@article{Ashburner2002GO,
  title={Gene Ontology: tool for the unification of biology},
  volume={25},
  ISSN={1061-4036},
  url={https://doi.org/10.1038/75556},
  DOI={10.1038/75556},
  number={1},
  journal={Nature Genetics},
  publisher={Springer Science and Business Media LLC},
  author={Ashburner, Michael and Ball, Catherine A. and Blake, Judith A. and Botstein, David and Butler, Heather and Cherry, J. Michael and Davis, Allan P. and Dolinski, Kara and Dwight, Selina S. and Eppig, Janan T. and Harris, Midori A. and Hill, David P. and Issel-Tarver, Laurie and Kasarskis, Andrew and Lewis, Suzanna and Matese, John C. and Richardson, Joel E. and Ringwald, Martin and Rubin, Gerald M. and Sherlock, Gavin},
  year={2002},
  pages={25-29}
}

@article{Ristoski2016RDF2Vec,
  title={RDF2Vec: RDF Graph Embeddings for Data Mining},
  volume={},
  ISSN={0302-9743},
  url={https://doi.org/10.1007/978-3-319-46523-4_30},
  DOI={10.1007/978-3-319-46523-4_30},
  number={},
  journal={Lecture Notes in Computer Science},
  publisher={Springer International Publishing},
  author={Ristoski, Petar and Paulheim, Heiko},
  year={2016},
  pages={498-514}
}

@inproceedings{Mikolov2013EfficientEO,
  title={Efficient Estimation of Word Representations in Vector Space},
  author={Tomas Mikolov and Kai Chen and Gregory S. Corrado and Jeffrey Dean},
  booktitle={International Conference on Learning Representations},
  year={2013},
  url={https://api.semanticscholar.org/CorpusID:5959482}
}

@article{Ying2019GNNExplainerGE,
  title={GNNExplainer: Generating Explanations for Graph Neural Networks},
  author={Rex Ying and Dylan Bourgeois and Jiaxuan You and Marinka Zitnik and Jure Leskovec},
  journal={Advances in neural information processing systems},
  year={2019},
  volume={32},
  pages={
          9240-9251
        },
  url={https://api.semanticscholar.org/CorpusID:202572927}
}

@article{Luo2020ParameterizedEF,
  title={Parameterized Explainer for Graph Neural Network},
  author={Dongsheng Luo and Wei Cheng and Dongkuan Xu and Wenchao Yu and Bo Zong and Haifeng Chen and Xiang Zhang},
  journal={ArXiv},
  year={2020},
  volume={abs/2011.04573},
  url={https://api.semanticscholar.org/CorpusID:226281363}
}

@inproceedings{yuan2021on,
  title     = {On Explainability of Graph Neural Networks via Subgraph Explorations},
  author    = {Yuan, Hao and Yu, Haiyang and Wang, Jie and Li, Kang and Ji, Shuiwang},
  booktitle = {Proceedings of the 38th International Conference on Machine Learning},
  series    = {Proceedings of Machine Learning Research},
  volume    = {139},
  pages     = {12241--12252},
  year      = {2021},
  editor    = {Meila, Marina and Zhang, Tong},
  publisher = {PMLR},
  url       = {http://proceedings.mlr.press/v139/yuan21c.html}
}

@inproceedings{pezeshkpour2019investigating,
    title = "{Investigating Robustness and Interpretability of Link Prediction via Adversarial Modifications}",
    author = "Pezeshkpour, Pouya  and
      Tian, Yifan  and
      Singh, Sameer",
    editor = "Burstein, Jill  and
      Doran, Christy  and
      Solorio, Thamar",
    booktitle = "Proceedings of the 2019 Conference of the North {A}merican Chapter of the Association for Computational Linguistics: Human Language Technologies, Volume 1 (Long and Short Papers)",
    month = jun,
    year = "2019",
    address = "Minneapolis, Minnesota",
    publisher = "Association for Computational Linguistics",
    url = "https://aclanthology.org/N19-1337",
    doi = "10.18653/v1/N19-1337",
    pages = "3336--3347",
}

@inproceedings{bhardwaj2021adversarial,
    title = "{Adversarial Attacks on Knowledge Graph Embeddings via Instance Attribution Methods}",
    author = "Bhardwaj, Peru  and
      Kelleher, John  and
      Costabello, Luca  and
      O{'}Sullivan, Declan",
    editor = "Moens, Marie-Francine  and
      Huang, Xuanjing  and
      Specia, Lucia  and
      Yih, Scott Wen-tau",
    booktitle = "Proceedings of the 2021 Conference on Empirical Methods in Natural Language Processing",
    month = nov,
    year = "2021",
    address = "Online and Punta Cana, Dominican Republic",
    publisher = "Association for Computational Linguistics",
    url = "https://aclanthology.org/2021.emnlp-main.648",
    doi = "10.18653/v1/2021.emnlp-main.648",
    pages = "8225--8239",
}

@article{Dao2022FlashAttentionFA,
  title={FlashAttention: Fast and Memory-Efficient Exact Attention with IO-Awareness},
  author={Tri Dao and Daniel Y. Fu and Stefano Ermon and Atri Rudra and Christopher R'e},
  journal={ArXiv},
  year={2022},
  volume={abs/2205.14135},
  url={https://api.semanticscholar.org/CorpusID:249151871}
}

@inproceedings{Zadrozny2001ObtainingCP,
  title={Obtaining calibrated probability estimates from decision trees and naive Bayesian classifiers},
  author={Bianca Zadrozny and Charles Peter Elkan},
  booktitle={International Conference on Machine Learning},
  year={2001},
  url={https://api.semanticscholar.org/CorpusID:9594071}
}

@inproceedings{Platt1999ProbabilisticOF,
  title={Probabilistic Outputs for Support vector Machines and Comparisons to Regularized Likelihood Methods},
  author={John Platt},
  year={1999},
  url={https://api.semanticscholar.org/CorpusID:56563878}
}

@inproceedings{Vaswani2017AttentionIA,
  title={Attention is All you Need},
  author={Ashish Vaswani and Noam Shazeer and Niki Parmar and Jakob Uszkoreit and Llion Jones and Aidan N. Gomez and Lukasz Kaiser and Illia Polosukhin},
  booktitle={Neural Information Processing Systems},
  year={2017},
  url={https://api.semanticscholar.org/CorpusID:13756489}
}

@inproceedings{rossi2022explaining,
    author = {Rossi, Andrea and Firmani, Donatella and Merialdo, Paolo and Teofili, Tommaso},
    title = {{Explaining Link Prediction Systems based on Knowledge Graph Embeddings}},
    year = {2022},
    isbn = {9781450392495},
    publisher = {Association for Computing Machinery},
    address = {New York, NY, USA},
    url = {https://doi.org/10.1145/3514221.3517887},
    doi = {10.1145/3514221.3517887},
    booktitle = {Proceedings of the 2022 International Conference on Management of Data},
    pages = {2062–2075},
    numpages = {14},
    location = {Philadelphia, PA, USA},
    series = {SIGMOD '22}
}

@InProceedings{barile2025addititve,
author="Barile, Roberto
and d'Amato, Claudia
and Fanizzi, Nicola",
editor="Alam, Mehwish
and Rospocher, Marco
and van Erp, Marieke
and Hollink, Laura
and Gesese, Genet Asefa",
title={{Additive Counterfactuals for Explaining Link Predictions on Knowledge Graphs}},
booktitle="Knowledge Engineering and Knowledge Management",
year="2024",
publisher="Springer Nature Switzerland",
address="Cham",
pages="346--363",
isbn="978-3-031-77792-9"
}

@article{Nickel2015ARO,
  title={A Review of Relational Machine Learning for Knowledge Graphs},
  author={Maximilian Nickel and Kevin P. Murphy and Volker Tresp and Evgeniy Gabrilovich},
  journal={Proceedings of the IEEE},
  year={2015},
  volume={104},
  pages={11-33},
  url={https://api.semanticscholar.org/CorpusID:12161567}
}

@article{Lonardi2025UnifyingPE,
  title={Unifying Post-hoc Explanations of Knowledge Graph Completions},
  author={Alessandro Lonardi and Samy Badreddine and Tarek R. Besold and Pablo Sanchez Martin},
  journal={ArXiv},
  year={2025},
  volume={abs/2507.22951},
  url={https://api.semanticscholar.org/CorpusID:280400998}
}

@article{Peng2023KnowledgeGO,
  title={Knowledge Graphs: Opportunities and Challenges},
  author={Ciyuan Peng and Feng Xia and Mehdi Naseriparsa and Francesco Osborne},
  journal={Artificial Intelligence Review},
  year={2023},
  pages={1 - 32},
  url={https://api.semanticscholar.org/CorpusID:257757244}
}

@article{Barabsi2004NetworkBU,
  title={Network biology: understanding the cell's functional organization},
  author={Albert-Laszl{\'o} Barab{\'a}si and Zolt{\'a}n N. Oltvai},
  journal={Nature Reviews Genetics},
  year={2004},
  volume={5},
  pages={101-113},
  url={https://api.semanticscholar.org/CorpusID:10950726}
}

@inproceedings{zhang2023page,
author = {Zhang, Shichang and Zhang, Jiani and Song, Xiang and Adeshina, Soji and Zheng, Da and Faloutsos, Christos and Sun, Yizhou},
title = {PaGE-Link: Path-based Graph Neural Network Explanation for Heterogeneous Link Prediction},
year = {2023},
isbn = {9781450394161},
publisher = {Association for Computing Machinery},
address = {New York, NY, USA},
url = {https://doi.org/10.1145/3543507.3583511},
doi = {10.1145/3543507.3583511},
booktitle = {Proceedings of the ACM Web Conference 2023},
pages = {3784–3793},
numpages = {10},
location = {Austin, TX, USA},
series = {WWW '23}
}

@Article{huang2024foundation,
author={Huang, Kexin
and Chandak, Payal
and Wang, Qianwen
and Havaldar, Shreyas
and Vaid, Akhil
and Leskovec, Jure
and Nadkarni, Girish N.
and Glicksberg, Benjamin S.
and Gehlenborg, Nils
and Zitnik, Marinka},
title={A foundation model for clinician-centered drug repurposing},
journal={Nature Medicine},
year={2024},
month={Sep},
day={25},
issn={1546-170X},
doi={10.1038/s41591-024-03233-x},
url={https://doi.org/10.1038/s41591-024-03233-x}
}

@inproceedings{chang2024path,
series={KDD ’24},
   title={Path-based Explanation for Knowledge Graph Completion},
   url={http://dx.doi.org/10.1145/3637528.3671683},
   DOI={10.1145/3637528.3671683},
   booktitle={Proceedings of the 30th ACM SIGKDD Conference on Knowledge Discovery and Data Mining},
   publisher={ACM},
   author={Chang, Heng and Ye, Jiangnan and Lopez-Avila, Alejo and Du, Jinhua and Li, Jia},
   year={2024},
   month=aug, pages={231–242},
   collection={KDD ’24} }

@misc{ma2024kgexplainer,
      title={{KGExplainer: Towards Exploring Connected Subgraph Explanations for Knowledge Graph Completion}}, 
      author={Tengfei Ma and Xiang song and Wen Tao and Mufei Li and Jiani Zhang and Xiaoqin Pan and Jianxin Lin and Bosheng Song and xiangxiang Zeng},
      year={2024},
      eprint={2404.03893},
      archivePrefix={arXiv},
      primaryClass={cs.AI},
      url={https://arxiv.org/abs/2404.03893}, 
}

@inproceedings{zhang2010iteratively,
author = {Zhang, Wen and Paudel, Bibek and Wang, Liang and Chen, Jiaoyan and Zhu, Hai and Zhang, Wei and Bernstein, Abraham and Chen, Huajun},
title = {{Iteratively Learning Embeddings and Rules for Knowledge Graph Reasoning}},
year = {2019},
isbn = {9781450366748},
publisher = {Association for Computing Machinery},
address = {New York, NY, USA},
url = {https://doi.org/10.1145/3308558.3313612},
doi = {10.1145/3308558.3313612},
booktitle = {The World Wide Web Conference},
pages = {2366–2377},
numpages = {12},
location = {San Francisco, CA, USA},
series = {WWW '19}
}

@inproceedings{sadeghian2019drum,
 author = {Sadeghian, Ali and Armandpour, Mohammadreza and Ding, Patrick and Wang, Daisy Zhe},
 booktitle = {Advances in Neural Information Processing Systems},
 editor = {H. Wallach and H. Larochelle and A. Beygelzimer and F. d\textquotesingle Alch\'{e}-Buc and E. Fox and R. Garnett},
 pages = {},
 publisher = {Curran Associates, Inc.},
 title = {{DRUM: End-To-End Differentiable Rule Mining On Knowledge Graphs}},
 volume = {32},
 year = {2019}
}

@inproceedings{
arakelyan2021complex,
title={Complex Query Answering with Neural Link Predictors},
author={Erik Arakelyan and Daniel Daza and Pasquale Minervini and Michael Cochez},
booktitle={International Conference on Learning Representations},
year={2021},
url={https://openreview.net/forum?id=Mos9F9kDwkz}
}

@inproceedings{gu2023iae,
    title = "{{IAE}val: A Comprehensive Evaluation of Instance Attribution on Natural Language Understanding}",
    author = "Gu, Peijian  and
      Shen, Yaozong  and
      Wang, Lijie  and
      Wang, Quan  and
      Wu, Hua  and
      Mao, Zhendong",
    editor = "Bouamor, Houda  and
      Pino, Juan  and
      Bali, Kalika",
    booktitle = "Findings of the Association for Computational Linguistics: EMNLP 2023",
    month = dec,
    year = "2023",
    address = "Singapore",
    publisher = "Association for Computational Linguistics",
    url = "https://aclanthology.org/2023.findings-emnlp.801/",
    doi = "10.18653/v1/2023.findings-emnlp.801",
    pages = "11966--11977",
}

@inproceedings{zhang2019data,
  title     = {{Data Poisoning Attack against Knowledge Graph Embedding}},
  author    = {Zhang, Hengtong and Zheng, Tianhang and Gao, Jing and Miao, Chenglin and Su, Lu and Li, Yaliang and Ren, Kui},
  booktitle = {Proceedings of the Twenty-Eighth International Joint Conference on
               Artificial Intelligence, {IJCAI-19}},
  publisher = {International Joint Conferences on Artificial Intelligence Organization},
  pages     = {4853--4859},
  year      = {2019},
  month     = {7},
  doi       = {10.24963/ijcai.2019/674},
  url       = {https://doi.org/10.24963/ijcai.2019/674},
}

@incollection{hamilton2017inductive,
  title     = {Inductive Representation Learning on Large Graphs},
  author    = {Hamilton, William L. and Ying, Rex and Leskovec, Jure},
  booktitle = {Advances in Neural Information Processing Systems},
  volume    = {30},
  year      = {2017},
  publisher = {Curran Associates, Inc.}
}

@article{zeng2016link,
  title={Link prediction based on local information considering preferential attachment},
  author={Zeng, Shan},
  journal={Physica A: Statistical Mechanics and its Applications},
  volume={443},
  pages={537--542},
  year={2016},
  publisher={Elsevier}
}

@article{Zhao2023KEXTS,
  title={KE-X: Towards subgraph explanations of knowledge graph embedding based on knowledge information gain},
  author={Dong Zhao and Guojia Wan and Yibing Zhan and Zengmao Wang and Liang Ding and Zhigao Zheng and Bo Du},
  journal={Knowl. Based Syst.},
  year={2023},
  volume={278},
  pages={110772},
  url={https://api.semanticscholar.org/CorpusID:260173519}
}

@article{Kosan2022GlobalCE,
  title={Global Counterfactual Explainer for Graph Neural Networks},
  author={Mert Kosan and Zexi Huang and Sourav Medya and Sayan Ranu and Ambuj K. Singh},
  journal={Proceedings of the Sixteenth ACM International Conference on Web Search and Data Mining},
  year={2022},
  url={https://api.semanticscholar.org/CorpusID:253080473}
}

@inproceedings{Bajaj2021RobustCE,
  title={Robust Counterfactual Explanations on Graph Neural Networks},
  author={Mohit Bajaj and Lingyang Chu and Zihui Xue and Jian Pei and Lanjun Wang and Peter Cho-Ho Lam and Yong Zhang},
  booktitle={Neural Information Processing Systems},
  year={2021},
  url={https://api.semanticscholar.org/CorpusID:235790538}
}

@inproceedings{Lucic2021CFGNNExplainerCE,
  title={CF-GNNExplainer: Counterfactual Explanations for Graph Neural Networks},
  author={Ana Lucic and Maartje ter Hoeve and Gabriele Tolomei and M. de Rijke and Fabrizio Silvestri},
  booktitle={International Conference on Artificial Intelligence and Statistics},
  year={2021},
  url={https://api.semanticscholar.org/CorpusID:231839528}
}

@article{Huang2020GraphLIMELI,
  title={GraphLIME: Local Interpretable Model Explanations for Graph Neural Networks},
  author={Q. Huang and Makoto Yamada and Yuan Tian and Dinesh Singh and Dawei Yin and Yi Chang},
  journal={IEEE Transactions on Knowledge and Data Engineering},
  year={2020},
  volume={35},
  pages={6968-6972},
  url={https://api.semanticscholar.org/CorpusID:210714016}
}

@inproceedings{hamilton2017graphsage,
  author    = {William L. Hamilton and Zhitao Ying and Jure Leskovec},
  title     = {Inductive Representation Learning on Large Graphs},
  booktitle = {Advances in Neural Information Processing Systems 30: Annual Conference on Neural Information Processing Systems 2017, December 4--9, 2017, Long Beach, CA, USA},
  pages     = {1024--1034},
  year      = {2017},
  url       = {https://proceedings.neurips.cc/paper/2017/hash/5dd9db5e033da9c6fb5ba83c7a7ebea9-Abstract.html}
}

@inproceedings{vaswani2017attention,
  author    = {Ashish Vaswani and Noam Shazeer and Niki Parmar and Jakob Uszkoreit and Llion Jones and Aidan N. Gomez and Lukasz Kaiser and Illia Polosukhin},
  title     = {Attention is All You Need},
  booktitle = {Advances in Neural Information Processing Systems 30: Annual Conference on Neural Information Processing Systems 2017, December 4--9, 2017, Long Beach, CA, USA},
  pages     = {5998--6008},
  year      = {2017},
  url       = {https://proceedings.neurips.cc/paper/2017/hash/3f5ee243547dee91fbd053c1c4a845aa-Abstract.html}
}

@inproceedings{lin2017focal,
  author    = {Tsung{-}Yi Lin and Priya Goyal and Ross B. Girshick and Kaiming He and Piotr Doll{\'{a}}r},
  title     = {Focal Loss for Dense Object Detection},
  booktitle = {IEEE International Conference on Computer Vision (ICCV), Venice, Italy, October 22--29, 2017},
  pages     = {2999--3007},
  publisher = {IEEE Computer Society},
  year      = {2017},
  doi       = {10.1109/ICCV.2017.324},
  url       = {https://doi.org/10.1109/ICCV.2017.324}
}

@article{dynamicloss,
  author    = {K. Ruwani M. Fernando and Chris P. Tsokos},
  title     = {Dynamically Weighted Balanced Loss: Class Imbalanced Learning and Confidence Calibration of Deep Neural Networks},
  journal   = {IEEE Transactions on Neural Networks and Learning Systems},
  year      = {2022},
  volume    = {33},
  number    = {7},
  pages     = {2940--2951},
  doi       = {10.1109/TNNLS.2020.3047335}
}

@article{bairos2024,
  title={Sterol O-acyltransferase (SOAT/ACAT) activity is required to form cholesterol crystals in hepatocyte lipid droplets},
  author={Bairos, J. A. and others},
  journal={BBA-Molecular and Cell Biology Lipids},
  volume={1869},
  pages={159512},
  year={2024}
}

@article{callahan2014,
  title={Paradoxical activation of T cells via augmented ERK signaling mediated by a RAF inhibitor},
  author={Callahan, M. K. and Masters, G. and Pratilas, C. A. and Ariyan, C. and Katz, J. and others},
  journal={Cancer Immunol. Res.},
  volume={2},
  number={1},
  pages={70--79},
  year={2014}
}

@misc{medchemexpress2025a,
  author={MedChemExpress},
  title={Product data sheet: AZ-628 (Catalog No. HY-11004)},
  year={2025},
  note={Available at: \url{https://file.medchemexpress.com/batch_PDF/HY-11004/}}
}

@misc{medchemexpress2025b,
  author={MedChemExpress},
  title={Product data sheet: ZM-336372 (Catalog No. HY-13343)},
  year={2025},
  note={Available at: \url{https://file.medchemexpress.com/batch_PDF/HY-13343/}}
}

@article{molina2006,
  title={The Ras/Raf/MAPK pathway},
  author={Molina, J. R. and Adjei, A. A.},
  journal={J. Thorac. Oncol.},
  volume={1},
  number={1},
  pages={7--9},
  year={2006}
}

@article{parameswaran2010,
  title={Tumour Necrosis Factor-$\alpha$ Signaling in Macrophages},
  author={Parameswaran, N. and Patial, S.},
  journal={Crit. Rev. Eukaryot. Gene Expr.},
  volume={20},
  number={2},
  pages={87--103},
  year={2010}
}

@article{sekiya2004,
  title={Transcriptional regulation of the TGF-$\beta$ pseudoreceptor BAMBI by TGF-$\beta$ signaling},
  author={Sekiya, T. and Oda, T. and Matsuura, K. and Akiyama, T.},
  journal={Biochem. Biophys. Res. Commun.},
  volume={320},
  pages={680--684},
  year={2004}
}

@article{solyakov2009,
  title={Regulation of p53 expression, phosphorylation and subcellular localization by a G-protein-coupled receptor},
  author={Solyakov, L. and Sayan, E. and Riley, J. and Pointon, A. and Tobin, A. B.},
  journal={Oncogene},
  volume={28},
  pages={3619--3630},
  year={2009}
}

@misc{tocris2016,
  author={Tocris Bioscience},
  title={GW5074 [Product datasheet]},
  year={2016},
  note={Available at: \url{https://documents.tocris.com/pdfs/tocris_coa/4836_1_coa.pdf}}
}

@inproceedings{sybrandt_agatha_2020,
	address = {New York, NY, USA},
	series = {{CIKM} '20},
	title = {{AGATHA}: {Automatic} {Graph} {Mining} {And} {Transformer} based {Hypothesis} {Generation} {Approach}},
	isbn = {978-1-4503-6859-9},
	shorttitle = {{AGATHA}},
	url = {https://dl.acm.org/doi/10.1145/3340531.3412684},
	doi = {10.1145/3340531.3412684},
	urldate = {2026-02-27},
	booktitle = {Proceedings of the 29th {ACM} {International} {Conference} on {Information} \& {Knowledge} {Management}},
	publisher = {Association for Computing Machinery},
	author = {Sybrandt, Justin and Tyagin, Ilya and Shtutman, Michael and Safro, Ilya},
	year = {2020},
	pages = {2757--2764},
}

@inproceedings{singer_node_2019,
	address = {Macao, China},
	title = {Node {Embedding} over {Temporal} {Graphs}},
	isbn = {978-0-9992411-4-1},
	url = {https://www.ijcai.org/proceedings/2019/640},
	doi = {10.24963/ijcai.2019/640},
	language = {en},
	urldate = {2026-02-27},
	booktitle = {Proceedings of the {Twenty}-{Eighth} {International} {Joint} {Conference} on {Artificial} {Intelligence}},
	publisher = {International Joint Conferences on Artificial Intelligence Organization},
	author = {Singer, Uriel and Guy, Ido and Radinsky, Kira},
	month = aug,
	year = {2019},
	pages = {4605--4612},
}

@misc{trouillon_complex_2016,
	title = {Complex {Embeddings} for {Simple} {Link} {Prediction}},
	url = {http://arxiv.org/abs/1606.06357},
	doi = {10.48550/arXiv.1606.06357},
	urldate = {2026-02-27},
	publisher = {arXiv},
	author = {Trouillon, Theo and Welbl, Johannes and Riedel, Sebastian and Gaussier, Eric and Bouchard, Guillaume},
	month = jun,
	year = {2016},
	note = {arXiv:1606.06357 [cs]},
}

@article{akujuobi_link_2024,
	title = {Link prediction for hypothesis generation: an active curriculum learning infused temporal graph-based approach},
	volume = {57},
	issn = {1573-7462},
	shorttitle = {Link prediction for hypothesis generation},
	url = {https://doi.org/10.1007/s10462-024-10885-1},
	doi = {10.1007/s10462-024-10885-1},
	language = {en},
	number = {9},
	urldate = {2026-02-27},
	journal = {Artificial Intelligence Review},
	author = {Akujuobi, Uchenna and Kumari, Priyadarshini and Choi, Jihun and Badreddine, Samy and Maruyama, Kana and Palaniappan, Sucheendra K. and Besold, Tarek R.},
	month = aug,
	year = {2024},
	pages = {244},
}

\end{document}